\documentclass{article}

\usepackage[preprint]{neurips_2026}

\usepackage[utf8]{inputenc} 
\usepackage[T1]{fontenc}    
\usepackage{url}            
\usepackage{booktabs}       
\usepackage{nicefrac}       
\usepackage{microtype}      

\usepackage{cite}
\usepackage{xcolor}
\usepackage{amsmath,amssymb,amsfonts}
\usepackage{graphicx}
\usepackage{tabularx}
\usepackage{textcomp}
\usepackage{lineno,hyperref}
\usepackage{multirow}
\usepackage{subcaption}

\usepackage[textsize=tiny]{todonotes}

\title{Dataset Scarcity Limits Robust Evaluation of Multilingual Embedding Models: A Case Study of Slavic Languages}

\author{
  Ana Gjorgjevikj\\
  Computer Systems Department\\
  Jožef Stefan Institute\\
  Ljubljana, Slovenia\\
  \texttt{ana.gjorgjevikj@ijs.si} \\
  \AND
  Barbara Korou\v{s}i\'{c} Seljak \\
  Computer Systems Department\\
  Jožef Stefan Institute\\
  Ljubljana, Slovenia\\
  \texttt{barbara.korousic@ijs.si} \\
  \AND
  Tome Eftimov \\
  Computer Systems Department\\
  Jožef Stefan Institute\\
  Ljubljana, Slovenia\\
  \texttt{tome.eftimov@ijs.si} \\
}

\begin{document}

\maketitle

\begin{abstract}
Multilingual text embedding models enable cross-lingual transfer of knowledge across a wide range of NLP tasks, but their evaluation remains highly uneven across high-, mid-, and low-resource languages. In this paper, we propose a two-dimensional framework, specifically tailored for analyzing robustness of multilingual embedding benchmarks under dataset scarcity, and apply it on the Slavic-language subset of the MTEB benchmark. The framework distinguishes between task-specific and cross-task evaluation, while jointly analyzing three complementary aspects: (1) ranking robustness, (2) model consistency, and (3) evidence strength. At the task-specific level, we evaluate the stability of model rankings under changes in ranking methodology and benchmark dataset composition. At the cross-task level, we assess the robustness of models across diverse tasks within a language. To quantify the reliability of benchmark conclusions, we introduce an Evidence Strength Score that accounts for dataset availability, diversity, and robustness assessability. Our analysis reveals severe benchmark sparsity, with many Slavic language-task pairs relying on a single dataset or highly correlated benchmark collections in the analyzed benchmark snapshot, limiting the ability to draw robust conclusions. The cross-task analysis reveals a small group of highly robust models, most notably \textit{llama-embed-nemotron-8b}, \textit{multilingual-e5-large-instruct}, and \textit{Qwen3-Embedding} variants, that consistently perform well across Slavic languages and tasks. Overall, the results demonstrate that benchmark rankings and robustness conclusions must be interpreted jointly with certain notation of their evidence strength and highlight benchmark scarcity as a major obstacle to trustworthy multilingual evaluation.

\end{abstract}

\section{Introduction}

Large-scale multilingual text embedding models have become a cornerstone of modern natural language processing~(NLP), enabling knowledge transfer from high-resource to low-resource languages~\citep{feng2022language,reimers2019sentence,wang2024multilingual}. Despite their promise, their evaluation remains highly uneven across languages, particularly for mid- and low-resource languages where both dataset availability and task coverage are limited. Multilingual benchmarking platforms such as MTEB~\citep{muennighoff2023mteb} and its massively multilingual extension MMTEB~\citep{enevoldsen2025mmteb} report results across hundreds of languages, tasks, and datasets, yet provide limited insight into how reliably these results reflect genuine model robustness under sparse and imbalanced evaluation conditions.
 
Multilingual benchmarks typically aggregate performance through simple averaging of heterogeneous metrics (e.g., accuracy, F1), implicitly assuming comparability between datasets while ignoring correlations among them and imbalances in task and language coverage. This limitation is particularly pronounced in multilingual settings, where dataset availability varies substantially across tasks and languages~\citep{gjorgjevikj2026robustness}. Critically, apparently stable benchmark conclusions may rest on very limited evidence. For example, a model may consistently rank first across different aggregation methods when only a single dataset exists, yet this agreement provides no evidence of genuine robustness. Likewise, multiple highly correlated datasets can create the illusion of broad evaluation coverage while contributing little independent information. Prior work has addressed ranking instability in monolingual settings using alternative aggregation strategies~\citep{colombo2022best,rofin2023vote} and has shown that performance can be sensitive to input variation~\citep{frank2026pteb}, but these approaches do not account for the qualitative and quantitative reliability of the underlying evaluation evidence - a distinction that becomes central in low-resource multilingual settings.
 
Slavic languages constitute a major branch of the Indo-European family, encompassing East (e.g., Russian, Ukrainian, Belarusian), West~(e.g., Polish, Czech, Slovak), and South Slavic languages (e.g., Slovene, Croatian, Serbian, Bulgarian, Macedonian). Collectively spoken by over 300 million people~\citep{sussex2006slavic} and nearly one-third of the speakers of the European Union's official languages~\citep{piskorski2023proceedings}, they represent a linguistically significant and diverse group. Nevertheless, many Slavic languages remain underrepresented in NLP resources and systems~\citep{piskorski2025proceedings,piskorski2023proceedings}. Their linguistic properties, e.g., rich morphology, relatively free word order, and the coexistence of Latin and Cyrillic scripts, pose additional challenges for NLP tools and cross-lingual generalization~\citep{piskorski2025proceedings,piskorski2023proceedings}, making them a valuable stress test for multilingual embedding evaluation under scarce and imbalanced benchmarking conditions.
 
\paragraph{Contribution}
 
We propose a two-dimensional framework for analyzing multilingual embedding benchmarks under dataset scarcity and imbalance. The framework operates across two evaluation scopes:
\begin{itemize}
  \item \textit{Task-specific analysis}: evaluates overall ranking stability and model top-$k$ consistency within a fixed language and task.
  \item \textit{Cross-task analysis}: evaluates whether models consistently stay robust across task families within a language, measured via cross-task consistency analysis.
\end{itemize}
Within each scope, three complementary aspects are analyzed jointly: (i)~\textit{ranking stability}, which assesses whether benchmark rankings remain stable under different aggregation methods (\textit{aggregation stability}) and dataset compositions (\textit{composition stability}); (ii)~\textit{top-$k$ consistency}, which assesses whether individual models reliably remain among the top-performing methods across evaluation conditions; and (iii)~\textit{evidence strength}, which quantifies how much confidence can actually be placed in the stability conclusions for a given language--task pair.

The ranking stability and top-$k$ consistency components build on our prior work~\citep{gjorgjevikj2026robustness}. \textbf{The novel contribution of this paper} is the evidence strength component, introduced specifically to address the low-resource multilingual setting. We define a set of qualitative \textit{evidence levels} that characterize which types of stability analysis are feasible for each task-language pair, and a quantitative \textit{Evidence Strength Score~(ESS)} that measures how reliable the available benchmark evidence actually is. ESS combines dataset availability, effective dataset diversity (accounting for redundancy between correlated datasets), and stability assessability, i.e., whether aggregation stability, composition stability, or neither can be meaningfully evaluated. This separation between stability and evidence strength allows users to distinguish genuine model stability from apparent stability arising from sparse or redundant benchmark resources.
 
Applied to the Slavic-language subset of MTEB, our analysis reveals severe benchmark scarsity and redundancy, with most language-task pairs relying on a single dataset or lacking sufficient diversity for reliable robustness assessment. Despite this, a small group of large multilingual models (most notably \textit{llama-embed-nemotron-8b}, \textit{multilingual-e5-large-instruct}, and \textit{Qwen3-Embedding} variants) consistently demonstrates stable cross-task performance, while most models exhibit strong task- and language-specific behavior. Overall, the results highlight that benchmark rankings and robustness conclusions must be interpreted jointly with their evidence strength, and underscore benchmark scarcity as a major obstacle to trustworthy multilingual evaluation.

\section{Related Work}

\paragraph{Multilingual text embedding models}
Multilingual embedding models such as LaBSE~\citep{feng2022language}
and multilingual Sentence-BERT~\citep{reimers2019sentence} map multiple
languages into a shared space, enabling cross-lingual tasks such as
retrieval and semantic similarity.
Recent models build on large-scale contrastive training and foundation
model architectures.
For example, multilingual E5~\citep{wang2024multilingual} leverages
large-scale text pairs for improved cross-lingual alignment, while
newer LLM-based embedding models (e.g., Qwen3~\citep{zhang2025qwen3}
and LLaMA-based variants~\citep{babakhin2025llama}) further improve
performance across tasks and languages.
 
\paragraph{Multilingual embedding benchmarks}
MTEB~\citep{muennighoff2023mteb} introduced a unified benchmark
covering multiple tasks and languages, showing that no single model
dominates across all settings.
MMTEB~\citep{enevoldsen2025mmteb} extends this to over 500 datasets
and 250 languages.
However, both rely on aggregate scores, limiting language-specific
analysis.
Recent work emphasizes the need for language-specific evaluation.
Examples include PL-MTEB~\citep{poswiata2026pl} for Polish and
SEB~\citep{enevoldsen2024scandinavian} for Nordic languages, which
reveal performance differences not captured by general benchmarks.
 
\paragraph{Evaluation stability}
Benchmark-based evaluation has notable limitations.
Performance can be sensitive to input
variation~\citep{frank2026pteb}, and aggregating heterogeneous metrics
often obscures meaningful differences.
Standard leaderboards rely on simple averaging, ignoring dataset
correlations and imbalance.
Alternative ranking methods based on social choice
theory~\citep{colombo2022best,rofin2023vote} offer a more principled
approach, but still assume equal dataset importance and overlook
correlation between datasets.
Our prior work~\citep{gjorgjevikj2026robustness} analyzed how model
rankings vary across different ranking schemes and decorrelated datasets
compositions within each task, and introduced the top-$k$ consistency to
identify models that remain stable among the top performers under different tasks. These components form the stability backbone of the present
framework.
 
\paragraph{Low-resource languages and evaluation gaps}
Low-resource languages remain underrepresented in
NLP~\citep{joshi2020state}, a challenge also evident for Slavic
languages~\citep{piskorski2025proceedings,piskorski2023proceedings}.
Although resources~\citep{klemen2024si,knez2025semi,piskorski2024cross,
suppa2025sklep,koeva2012bulgarian} and efforts like Slavic NLP
Workshops exist~\citep{piskorski2025proceedings,
piskorski2023proceedings}, coverage is limited to specific languages
or tasks, and languages spoken by smaller populations remain
particularly underrepresented~\citep{piskorski2025proceedings}.
 
\paragraph{Summary}
While prior work has advanced both multilingual embeddings and
benchmark-based evaluation, two gaps remain unaddressed.
First, existing stability analyses treat all task-language pairs as
equally evaluable, ignoring that for many low-resource settings the
available evidence is too sparse or too redundant to support meaningful
stability conclusions.
Second, no prior work jointly analyzes task-specific ranking stability
and cross-task consistency while explicitly qualifying the
reliability of the underlying evaluation evidence.
This paper addresses both gaps through a structured, language-aware
framework that pairs stability analysis with evidence strength
quantification across two evaluation scopes.

\section{Methodology}

This paper proposes a two-dimensional framework for analyzing
multilingual embedding benchmark results when datasets are scarce or
their distribution across tasks is uneven.
The framework distinguishes between two evaluation scopes:
(i)~\textit{task-specific analysis}, which evaluates ranking stability
within a fixed language and task, and
(ii)~\textit{cross-task analysis}, which evaluates model rank
consistency across tasks within a language.
Within each scope, three complementary aspects are analyzed jointly:
(i)~\textit{ranking stability}, which evaluates whether benchmark
rankings remain stable under different aggregation methods
(\textit{aggregation stability}, $W_{RS}$) and dataset compositions
(\textit{composition stability}, $W_{DS}$);
(ii)~\textit{top-$k$ consistency}, which evaluates whether
individual models consistently remain among the top-performing models
across evaluation conditions; and
(iii)~\textit{evidence strength}, which quantifies how reliable the
available benchmark evidence is and, thus, how much confidence can be
placed in the stability conclusions.
A summary is given in Table~\ref{tab:framework}.
 
\textit{Ranking stability} is not evaluated in the cross-task setting
because model scores originate from different tasks (e.g., retrieval,
clustering, classification), each reflecting different objectives and
evaluation metrics.
Consequently, disagreement between task-level model scores does not
necessarily indicate instability but may instead reflect task
specialization.

\paragraph{Relationship to prior work}
The ranking stability component is adopted directly from our prior
work~\citep{gjorgjevikj2026robustness}, which analyzed how model
rankings vary across different ranking schemes and decorrelated dataset
compositions.
The top-$k$ consistency component is a quantitative extension
of that work: whereas prior work used binary top-$k$ membership to
identify consistently strong models, we replace it with a
rank-weighted formulation that additionally credits models according
to their actual position within the top-$k$ set, assigning
progressively lower weight to ranks further from first place.
\textbf{The novel contribution of this paper} is the evidence strength
component, introduced specifically to address the low-resource
multilingual setting.
We define a set of qualitative \textit{evidence levels} that
characterize which types of stability analysis are feasible for each
task--language pair, and a quantitative \textit{Evidence Strength
Score~(ESS)} that measures how reliable the available benchmark
evidence actually is.
ESS combines dataset availability, effective dataset diversity
(accounting for redundancy between correlated datasets), and stability
assessability, i.e., whether aggregation stability, composition
stability, or neither can be meaningfully evaluated.
This separation between stability and evidence strength allows users
to distinguish genuine model stability from apparent stability arising
from scarse or redundant benchmark resources.
The three aspects must always be interpreted jointly: high stability
under high ESS indicates reliable conclusions, whereas high stability
under low ESS suggests that apparent stability may primarily reflect
limited benchmark diversity.
 
\begin{table}[t]
\centering
\small
\caption{Overview of the proposed two-dimensional framework for
analyzing multilingual embedding benchmarks under dataset scarcity
and imbalance.}
\label{tab:framework}
\begin{tabular}{lccc}
\hline
Scope / Aspect & Evidence Strength & Ranking Stability &
Top-$k$ Consistency \\
\hline
Task-specific &
$ESS(t,l)$ &
Kendall's $W$ ($W_{RS}$, $W_{DS}$) &
Task-specific consistency \\
Cross-task &
$ESS(l)$ &
-- &
Cross-task consistency \\
\hline
\end{tabular}
\end{table}
 
\subsection{Problem Setup}
 
Let $\mathcal{L}=\{1,\dots,L\}$ denote the set of analyzed languages
and $\mathcal{T}=\{1,\dots,T\}$ the set of analyzed tasks.
For each language $l \in \mathcal{L}$, let
$\mathcal{M}_l=\{1,\dots,m_l\}$ denotes the set of evaluated models (i.e., the number of evaluated models across different tasks within the same language can differ).
For each task--language pair $(t,l)$, let
$\mathcal{D}_{t,l}=\{1,\dots,n_{t,l}\}$ denote the set of available
datasets.
The evaluation results are represented by a performance matrix
$X^{(t,l)} \in \mathbb{R}^{m_l \times n_{t,l}}$, where
$x_{i,j}^{(t,l)}$ denotes the performance of model $i$ on dataset $j$
according to a task-specific metric.
 
\paragraph{Ranking schemes}
We compare models using a set of ranking schemes
$\mathcal{S}=\{1,\dots,K\}$.
Each ranking scheme $s \in \mathcal{S}$ defines a ranking operator
$R^{(s)}: \mathbb{R}^{m_l \times n} \rightarrow \{1,\dots,m_l\}^{m_l},$
which maps a performance matrix to a ranking vector.
For task $t$, language $l$, and ranking scheme $s$, model rankings are
computed as
$r^{(s,t,l)} = R^{(s)}\!\left(X^{(t,l)}\right),$
where lower ranks indicate better performance.
 
\paragraph{Dataset composition}
To reduce redundancy between highly similar datasets, we compare
datasets by their model-performance profiles.
For dataset $j \in \mathcal{D}_{t,l}$, we define the performance profile
$p_j^{(t,l)} = X_{\cdot,j}^{(t,l)} \in \mathbb{R}^{m_l}$.
For two datasets $j,j' \in \mathcal{D}_{t,l}$, similarity is measured
using the Pearson correlation $\rho_{j,j'}^{(t,l)}$.
Given a threshold $\tau \in [0,1]$, datasets are considered redundant
whenever $\rho_{j,j'}^{(t,l)} > \tau$.
 
Let $\mathcal{C}_{t,l}=\{C_1^{(t,l)},\dots,C_{q_{t,l}}^{(t,l)}\}$
denote the resulting set of groups of correlated dataset, where
$q_{t,l}=|\mathcal{C}_{t,l}|$ is the number of such groups.
 The composition of datasets $u$ is constructed by selecting a
representative dataset from each group
  $D^{(t,l,u)}
  =
  \bigl\{d_1^{(t,l,u)},\dots,d_{q_{t,l}}^{(t,l,u)}\bigr\},
  \quad
  d_a^{(t,l,u)} \in C_a^{(t,l)}.$
Therefore, $\mathcal{U}_{t,l}=\{1,\dots,U_{t,l}\}$ denotes the set of
generated dataset compositions, each defining a reduced performance
matrix $X^{(t,l,u)} = X_{\cdot,D^{(t,l,u)}}^{(t,l)}$.
 
\subsection{Evidence Levels and Evidence Strength Score}
 
The \textit{evidence strength} component addresses a fundamental limitation of existing multilingual
evaluation: benchmark conclusions may appear stable simply because very
limited evaluation evidence is available, not because model ranks are
genuinely stable.
We introduce two complementary tools to characterize evidence
reliability, i.e., a qualitative categorization via \textit{evidence levels}
and a quantitative measure via the \textit{ESS}.
 
\paragraph{Evidence levels}
Since dataset availability and effective dataset diversity differ
substantially across languages and tasks, each task--language pair is
assigned to one of five evidence levels defined in
Table~\ref{tab:evidence_levels}, based on the number of available
datasets $n_{t,l}$ and the number of decorrelated dataset groups
$q_{t,l}$.
 
Pairs in $E_0$ provide no evaluation evidence, while pairs in $E_1$
provide only a single-dataset evidence.
Pairs in $E_{SC}$ contain multiple datasets that form a single
correlated group. Under correlation-aware subsampling, each
composition contains only one dataset from the group, making aggregation across
datasets inapplicable. Although different datasets can be sampled,
they originate from the same correlated group and therefore do
not support meaningful composition stability analysis. Effectively, this setting reduces to only one non-redundant dataset.
Pairs in $E_{RS}$ contain multiple independent datasets, but each forms its own group, allowing only aggregation stability analysis across ranking schemes. Finally, pairs in $E_{RS+DS}$ contain multiple non-trivial dataset groups, enabling full stability analysis across both aggregation methods and dataset compositions.

These levels explicitly distinguish absence of evidence, single-dataset
evidence, redundant multi-dataset evidence, and two degrees of genuine
stability assessability.
This is particularly important in low-resource settings, where
stability conclusions may appear reliable simply because evaluation
coverage is too limited to reveal inconsistencies.
For example, a model may consistently rank first across aggregation
methods when only a single dataset exists, yet such agreement provides
no evidence of robustness and stability.
 
\begin{table}[t]
\centering
\small
\caption{Evidence levels used in the proposed stability framework.}
\label{tab:evidence_levels}
\begin{tabularx}{\linewidth}{llX}
\hline
Evidence Level & Definition & Interpretation \\
\hline
$E_0$ &
$n_{t,l}=0$ &
No evaluation datasets available. \\
$E_1$ &
$n_{t,l}=1$ &
Single-dataset evidence only - stability cannot be assessed. \\
$E_{SC}$ &
$n_{t,l}\geq2 \wedge q_{t,l}=1$ &
Multiple datasets exist but form a single correlated group - the evidence is redundant.
Each decorrelated composition contains only one dataset, so neither aggregation nor composition stability is
meaningfully assessable. \\
$E_{RS}$ &
$n_{t,l}\geq2 \wedge q_{t,l}=n_{t,l}$ &
All datasets are uncorrelated and form separate groups, resulting
in a single dataset composition - only aggregation stability can
be assessed. \\
$E_{RS+DS}$ &
$n_{t,l}\geq2 \wedge 1<q_{t,l}<n_{t,l}$ &
Multiple non-trivial dataset groups exist - both aggregation
stability and composition stability can be assessed. \\
\hline
\end{tabularx}
\end{table}
 
\paragraph{Evidence Strength Score}
Evidence levels determine \textit{which} stability analyses are
applicable.
The ESS quantifies \textit{how much} benchmark evidence supports those
analyses.
It combines four complementary factors:
(i)~dataset availability,
(ii)~effective dataset diversity,
(iii)~the ability to evaluate aggregation stability across ranking
schemes, and
(iv)~the ability to evaluate composition stability across dataset
compositions.
 
\textit{Normalized dataset availability} measures the number of available datasets for a task-language pair relative to a predefined saturation threshold (Eq.~\ref{eq:datasetAvailability}), where $n_{\min}$ denotes the minimum number of datasets considered
sufficient for reliable evaluation evidence. 
Its initialization is later explained in the experimental design setup.

\begin{equation}
\label{eq:datasetAvailability}
    A(t,l) = \min\!\left(\frac{n_{t,l}}{n_{\min}},\,1\right)
\end{equation}

\textit{Effective dataset diversity} accounts for redundancy between
datasets, defined as the ratio of uncorrelated datasets (Eq.~\ref{eq:datasetDiversity}). Values close to 1 indicate that datasets provide largely independent
evaluation evidence, while lower values indicate substantial redundancy.

\begin{equation}
\label{eq:datasetDiversity}
  \mathrm{Div}(t,l) = \frac{q_{t,l}}{n_{t,l}}
\end{equation}

\textit{Aggregation stability availability} (Eq.~\ref{eq:aggregationStability}) and \textit{composition stability availability} (Eq.~\ref{eq:compositionStability}) are indicator
functions that capture whether each type of stability analysis is meaningfully assessable under the available dataset structure. For $E_{SC}$, both indicators are zero because
each decorrelated composition contains only one
dataset, and all alternative compositions originate
from the same correlated group.

\begin{equation}
\label{eq:aggregationStability}
  RS(t,l) = \mathbf{1}\!\left[(t,l)\in E_{RS}\cup E_{RS+DS}\right]
\end{equation}

\begin{equation}
\label{eq:compositionStability}
  DS(t,l) = \mathbf{1}\!\left[(t,l)\in E_{RS+DS}\right]
\end{equation}

The task-level ESS combines the   four above-defined factors into one score that characterizes the adequacy of the available benchmark evidence in terms of dataset availability, dataset diversity, and stability assessability (see Eq.~\ref{eq:ess}). The four factors are combined with equal weights since equal weighting preserves interpretability and avoids introducing
additional assumptions about the relative importance of dataset
quantity, diversity, and stability assessability. For applications in which certain factors are more important than others, practitioners can customize the score by assigning greater weights to those factors. However, from an academic and research perspective, we consider all factors equally important for assessing the evidence supporting the claims made on the basis of a benchmark/leaderboard.
High $ESS(t,l)$ indicates that stability conclusions are supported by
sufficiently large, diverse, and non-redundant evaluation evidence, while 
low values signal that conclusions should be interpreted cautiously. We emphasize that the score should not be interpreted as a measure of statistical confidence in model performance or ranking outcomes.
Further discussion of design choices is provided in~\ref{appendix:ess} and details on the qualitative interpretation of the ESS score in ~\ref{appendix:interpretationESS}. Additional sensitivity analysis of the weighting is done in~\ref{appendix:ess_wighting}.

\begin{equation}
\label{eq:ess}
  ESS(t,l)
  =
  \frac{1}{4}
  \bigl(
    A(t,l)
    +
    \mathrm{Div}(t,l)
    +
    RS(t,l)
    +
    DS(t,l)
  \bigr)
\end{equation}

\subsection{Task-Specific Analysis Within a Language}
 
The task-specific analysis evaluates ranking stability and top-$k$ consistency within each task-language pair $(t,l)$. The previously presented evidence levels determine which analyses are applicable within task-language pairs, and ESS quantifies the evidence that supports the results.
 
\paragraph{Ranking stability}
Ranking stability evaluates whether benchmark conclusions remain
stable under different aggregation methods and dataset compositions,
measured separately as \textit{aggregation stability}~($W_{RS}$) and
\textit{composition stability}~($W_{DS}$), respectively.
To calculate the stability, we use Kendall's coefficient of concordance~($W$) (i.e., larger values
indicate stronger agreement among rankings). Let $\mathcal{U}_{t,l}^{rob} = \{u_1,\ldots,u_{U_{t,l}}\}$ denote the
set of dataset compositions generated through
correlation-aware subsampling. The aggregation stability ($W_{RS}$) is computed for each dataset composition across all ranking schemes using Eq.~\ref{eq:wrs_ds}, and
composition stability ($W_{DS}$) is computed for each ranking scheme $s$ across all dataset compositions using Eq.~\ref{eq:wds}.

\begin{equation}
\label{eq:wrs_ds}
  W_{RS}^{(t,l,u)}
  =
  \frac{
    12
    \sum_{i=1}^{m_l}
    \!\left(
      R_i^{(t,l,u)}
      -
      \frac{K(m_l+1)}{2}
    \right)^{\!2}
  }{
    K^2(m_l^3-m_l)
  },
  \qquad
  R_i^{(t,l,u)}
  =
  \sum_{s=1}^{K}
  r_i^{(s,t,l,u)}.
\end{equation}
 
\begin{equation}
\label{eq:wds}
  W_{DS}^{(s,t,l)}
  =
  \frac{
    12
    \sum_{i=1}^{m_l}
    \!\left(
      Q_i^{(s,t,l)}
      -
      \frac{U_{t,l}(m_l+1)}{2}
    \right)^{\!2}
  }{
    U_{t,l}^2(m_l^3-m_l)
  },
  \qquad
  Q_i^{(s,t,l)}
  =
  \sum_{u\in\mathcal{U}_{t,l}^{rob}}
  r_i^{(s,t,l,u)}.
\end{equation}

We clarify that Eqs.~\ref{eq:wrs_ds} and~\ref{eq:wds} correspond to the $E_{RS+DS}$ scenario. In the $E_{RS}$ scenario, where only a single dataset composition is available ($|\mathcal{U}_{t,l}^{rob}| = 1$), $W_{RS}^{(t,l,u)} = W_{RS}^{(t,l)}$ and is computed only once, whereas $W_{DS}^{(s,t,l)}$ cannot be computed. 
High values of $W_{RS}$ indicate high aggregation stability, i.e.,
rankings remain consistent across different aggregation methods.
High values of $W_{DS}$ indicate high composition stability, i.e.,
model rankings remain largely unchanged under dataset-composition
perturbations generated through correlation-aware subsampling.
Together, these two measures quantify the extent to which benchmark
conclusions depend on the choice of aggregation method and the
composition of the evaluation benchmark.

\paragraph{Top-$k$ consistency}
Ranking stability evaluates agreement between complete rankings, but
does not identify which individual models consistently remain among the strongest performers. We therefore introduce top-$k$ consistency, which quantifies how reliably a model achieves high ranks across evaluation conditions. Unlike binary top-$k$ membership as in our previous study, the proposed measure additionally accounts for the actual position of a model within the top-$k$ set, assigning higher credit to models that consistently appear near the top of the ranking. Eq.~\ref{eq:rank_weight} defines the top-$k$ rank weight for a model at ranking position $r_i$. A model ranked first receives weight~1, while a model ranked exactly
at position~$k$ receives weight~$1/k$.
Models ranked outside the top-$k$ receive no credit.

\begin{equation}
\label{eq:rank_weight}
  w_k(r_i)=
  \begin{cases}
    \dfrac{k-r_i+1}{k}, & r_i \leq k,\\[4pt]
    0, & r_i > k.
  \end{cases}
\end{equation}

The top-$k$ consistency of each model, $i$, is computed as the mean of its top-$k$ rank weights across all combinations of the $K$ ranking schemes and $U_{t,l}$ dataset compositions (Eq.~\ref{eq:wtc_ds}).

\begin{equation}
\label{eq:wtc_ds}
  TC_{RS+DS}^{(t,l)}(i)
  =
  \frac{1}{KU_{t,l}}
  \sum_{s=1}^{K}
  \sum_{u=1}^{U_{t,l}}
  w_k\!\left(r_i^{(s,t,l,u)}\right).
\end{equation}

In cases when a single dataset composition is available, as in the case of $E_{RS}$ scenario, the task-specific consistency of each model, $i$, is computed as the mean of its top-$k$ rank weights only across the ranking schemes (Eq.~\ref{eq:wtc_rs}). The $E_1$ and $E_{SC}$ scenarios are special cases in which the single dataset composition consists of a single dataset, and all ranking schemes yield the same effective ranking, so the $TC(i)$ score is equal to the $w_k(r_i)$ calculated on the ranks on the single available/sampled dataset.

\begin{equation}
\label{eq:wtc_rs}
  TC_{RS}^{(t,l)}(i)
  =
  \frac{1}{K}
  \sum_{s=1}^{K}
  w_k\!\left(r_i^{(s,t,l)}\right).
\end{equation}

 Values of the top-$k$ consistency close to one indicate that a model consistently occupies the highest-ranking positions across evaluation conditions, whereas values near zero indicate that the model rarely appears among the top-performing methods.

\subsection{Cross-Task Analysis Within a Language}
 
The cross-task analysis evaluates the consistency with which
models achieve high ranks across the different evaluated task families
within a language. As in the task-specific analysis, evidence levels and ESS are
essential for qualifying the conclusions.
 
\paragraph{Evidence strength}
The language-level ESS is a generalization of the task-language $ESS(t,l)$ score. It is computed as the mean of all task-language $ESS(t,l)$ scores within the language (Eq~\ref{eq:ess_l}). The calculation of the language-level ESS also includes tasks that fall under $E_0$ or for which no dataset is available. In these cases, we assign $ESS(t,l)=0$, so missing benchmark coverage directly lowers the language-level score. The resulting score therefore reflects both the quality of the available evidence and the availability of benchmarks across tasks.

\begin{equation}
\label{eq:ess_l}
    ESS(l) = \frac{1}{|\mathcal{T}|} \sum_{t\in\mathcal{T}} ESS(t,l)
\end{equation}

\paragraph{Cross-task consistency} 
The \textit{cross-task consistency} measures how consistently a model achieves high ranks across the tasks available for a language, while accounting for how many tasks can be evaluated in that language. We compute it in four steps.

\paragraph{Step 1: Compute the contribution of each task}
For each evaluable task, we score a model's top-$k$ ranking performance using the strongest evidence available for that task-language pair:

\begin{itemize}
  \item For pairs in $E_{RS+DS}$, we average the top-$k$ weights across the robust dataset compositions.
  \item For pairs in $E_1$ or $E_{SC}$, we use the only possible ranking, which is based on a single available/representative dataset.
  \item For pairs in $E_{RS}$, we use the ranking from the original \texttt{all} composition.
\end{itemize}

\noindent If a model has no evaluation for a task, its contribution for that task is zero. Formally, the contribution of model $i$ under ranking scheme $s$ is:

\begin{equation}
\label{eq:taskLevelContribution}
\widetilde{C}^{(s,t,l)}(i)
=
\begin{cases}
0,
& i \text{ is not evaluated on } (t,l),\\[4pt]
\dfrac{1}{U_{t,l}}
\displaystyle\sum_{u=1}^{U_{t,l}}
w_k\!\left(r_i^{(s,t,l,u)}\right),
& (t,l)\in E_{RS+DS},\\[8pt]
w_k\!\left(r_i^{(t,l)}\right),
& (t,l)\in E_1\cup E_{SC},\\[6pt]
w_k\!\left(r_i^{(s,t,l,\mathrm{all})}\right),
& (t,l)\in E_{RS}
\end{cases}
\end{equation}

\paragraph{Step 2: Average across tasks}
For each language $l$, the evaluable task set contains the tasks with at least one available evaluation, $\mathcal{T}^{eval}_l = \{t \in \mathcal{T}: n_{t,l}>0\}$. We average the task contributions over this set (Eq.~\ref{eq:cross_task_adaptive}). Missing model evaluations remain in this average as zeros.

\begin{equation}
\label{eq:cross_task_adaptive}
  C^{(s,l)}(i)
  =
  \frac{1}{|\mathcal{T}^{eval}_l|}
  \sum_{t\in\mathcal{T}^{eval}_l}
  \widetilde{C}^{(s,t,l)}(i)
\end{equation}

\paragraph{Step 3: Average across ranking schemes}
We then average the result over the $K$ ranking schemes (Eq.~\ref{eq:cross_task_average}).

\begin{equation}
\label{eq:cross_task_average}
  C^{(l)}(i)
  =
  \frac{1}{K}
  \sum_{s=1}^{K}
  C^{(s,l)}(i)
\end{equation}

\paragraph{Step 4: Adjust for language-level task coverage}
Task coverage is the share of all tasks that can be evaluated for language $l$:

\begin{equation}
\label{eq:task_coverage}
  \mathrm{Coverage}(l)
  =
  \frac{|\mathcal{T}^{eval}_l|}{|\mathcal{T}|}
\end{equation}

\noindent Multiplying the consistency score by this coverage gives the final score (Eq.~\ref{eq:cross_task_consistency}).

\begin{equation}
\label{eq:cross_task_consistency}
  CW^{(l)}(i)
  =
  \mathrm{Coverage}(l)
  \cdot
  C^{(l)}(i)
\end{equation}

\noindent This is equivalent to averaging over all tasks, with non-evaluable tasks contributing zero. As a result, the score rewards consistently high ranks across many tasks and penalizes conclusions drawn from languages with limited task coverage.

\paragraph{Differences in candidate sets}
The set of eligible models may differ across tasks within a language because of task-specific evaluation availability and zero-shot filtering. Each rank is therefore relative to the models available for that task, and the number of competing models may vary. Because the top-$k$ weights depend on absolute rank positions, they do not adjust for candidate-set size. The score therefore summarizes relative ranking performance across task-specific candidate sets, rather than comparing models against a fixed set of competitors.

\section{Experimental Design}
 
The experimental data used in this study are obtained from a snapshot of the MTEB Multilingual leaderboard v2\footnote{https://huggingface.co/spaces/mteb/leaderboard,
accessed on 26-12-2025}.
Our analysis includes the performance of embedding models across eight
tasks\footnote{We use \textit{task} in this paper to refer to
\textit{task category} in MTEB terminology, while \textit{dataset}
refers to \textit{task} in MTEB.}, including classification,
clustering, retrieval, reranking, semantic textual similarity~(STS),
pair classification, multilabel classification, and bitext mining.
Within each task, we restrict our analysis to open, fully zero-shot
models, i.e., models not trained on any task-specific benchmark data. The number of models by task, included in the analysis after the zero-shot filtering, is given in Table~\ref{table:models_per_task}.
Since each MTEB dataset is associated with one or multiple languages,
we construct language-specific, task-specific subsets of datasets for
each Slavic language.
We exclude languages with a single dataset in total across all tasks.

\begin{table}[htbp]
\centering
\small
\caption{Number of models included in the study per task, after the zero-shot filtering.}
\label{table:models_per_task}
\begin{tabular}{lccccccccc}
\hline
\textbf{Task} 
& Clss. 
& Clst.
& Pair Clss. 
& Rerank. 
& Retriev.
& STS 
& Mult. Clss. 
& Instr. Rerank.
& Bitext Min. 
\\
\hline
\textbf{Num.}
& 89 & 89 & 94 & 91 & 52 & 88 & 103 & 109 & 96 \\
\hline
\end{tabular}
\end{table}

We set the minimum number of datasets considered sufficient for reliable evaluation evidence to five ($n_{min}=5$). This threshold ensures that, in the ideal case in which the datasets form three mutually decorrelated groups, at least three different dataset compositions can be constructed by selecting one dataset per group, allowing $E_{RS+DS}$ to be performed. Five is the smallest number of datasets that satisfies this requirement: three datasets yield only one composition ($1 \times 1 \times 1$) and four yield two ($2 \times 1 \times 1$), whereas five yield three ($3 \times 1 \times 1$) or four ($2 \times 2 \times 1$), depending on how they are distributed across the groups.  To assess the sensitivity of the analysis to this choice, we also varied $n_{min}$ from 5 to 10 in steps of 1 and examined the stability of the task-language-specific and cross-task language scores across them (see~\ref{appendix:ess_saturation}).
 
The datasets were decorrelated using the Pearson correlation coefficient with a threshold of $\tau=0.9$. Datasets exceeding this correlation are grouped together, and we
construct decorrelated subsets by randomly sampling one dataset per
group. In addition, a sensitivity analysis has been performed using $\tau=0.7$ and $\tau=0.8$ in~\ref{appendix:ess_saturation}. These thresholds were selected based on the interpretation of strong (0.70, 0.90) and very strong [0.9,) correlations provided by~\citep{schober2018correlation}. 
This procedure is repeated three times to account for sampling
variability, resulting in one dataset composition including all
datasets and three decorrelated dataset compositions.
If no highly correlated datasets exist, only the one composition
including all datasets is generated.
 
To rank models for each language, task, and dataset composition, we
apply a diverse set of multi-criteria decision-making~(MCDM) methods:
Weighted Sum Model~(WSM), TOPSIS~\citep{chen1992fuzzy},
VIKOR~\citep{opricovic1998multicriteria}, and
PROMETHEE~II~\citep{brans1982ingenierie}.
For PROMETHEE~II, we use the usual and Gaussian preference functions.
The selected methods represent the main methodological families,
ensuring conceptual diversity and complementary decision logic.
They cover value-based aggregation~(WSM), distance-based
ranking~(TOPSIS), compromise-based optimization~(VIKOR), and
outranking approaches~(PROMETHEE~II), providing a representative and
non-redundant portfolio.
To further capture ranking variability, we combine these methods with
three objective weighting strategies: equal weighting,
MEREC~\citep{keshavarz2021determination}, and
CRITIC~\citep{diakoulaki1995determining}, which assign weights to the
datasets using the performance matrix.
This results in 15 ranking schemes (5 MCDM methods $\times$ 3
weighting strategies), where datasets act as criteria in each
language-task setting.

The weights of the four components constituting the EES score, i.e., dataset availability, dataset diversity, ranking scheme robustness assessability, and dataset composition robustness assessability, were set to equal value of 0.25. We varied the weights of the four ESS components individually, redistributing the remaining weight equally among the other components, to assess the equal-weight formulation (see~\ref{appendix:ess_wighting}).

The value of the parameter $k$ in the task-specific top-$k$ consistency was set to value of 10, but we further evaluated the changes in the consistency score for $k \in \{5,8,10,12,15,20\}$, using $k=10$ as the reference and keeping the evaluated model set fixed across configurations (see~\ref{appendix:topk_sensitivity}).

\section{Results}
 
\subsection{Language-Specific Task Coverage}
 
\begin{figure*}[!ht]
\centering
  \includegraphics[width=\linewidth]{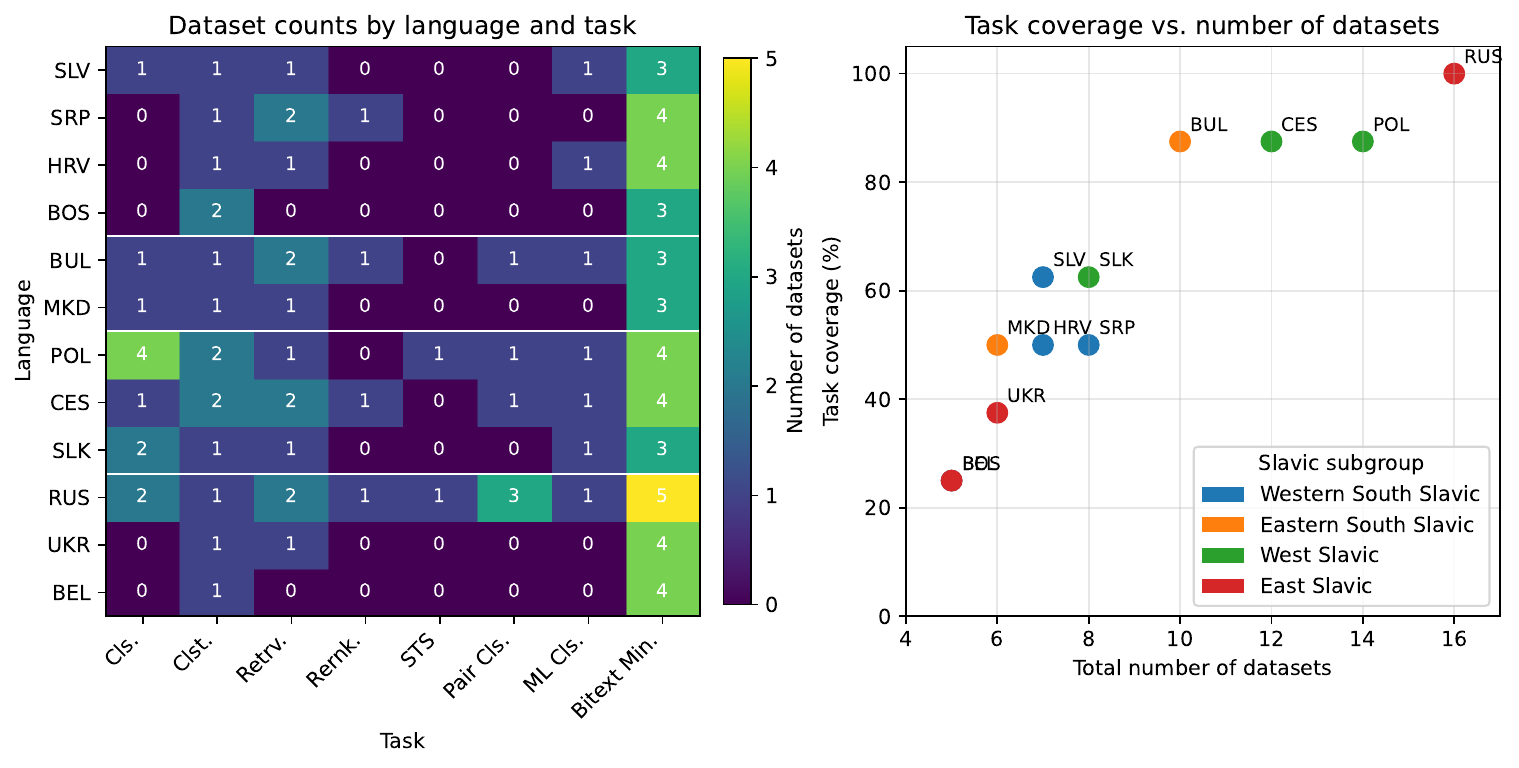}
  \caption{Dataset distribution across tasks and languages~(left).
  Task coverage (percentage of tasks with at least one dataset) against
  the total number of datasets per language, showing substantial
  imbalance across languages~(right).}
  \label{fig:dataset_distribution}
\end{figure*}

The analysis of embedding model rank consistency in different languages
critically depends on the availability and balance of datasets across
tasks.
Figure~\ref{fig:dataset_distribution}~(left) summarizes the available
dataset distribution per language, revealing a strong imbalance.
For most languages, a large proportion of datasets is concentrated in
the bitext mining task (e.g., 3--5 datasets per language), while
several other tasks are either underrepresented or entirely lacking
datasets.
Even in languages with relatively high task coverage, such as
Polish~(POL), Czech~(CES), and Bulgarian~(BUL) (87.5\% task
coverage), many tasks are supported by a single dataset.
Notable exceptions include classification in Polish (4 datasets) and
pair classification in Russian~(RUS) (3 datasets), which provide
richer task-specific coverage.
 
Figure~\ref{fig:dataset_distribution}~(right) shows that task coverage
varies substantially across languages.
Russian achieves full coverage (100\%) across all tasks, while
Bulgarian, Polish, and Czech cover 7 out of 8 tasks~(87.5\%).
Slovenian~(SLV) and Slovak~(SLK) reach moderate coverage~(62.5\%),
whereas Serbian~(SRP), Croatian~(HRV), and Macedonian~(MKD) cover
only half of the tasks~(50\%).
The lowest coverage is observed for Ukrainian~(UKR) (37.5\%),
Bosnian~(BOS), and Belarusian~(BEL) (both 25.0\%), with only 2--3
tasks represented.
The figure clearly separates high-resource Slavic languages~(Russian,
Bulgarian, Polish, and Czech), mid-resource~(Slovenian, Slovak,
Serbian, Croatian, and Macedonian), and low-resource~(Ukrainian,
Bosnian, and Belarusian) in terms of dataset availability.
This uneven distribution has direct implications for our stability
analysis. Namely, languages with broader task coverage enable task-specific
and cross-task analysis, while those with limited coverage restrict
the analysis to a small subset of tasks and reduce the reliability of
cross-task consistency estimates.
 
\begin{figure*}[!ht]
\centering
  \includegraphics[width=0.8\linewidth]{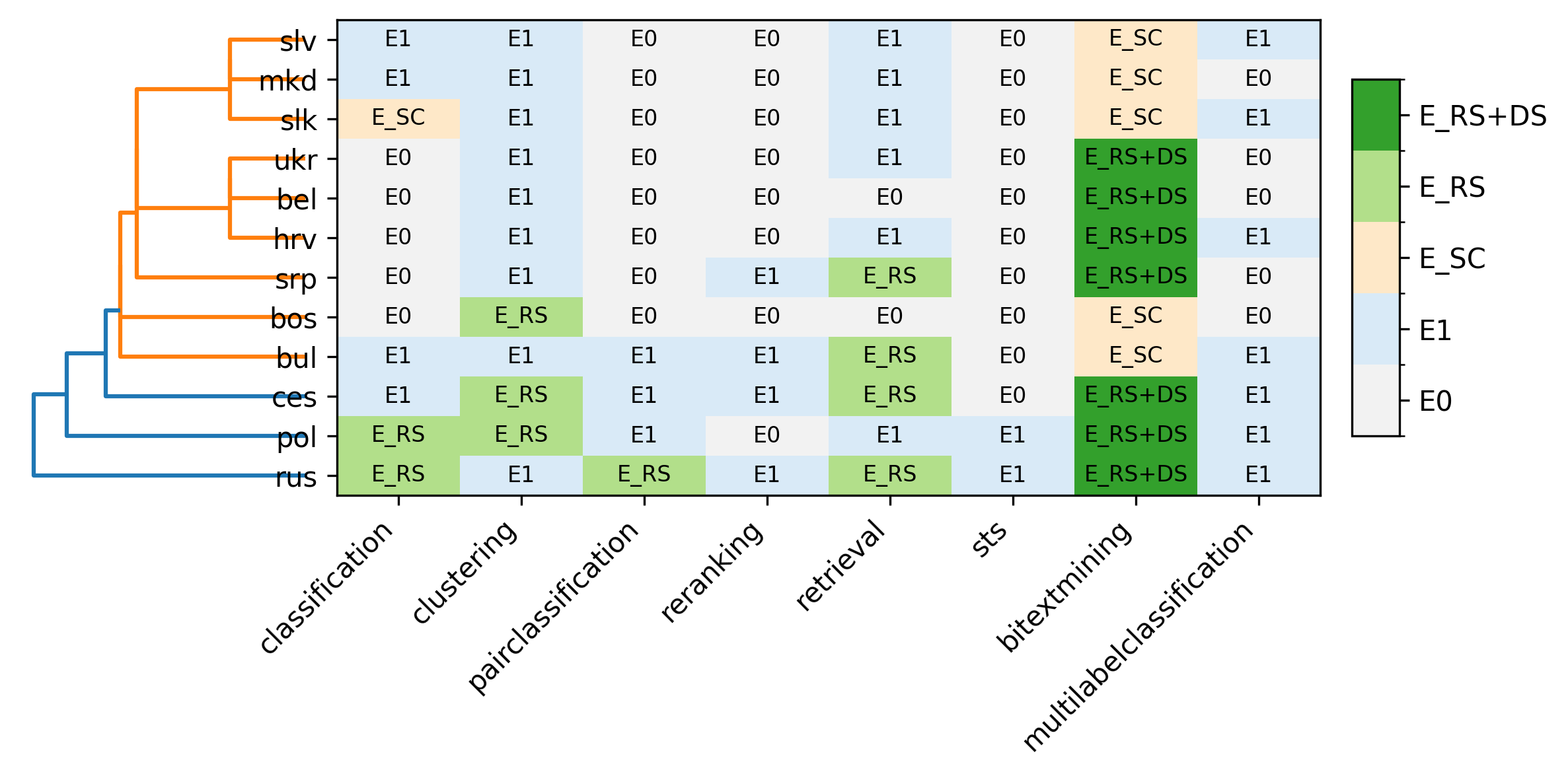}
  \caption{Evidence levels by language and task. Each cell shows the
  evidence category ($E_0$--$E_{RS+DS}$) for the corresponding
  language-task pair, reflecting dataset availability, redundancy,
  and stability assessability.}
  \label{fig:evidence_levels_heatmap}
\end{figure*}

Figure~\ref{fig:evidence_levels_heatmap} summarizes the evidence
levels across languages and tasks, revealing substantial disparities
in benchmark coverage, dataset diversity, and stability assessability.
Several tasks, particularly \textit{STS}, remain severely
underrepresented, with nearly all languages in the no-evidence
regime~($E_0$).
Similarly, \textit{reranking} and \textit{pair classification} are
dominated by sparse evidence levels, most frequently single-dataset
evidence~($E_1$) or complete absence of evaluation resources.
In contrast, \textit{bitext mining} consistently exhibits the
strongest evidence levels across most languages, frequently reaching
the full stability setting~($E_{RS+DS}$), indicating that both
ranking scheme stability and dataset composition stability can be meaningfully
assessed.

The figure additionally highlights strong language-level disparities.
Russian, Polish, and Czech form a cluster characterized by broader
benchmark coverage and more robust evidence levels across tasks,
including multiple $E_{RS}$ and $E_{RS+DS}$ cases.
Bulgarian also exhibits relatively strong retrieval evidence despite
otherwise sparse task coverage.
In contrast, several South Slavic languages, including Bosnian,
Slovak, Macedonian, and Slovene, remain dominated by $E_0$ and $E_1$
cases, indicating highly limited benchmark support.
The presence of multiple $E_{SC}$ cases, particularly for
\textit{classification} and \textit{bitext mining}, further
demonstrates that multiple available datasets do not necessarily
correspond to strong evaluation evidence, as many datasets remain
highly correlated and provide effectively redundant information.
Overall, the evidence-level structure confirms that benchmark sparsity
and redundancy remain major obstacles for reliable multilingual
stability analysis in lower-coverage Slavic languages.
 
\subsection{Language-Specific Task-Specific Stability}
 
\subsubsection{Evidence Strength}
 
\begin{figure*}[!ht]
\centering
  \includegraphics[width=0.7\linewidth]{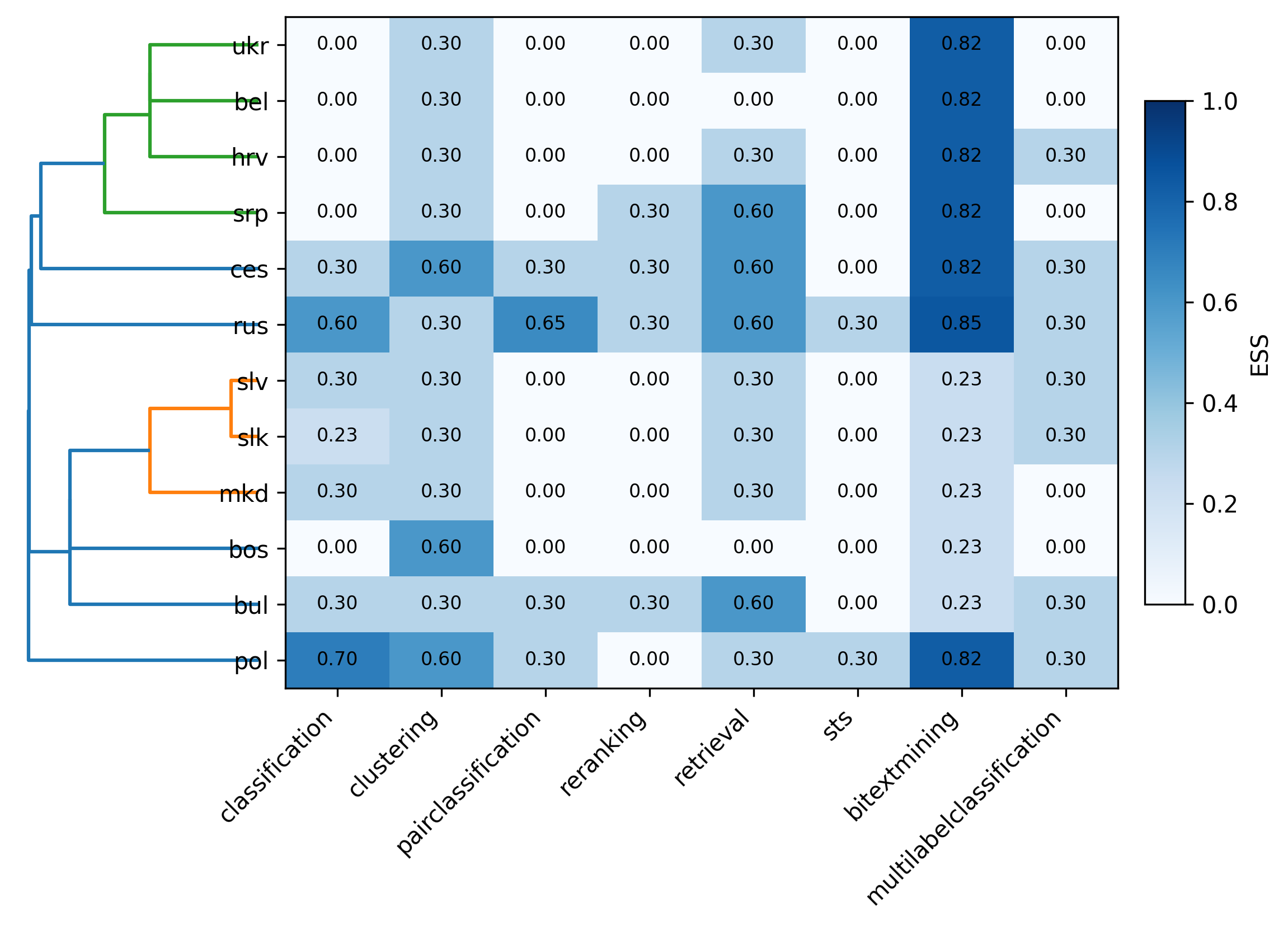}
  \caption{Task-specific ESS by language
  and task. Higher values indicate stronger, more diverse, and
  stability-assessable benchmark evidence.}
  \label{fig:evidence_strength_heatmap}
\end{figure*}

Figure~\ref{fig:evidence_strength_heatmap} presents the task-specific
ESS across Slavic languages and tasks, revealing substantial
disparities in benchmark coverage, dataset diversity, and stability
assessability.
Overall, the strongest evidence is consistently observed for
\textit{bitext mining}, where half of the languages achieve ESS values above
$0.8$, indicating relatively rich, diverse, and
stability-assessable evaluation evidence.
Russian achieves the highest ESS in this task~($0.85$), followed
closely by Polish, Czech, Serbian, Croatian, Belarusian, and Ukrainian
($0.82$).
In contrast, Bulgarian, Macedonian, Slovak, Slovene, and Bosnian
exhibit substantially weaker evidence due to correlated datasets forming a single cluster.
 
Moderate evidence strength is observed for \textit{classification},
\textit{clustering}, and \textit{retrieval}, although substantial
language-level variation remains present.
Russian and Polish consistently exhibit the strongest overall evidence
profiles across multiple tasks, while Czech also demonstrates
comparatively strong evidence, particularly for \textit{clustering}
and \textit{retrieval}.
In contrast, Bosnian, Slovene, Macedonian, and Slovak remain dominated
by weak evidence levels across most tasks.
Several tasks remain severely underrepresented across nearly all
languages: \textit{STS} exhibits near-zero ESS for almost all
languages except Russian and Polish, while \textit{reranking},
\textit{pair classification}, and \textit{multilabel classification}
are largely characterized by missing or single-dataset evidence, resulting in
limited stability assessability.
 
The hierarchical clustering separates languages according to benchmark
coverage and evidence quality.
Russian and Polish form the strongest evidence cluster, characterized
by broader task coverage and more diverse evaluation resources.
Czech, Serbian, and Croatian form an
intermediate cluster with moderate evidence concentrated primarily in
\textit{bitext mining} and \textit{retrieval}.
Bosnian, Slovak, Slovene, and Macedonian cluster together as
the weakest evidence group, reflecting low evidence
across most evaluated tasks.
 
Importantly, the ESS analysis provides critical context for
interpreting both ranking stability and top-$k$ consistency
results.
High ranking scheme or dataset composition stability should not automatically be
interpreted as evidence of genuinely stable model behavior when ESS
values remain low, since stable rankings may simply reflect scarce or
redundant evaluation settings.
Consequently, stability conclusions for tasks with weak ESS (particularly \textit{STS}, \textit{reranking}, and \textit{pair
classification}) should be interpreted cautiously.
In contrast, the high ESS observed for \textit{bitext mining} and for
languages such as Russian, Polish, and Czech indicates that the
corresponding stability and consistency results are supported
by substantially stronger and more reliable benchmark evidence.
 
To assess the robustness of these evidence strength findings, we conducted a sensitivity analysis of ESS with respect to the dataset correlation threshold $\tau$ and dataset saturation threshold $n_{\min}$ (see~\ref{appendix:ess_saturation}). We evaluated 18 parameter configurations, combining $\tau \in \{0.7, 0.8, 0.9\}$ with $n_{\min} \in \{5, 6, 7, 8, 9, 10\}$, and measured agreement between the resulting ESS-based task--language orderings using Kendall's coefficient of concordance. The results show near-perfect agreement across saturation thresholds ($W > 0.999$) and high agreement across correlation thresholds ($W = 0.972$--$0.977$). Although the correlation threshold has a greater influence on ESS than the saturation threshold, the relative ordering of task-language combinations remains largely preserved. These findings indicate that the task-specific ESS is robust to the examined parameter choices.

The ESS component-weighting sensitivity analysis in~\ref{appendix:ess_wighting} indicates that ESS values are relatively stable under moderate deviations from equal weighting, while larger changes in component importance produce more pronounced differences. Dataset diversity exhibits the greatest sensitivity. Increasing its weight to 0.60 yields a mean absolute difference of 0.1548 from the equal-weight baseline, compared with 0.0493 when the dataset availability weight is increased to the same value. These results support equal weighting as a transparent reference configuration, while emphasizing that ESS value depends on the relative importance assigned to its components.
 
\subsubsection{Ranking Stability}
 
Kendall's coefficient of concordance ($W$) indicates high ranking stability across the task-language pairs for which robustness can be assessed ($E_{RS}$ and $E_{RS+DS}$). On the original (all) dataset compositions, ranking-scheme stability is consistently high, with a mean of $\overline{W}_{RS}^{all}=0.986$ and a median of $\widetilde{W}_{RS}^{all}=0.990$. This indicates that the 15 aggregation schemes generally produce similar model orderings. The highest mean stability is observed for pair classification ($\overline{W}_{RS}^{all}=0.993$), followed by bitext mining ($0.992$), retrieval ($0.988$), clustering ($0.985$), and classification ($0.961$).

For task-language pairs classified as $E_{RS+DS}$, which occur only in bitext mining, stability was additionally assessed using robust dataset compositions obtained through correlation-aware dataset subsampling. Aggregation stability remains high within these compositions, with a mean of $\overline{W}_{RS}^{rob}=0.992$ and values ranging from $0.984$ to $0.996$. This indicates that the aggregation schemes produce highly consistent model orderings across the evaluated robust compositions.

Composition stability was assessed separately by comparing model rankings across the three robust compositions for each aggregation scheme. The resulting mean concordance is $\overline{W}_{DS}^{rob}=0.976$, with a median of $\widetilde{W}_{DS}^{rob}=0.974$. These results indicate that model rankings are also largely preserved when the dataset composition changes. Taken together, the findings suggest that, for the assessable task-language pairs, model rankings are robust to the choice of aggregation scheme and, where multiple robust compositions are available, to correlation-aware dataset selection.

The stability results are further supported by the evidence-strength
analysis.
Task-language pairs belonging to $E_{RS}$ have an average ESS of
$0.615$, while those belonging to $E_{RS+DS}$ achieve an average ESS
of $0.829$, substantially higher than the scores observed for $E_1$
($0.300$) and $E_{SC}$~($0.232$).
Consequently, the cases for which stability can be evaluated are also
those supported by the strongest benchmark evidence.
This is particularly important because high Kendall agreement alone
does not guarantee reliable conclusions, i.e., rankings may appear stable
simply because only a single dataset is available.
 
At the task level, the results reinforce the distinction between
stability and evidence availability.
Tasks such as classification, clustering, retrieval, and pair
classification exhibit high $W_{RS}^{all}$ values, but only for a
small subset of languages belonging to $E_{RS}$.
Therefore, these high agreement values indicate aggregation stability,
but do not provide evidence of composition stability.
In contrast, bitext mining is the only task that supports both
aggregation stability and composition stability analysis.
Its simultaneously high values of $W_{RS}^{all}$, $W_{RS}^{rob}$,
$W_{DS}^{rob}$, and ESS provide the strongest evidence that benchmark
conclusions remain stable under both methodological and
dataset-composition perturbations.
 
Overall, the Kendall's~$W$ results show that, whenever stability can
be evaluated, multilingual embedding rankings are highly concordant.
The choice of aggregation method has limited influence on the
resulting model orderings, and in the case of bitext mining, rankings
also remain stable across robust dataset compositions.
Nevertheless, the primary limitation of current multilingual
evaluation remains benchmark scarcity rather than ranking instability.
The evidence-strength analysis reveals that only a small subset of
task-language pairs provides sufficiently rich evaluation evidence to
support comprehensive stability assessment, highlighting the need for
broader and more diverse benchmark resources. For the full results, see the project repository.
 
\subsubsection{Top-$k$ Consistency}
 
The following tables report task-specific top-$k$ consistency
results for each evaluated task and language.
For each language, we list the model with the highest rank-weighted
top-$k$ consistency score, together with the evidence level,
ESS, dataset statistics, and the number of
ranking instances used to compute the score.
The evidence level indicates which type of stability assessment is
possible, while ESS quantifies the reliability of the available
benchmark evidence.
The columns $n$, $q$, and $U$ denote the number of datasets,
decorrelated dataset clusters, and dataset compositions respectively.
The \#Ranks column reports the number of ranking outputs used to
compute the consistency, and TC denotes the resulting
task-specific top-$k$ consistency score.
The tables should be interpreted by considering both the TC value and
the strength of the underlying evidence: high TC under strong evidence
indicates reliable model stability, whereas high TC under weak
evidence may reflect limited benchmark diversity rather than
genuinely stable model behavior.
 
\paragraph{Classification}
 
Table~\ref{tab:summary_classification} summarizes the models with the
highest task-specific top-$k$ consistency for the classification task,
together with the corresponding evidence levels and ESS values.
Overall, classification remains one of the sparsest evaluation
settings, with substantial disparities in benchmark coverage across
languages.
Most languages are supported by either a single dataset~($E_1$) or no
evaluation evidence at all~($E_0$), resulting in low ESS values
($ESS \leq 0.30$).
Consequently, top-$k$ consistency conclusions for Slovene, Bulgarian,
Macedonian, and Czech are based on only a single dataset and should
therefore be interpreted as identifying the best-performing model on
the single available dataset rather than providing a fully validated
stability assessment.
 
The strongest evidence is observed for Polish~($ESS=0.70$), which is
supported by four independent datasets~($n=q=4$) and enables
aggregation stability analysis~($E_{RS}$).
Russian provides the second strongest evidence
profile~($ESS=0.60$), supported by two independent datasets and
aggregation stability analysis.
These two languages therefore provide the most reliable classification
conclusions.
In contrast, Slovak represents a single-cluster evidence
setting~($E_{SC}$), where multiple datasets are available~($n=2$) but
belong to a single correlated cluster~($q=1$).
Therefore, the resulting ESS remains comparatively weak~($ESS=0.23$) due to limited dataset diversity.
 
The top-$k$ consistency scores further illustrate the
impact of benchmark coverage on stability interpretation.
Perfect consistency~($TC=1.00$) is observed for most languages with
available evidence, including Slovene, Bulgarian, Macedonian, Czech,
Slovak, and Russian.
However, these scores do not imply equally reliable conclusions.
For the $E_1$ languages, perfect consistency simply reflects agreement
within a single ranking instance, whereas the corresponding result for
Russian is supported by fifteen ranking schemes and substantially
stronger benchmark evidence.
Similarly, Polish exhibits a slightly lower consistency
score~($TC=0.99$) but provides stronger evidence overall due to its
broader and more diverse benchmark coverage.
 
Across languages, \textit{Qwen3-Embedding} variants emerge as the
dominant models, appearing as the most consistent model in
Slovene, Czech, Slovak, and Russian.
In contrast, \textit{Linq-Embed-Mistral} achieves the strongest
consistency in Bulgarian, \textit{Qwen3-Embedding-4B} in Macedonian,
and \textit{llama-embed-nemotron-8b} in Polish.
Overall, the classification results highlight that apparent consistency
is often constrained by limited benchmark evidence, and that the most
reliable conclusions are currently restricted to Polish and Russian,
where stability assessments are supported by sufficiently diverse and
non-redundant evaluation datasets.
 
\begin{table}[!ht]
  \caption{Models with highest task-specific top-$k$ consistency (TC) by language on the classification task.}
  \label{tab:summary_classification}
  \begin{center}
    \begin{small}
      \begin{tabularx}{\columnwidth}{llcccccccX}
        \hline
        Group & Lang. & Level & ESS & $n$ & $q$ & $U$ & \#Ranks & TC &
        Most Consistent Model \\
        \hline
        \multirow{4}{*}{W. South}
        & SLV & $E_1$     & 0.30 & 1 & 1 & 1 & 1  & 1.00 & Qwen3-Embedding-8B \\
        & SRP & $E_0$     & 0.00 & 0 & 0 & 0 & 0  & --   & -- \\
        & HRV & $E_0$     & 0.00 & 0 & 0 & 0 & 0  & --   & -- \\
        & BOS & $E_0$     & 0.00 & 0 & 0 & 0 & 0  & --   & -- \\
        \hline
        \multirow{2}{*}{E. South}
        & BUL & $E_1$     & 0.30 & 1 & 1 & 1 & 1  & 1.00 & Linq-Embed-Mistral \\
        & MKD & $E_1$     & 0.30 & 1 & 1 & 1 & 1  & 1.00 & Qwen3-Embedding-4B \\
        \hline
        \multirow{3}{*}{West}
        & POL & $E_{RS}$  & 0.70 & 4 & 4 & 1 & 15 & 0.99 & llama-embed-nemotron-8b \\
        & CES & $E_1$     & 0.30 & 1 & 1 & 1 & 1  & 1.00 & Qwen3-Embedding-8B \\
        & SLK & $E_{SC}$  & 0.23 & 2 & 1 & 1 & 1 & 1.00 & Qwen3-Embedding-8B \\
        \hline
        \multirow{3}{*}{East}
        & RUS & $E_{RS}$  & 0.60 & 2 & 2 & 1 & 15 & 1.00 & Qwen3-Embedding-8B \\
        & UKR & $E_0$     & 0.00 & 0 & 0 & 0 & 0  & --   & -- \\
        & BEL & $E_0$     & 0.00 & 0 & 0 & 0 & 0  & --   & -- \\
        \hline
      \end{tabularx}
    \end{small}
  \end{center}
\end{table}
 
\paragraph{Clustering}
 
Table~\ref{tab:summary_clustering} presents the models with the
highest task-specific top-$k$ consistency for the clustering task,
together with the corresponding evidence levels and ESS values.
Compared to classification, clustering exhibits a more homogeneous
consistency profile but remains characterized by limited benchmark
coverage.
Most languages are supported by only a single dataset~($E_1$),
resulting in low ESS values~($ESS=0.30$) and preventing meaningful
stability validation across datasets or ranking schemes.
Only Bosnian, Polish, and Czech provide sufficient benchmark coverage
to enable aggregation stability analysis~($E_{RS}$), yielding
substantially stronger evidence~($ESS=0.60$) based on two uncorrelated
datasets and fifteen ranking schemes.
 
A notable result is the complete dominance of
\textit{llama-embed-nemotron-8b}, which emerges as the most
consistent model for every evaluated language, achieving
$TC=1.00$ in all cases.
This degree of consistency is substantially stronger than that
observed for most other tasks and suggests a remarkably stable
performance profile for clustering.
 
However, the interpretation depends critically on the underlying
evidence strength.
For the majority of languages, perfect consistency is obtained under
single-dataset evidence~($E_1$), meaning that the observed ``stability"
reflects agreement within a single dataset rather than stability
validated across multiple evaluation conditions.
In contrast, the conclusions for Bosnian, Polish, and Czech are
supported by stronger evidence~($E_{RS}$, $ESS=0.60$), making the
corresponding stability assessments considerably more reliable.
The fact that \textit{llama-embed-nemotron-8b} remains the most
consistent model even in these higher-evidence settings provides
stronger support for its superiority on clustering tasks within this benchmark snapshot.
 
\begin{table}[!ht]
  \caption{Models with highest task-specific top-$k$ consistency (TC) by language on the clustering task.}
  \label{tab:summary_clustering}
  \begin{center}
    \begin{small}
      \begin{tabularx}{\columnwidth}{llcccccccX}
        \hline
        Group & Lang. & Level & ESS & $n$ & $q$ & $U$ & \#Ranks & TC &
        Most Consistent Model \\
        \hline
        \multirow{4}{*}{W. South}
        & SLV & $E_1$    & 0.30 & 1 & 1 & 1 & 1  & 1.00 & llama-embed-nemotron-8b \\
        & SRP & $E_1$    & 0.30 & 1 & 1 & 1 & 1  & 1.00 & llama-embed-nemotron-8b \\
        & HRV & $E_1$    & 0.30 & 1 & 1 & 1 & 1  & 1.00 & llama-embed-nemotron-8b \\
        & BOS & $E_{RS}$ & 0.60 & 2 & 2 & 1 & 15 & 1.00 & llama-embed-nemotron-8b \\
        \hline
        \multirow{2}{*}{E. South}
        & BUL & $E_1$    & 0.30 & 1 & 1 & 1 & 1  & 1.00 & llama-embed-nemotron-8b \\
        & MKD & $E_1$    & 0.30 & 1 & 1 & 1 & 1  & 1.00 & llama-embed-nemotron-8b \\
        \hline
        \multirow{3}{*}{West}
        & POL & $E_{RS}$ & 0.60 & 2 & 2 & 1 & 15 & 1.00 & llama-embed-nemotron-8b \\
        & CES & $E_{RS}$ & 0.60 & 2 & 2 & 1 & 15 & 1.00 & llama-embed-nemotron-8b \\
        & SLK & $E_1$    & 0.30 & 1 & 1 & 1 & 1  & 1.00 & llama-embed-nemotron-8b \\
        \hline
        \multirow{3}{*}{East}
        & RUS & $E_1$    & 0.30 & 1 & 1 & 1 & 1  & 1.00 & llama-embed-nemotron-8b \\
        & UKR & $E_1$    & 0.30 & 1 & 1 & 1 & 1  & 1.00 & llama-embed-nemotron-8b \\
        & BEL & $E_1$    & 0.30 & 1 & 1 & 1 & 1  & 1.00 & llama-embed-nemotron-8b \\
        \hline
      \end{tabularx}
    \end{small}
  \end{center}
\end{table}
 
\paragraph{Retrieval}
 
Table~\ref{tab:summary_retrieval} summarizes the models with the
highest task-specific top-$k$ consistency by language for the retrieval task.
Similar to clustering, retrieval exhibits highly uneven benchmark
coverage across languages.
Most languages are supported by only a single dataset~($E_1$),
resulting in low ESS values~($ESS=0.30$) and limiting conclusions to evidence extracted from individual datasets.
Aggregation stability analysis is possible only for Serbian,
Bulgarian, Czech, and Russian~($E_{RS}$), where two independent
datasets are available and the ESS increases to $0.60$.
No language provides sufficient benchmark diversity to support
composition stability analysis.
 
Unlike the classification and clustering tasks, retrieval does not
exhibit a single dominant model across all languages.
Instead, three models emerge as the most consistent under
different evaluation settings, i.e., \textit{LaBSE-ru-turbo},
\textit{bilingual-embedding-large}, and \textit{jina-embeddings-v3}.
\textit{LaBSE-ru-turbo} achieves perfect consistency~($TC=1.00$) in
most single-dataset settings.
In contrast, \textit{bilingual-embedding-large} emerges as the most
consistent model in the higher-evidence Serbian, Bulgarian, and Czech
settings.
For Russian, \textit{jina-embeddings-v3} achieves the highest consistency score~($TC=0.99$), slightly outperforming competing models despite the stronger stability requirements imposed by the aggregation stability analysis.
 
The retrieval results also differ from previous tasks in terms of
model composition.
Several models that consistently performed well in classification and
clustering, such as \textit{llama-embed-nemotron-8b}, do not appear
among the most consistent retrieval models.
This reflects the task-specific zero-shot filtering applied during
benchmark construction, which results in a different set
of eligible models for retrieval evaluation.
 
\begin{table}[!ht]
  \caption{Models with highest task-specific top-$k$ consistency (TC) by language on the retrieval task.}
  \label{tab:summary_retrieval}
  \begin{center}
    \begin{small}
      \begin{tabularx}{\columnwidth}{llcccccccX}
        \hline
        Group & Lang. & Level & ESS & $n$ & $q$ & $U$ & \#Ranks & TC &
        Most Consistent Model \\
        \hline
        \multirow{4}{*}{W. South}
        & SLV & $E_1$    & 0.30 & 1 & 1 & 1 & 1  & 1.00 & LaBSE-ru-turbo \\
        & SRP & $E_{RS}$ & 0.60 & 2 & 2 & 1 & 15 & 1.00 & bilingual-embedding-large \\
        & HRV & $E_1$    & 0.30 & 1 & 1 & 1 & 1  & 1.00 & LaBSE-ru-turbo \\
        & BOS & $E_0$    & 0.00 & 0 & 0 & 0 & 0  & --   & -- \\
        \hline
        \multirow{2}{*}{E. South}
        & BUL & $E_{RS}$ & 0.60 & 2 & 2 & 1 & 15 & 1.00 & bilingual-embedding-large \\
        & MKD & $E_1$    & 0.30 & 1 & 1 & 1 & 1  & 1.00 & LaBSE-ru-turbo \\
        \hline
        \multirow{3}{*}{West}
        & POL & $E_1$    & 0.30 & 1 & 1 & 1 & 1  & 1.00 & LaBSE-ru-turbo \\
        & CES & $E_{RS}$ & 0.60 & 2 & 2 & 1 & 15 & 1.00 & bilingual-embedding-large \\
        & SLK & $E_1$    & 0.30 & 1 & 1 & 1 & 1  & 1.00 & LaBSE-ru-turbo \\
        \hline
        \multirow{3}{*}{East}
        & RUS & $E_{RS}$ & 0.60 & 2 & 2 & 1 & 15 & 0.99 & jina-embeddings-v3 \\
        & UKR & $E_1$    & 0.30 & 1 & 1 & 1 & 1  & 1.00 & LaBSE-ru-turbo \\
        & BEL & $E_0$    & 0.00 & 0 & 0 & 0 & 0  & --   & -- \\
        \hline
      \end{tabularx}
    \end{small}
  \end{center}
\end{table}
 
\paragraph{Reranking}
 
Table~\ref{tab:summary_reranking} confirms that reranking is among
the most underrepresented evaluation settings.
Benchmark coverage is extremely sparse, with available evidence
limited to only four languages, i.e., Serbian, Bulgarian, Czech, and
Russian.
All four belong to the single-dataset evidence category~($E_1$,
$ESS=0.30$), so neither aggregation stability nor composition
stability can be evaluated, and top-$k$ consistency is derived from a ranking on a single dataset per language.
 
Despite this severe limitation, a consistent pattern
emerges: \textit{llama-embed-nemotron-8b} is identified as the most
consistent model in all four languages, achieving $TC=1.00$
in every case.
However, because all conclusions are based on single-dataset
evidence, the observed consistency should not be interpreted as
evidence of stable cross-dataset performance.
 
\begin{table}[!ht]
  \caption{Models with highest task-specific top-$k$ consistency (TC) by language on the reranking task.}
  \label{tab:summary_reranking}
  \begin{center}
    \begin{small}
      \begin{tabularx}{\columnwidth}{llcccccccX}
        \hline
        Group & Lang. & Level & ESS & $n$ & $q$ & $U$ & \#Ranks & TC &
        Most Consistent Model \\
        \hline
        \multirow{4}{*}{W. South}
        & SLV & $E_0$ & 0.00 & 0 & 0 & 0 & 0 & -- & -- \\
        & SRP & $E_1$ & 0.30 & 1 & 1 & 1 & 1 & 1.00 & llama-embed-nemotron-8b \\
        & HRV & $E_0$ & 0.00 & 0 & 0 & 0 & 0 & -- & -- \\
        & BOS & $E_0$ & 0.00 & 0 & 0 & 0 & 0 & -- & -- \\
        \hline
        \multirow{2}{*}{E. South}
        & BUL & $E_1$ & 0.30 & 1 & 1 & 1 & 1 & 1.00 & llama-embed-nemotron-8b \\
        & MKD & $E_0$ & 0.00 & 0 & 0 & 0 & 0 & -- & -- \\
        \hline
        \multirow{3}{*}{West}
        & POL & $E_0$ & 0.00 & 0 & 0 & 0 & 0 & -- & -- \\
        & CES & $E_1$ & 0.30 & 1 & 1 & 1 & 1 & 1.00 & llama-embed-nemotron-8b \\
        & SLK & $E_0$ & 0.00 & 0 & 0 & 0 & 0 & -- & -- \\
        \hline
        \multirow{3}{*}{East}
        & RUS & $E_1$ & 0.30 & 1 & 1 & 1 & 1 & 1.00 & llama-embed-nemotron-8b \\
        & UKR & $E_0$ & 0.00 & 0 & 0 & 0 & 0 & -- & -- \\
        & BEL & $E_0$ & 0.00 & 0 & 0 & 0 & 0 & -- & -- \\
        \hline
      \end{tabularx}
    \end{small}
  \end{center}
\end{table}
 
\paragraph{STS}
 
Table~\ref{tab:summary_sts} reveals the most severe benchmark
scarcity among all evaluated tasks.
Evaluation evidence is available only for Polish and Russian, while
all remaining Slavic languages fall into the no-evidence
category~($E_0$).
For both languages, only a single dataset is available~($E_1$,
$ESS=0.30$), so neither aggregation stability nor composition
stability can be assessed.
In both languages, \textit{Octen-Embedding-8B} achieves perfect
top-$k$ consistency~($TC=1.00$), though this result is derived from
a single ranking instance and therefore provides limited evidence
regarding broader rank stability.
 
\begin{table}[!ht]
  \caption{Models with highest task-specific top-$k$ consistency (TC) by language on the STS task.}
  \label{tab:summary_sts}
  \begin{center}
    \begin{small}
      \begin{tabularx}{\columnwidth}{llcccccccX}
        \hline
        Group & Lang. & Level & ESS & $n$ & $q$ & $U$ & \#Ranks & TC &
        Most Consistent Model \\
        \hline
        \multirow{4}{*}{W. South}
        & SLV & $E_0$ & 0.00 & 0 & 0 & 0 & 0 & -- & -- \\
        & SRP & $E_0$ & 0.00 & 0 & 0 & 0 & 0 & -- & -- \\
        & HRV & $E_0$ & 0.00 & 0 & 0 & 0 & 0 & -- & -- \\
        & BOS & $E_0$ & 0.00 & 0 & 0 & 0 & 0 & -- & -- \\
        \hline
        \multirow{2}{*}{E. South}
        & BUL & $E_0$ & 0.00 & 0 & 0 & 0 & 0 & -- & -- \\
        & MKD & $E_0$ & 0.00 & 0 & 0 & 0 & 0 & -- & -- \\
        \hline
        \multirow{3}{*}{West}
        & POL & $E_1$ & 0.30 & 1 & 1 & 1 & 1 & 1.00 & Octen-Embedding-8B \\
        & CES & $E_0$ & 0.00 & 0 & 0 & 0 & 0 & -- & -- \\
        & SLK & $E_0$ & 0.00 & 0 & 0 & 0 & 0 & -- & -- \\
        \hline
        \multirow{3}{*}{East}
        & RUS & $E_1$ & 0.30 & 1 & 1 & 1 & 1 & 1.00 & Octen-Embedding-8B \\
        & UKR & $E_0$ & 0.00 & 0 & 0 & 0 & 0 & -- & -- \\
        & BEL & $E_0$ & 0.00 & 0 & 0 & 0 & 0 & -- & -- \\
        \hline
      \end{tabularx}
    \end{small}
  \end{center}
\end{table}
 
\paragraph{Pair Classification}
 
Table~\ref{tab:summary_pairclassification} summarizes the models with
the highest task-specific top-$k$ consistency for the pair
classification task.
Similar to reranking and STS, pair classification suffers from limited
benchmark coverage, with evaluation evidence available for only four
languages, i.e., Bulgarian, Polish, Czech, and Russian.
Bulgarian, Polish, and Czech are supported by a single
dataset~($E_1$, $ESS=0.30$).
Russian provides the strongest evidence setting for this task,
supporting aggregation stability analysis~($E_{RS}$) through three datasets~($n=q=3$) and achieving $ESS=0.65$.
 
A highly consistent pattern emerges: \textit{Qwen3-Embedding-8B} is
identified as the most consistent model in Bulgarian, Czech,
and Russian, while \textit{Qwen3-Embedding-4B} achieves the highest
consistency in Polish.
All reported top models achieve $TC=1.00$.
However, for Bulgarian, Polish, and Czech, perfect consistency is
obtained from a single ranking instance, whereas the Russian result is
supported by fifteen ranking schemes and substantially stronger
benchmark evidence.
 
\begin{table}[!ht]
  \caption{Models with highest task-specific top-$k$ consistency (TC) by language on the pair classification task.}
  \label{tab:summary_pairclassification}
  \begin{center}
    \begin{small}
      \begin{tabularx}{\columnwidth}{llcccccccX}
        \hline
        Group & Lang. & Level & ESS & $n$ & $q$ & $U$ & \#Ranks & TC &
        Most Consistent Model \\
        \hline
        \multirow{4}{*}{W. South}
        & SLV & $E_0$    & 0.00 & 0 & 0 & 0 & 0  & --   & -- \\
        & SRP & $E_0$    & 0.00 & 0 & 0 & 0 & 0  & --   & -- \\
        & HRV & $E_0$    & 0.00 & 0 & 0 & 0 & 0  & --   & -- \\
        & BOS & $E_0$    & 0.00 & 0 & 0 & 0 & 0  & --   & -- \\
        \hline
        \multirow{2}{*}{E. South}
        & BUL & $E_1$    & 0.30 & 1 & 1 & 1 & 1  & 1.00 & Qwen3-Embedding-8B \\
        & MKD & $E_0$    & 0.00 & 0 & 0 & 0 & 0  & --   & -- \\
        \hline
        \multirow{3}{*}{West}
        & POL & $E_1$    & 0.30 & 1 & 1 & 1 & 1  & 1.00 & Qwen3-Embedding-4B \\
        & CES & $E_1$    & 0.30 & 1 & 1 & 1 & 1  & 1.00 & Qwen3-Embedding-8B \\
        & SLK & $E_0$    & 0.00 & 0 & 0 & 0 & 0  & --   & -- \\
        \hline
        \multirow{3}{*}{East}
        & RUS & $E_{RS}$ & 0.65 & 3 & 3 & 1 & 15 & 1.00 & Qwen3-Embedding-8B \\
        & UKR & $E_0$    & 0.00 & 0 & 0 & 0 & 0  & --   & -- \\
        & BEL & $E_0$    & 0.00 & 0 & 0 & 0 & 0  & --   & -- \\
        \hline
      \end{tabularx}
    \end{small}
  \end{center}
\end{table}
 
\paragraph{Multilabel Classification}
 
Table~\ref{tab:summary_multilabelclassification} summarizes the models
with the highest task-specific top-$k$ consistency for the multilabel
classification task.
Similar to reranking and pair classification, this task is
characterized by sparse and uneven benchmark coverage.
Evaluation evidence is available only for Slovene, Croatian,
Bulgarian, Polish, Czech, Slovak, and Russian, all with single-dataset
evidence~($E_1$, $ESS=0.30$).
Consequently, neither aggregation stability nor composition stability
can be assessed.
 
Despite the limited benchmark coverage, a consistent
pattern emerges: \textit{KaLM-Embedding-Gemma3-12B-2511} achieves the
highest top-$k$ consistency~($TC=1.00$) in every language with
available evidence except Russian.
However, because all conclusions are based on single-dataset
evaluations~($E_1$), the observed perfect consistency reflects
stability within individual benchmark instances rather than stability
validated across multiple datasets or ranking schemes.
Multilabel classification therefore provides one of the clearest
illustrations of the distinction between top-$k$ consistency and
evidence strength: the model is uniformly identified as the top
performer, yet the supporting evidence is insufficient to confirm this
as genuinely robust behavior.

\begin{table}[!ht]
  \caption{Models with highest task-specific top-$k$ consistency (TC) by language on the multilabel classification task.}
  \label{tab:summary_multilabelclassification}
  \begin{center}
    \begin{small}
      \begin{tabularx}{\columnwidth}{Xlcccccccl}
        \hline
        Group & Lang & L & ESS & $n$ & $q$ & $U$ & \#Rnk & TC &
        Most Consistent Model \\
        \hline
        \multirow{4}{*}{W. South}
        & SLV & $E_1$ & 0.30 & 1 & 1 & 1 & 1 & 1.00 &
          KaLM-Embedding-Gemma3-12B-2511 \\
        & SRP & $E_0$ & 0.00 & 0 & 0 & 0 & 0 & -- & -- \\
        & HRV & $E_1$ & 0.30 & 1 & 1 & 1 & 1 & 1.00 &
          KaLM-Embedding-Gemma3-12B-2511 \\
        & BOS & $E_0$ & 0.00 & 0 & 0 & 0 & 0 & -- & -- \\
        \hline
        \multirow{2}{*}{E. South}
        & BUL & $E_1$ & 0.30 & 1 & 1 & 1 & 1 & 1.00 &
          KaLM-Embedding-Gemma3-12B-2511 \\
        & MKD & $E_0$ & 0.00 & 0 & 0 & 0 & 0 & -- & -- \\
        \hline
        \multirow{3}{*}{West}
        & POL & $E_1$ & 0.30 & 1 & 1 & 1 & 1 & 1.00 &
          KaLM-Embedding-Gemma3-12B-2511 \\
        & CES & $E_1$ & 0.30 & 1 & 1 & 1 & 1 & 1.00 &
          KaLM-Embedding-Gemma3-12B-2511 \\
        & SLK & $E_1$ & 0.30 & 1 & 1 & 1 & 1 & 1.00 &
          KaLM-Embedding-Gemma3-12B-2511 \\
        \hline
        \multirow{3}{*}{East}
        & RUS & $E_1$ & 0.30 & 1 & 1 & 1 & 1 & 1.00 &
          llama-embed-nemotron-8b \\
        & UKR & $E_0$ & 0.00 & 0 & 0 & 0 & 0 & -- & -- \\
        & BEL & $E_0$ & 0.00 & 0 & 0 & 0 & 0 & -- & -- \\
        \hline
      \end{tabularx}
    \end{small}
  \end{center}
\end{table}
 
\paragraph{Bitext Mining}
 
Table~\ref{tab:summary_bitextmining} presents the models with the
highest task-specific top-$k$ consistency for bitext mining, which
stands out as the strongest evidence setting in the entire study.
Unlike all other tasks, bitext mining provides sufficient benchmark
coverage to support both aggregation stability and composition
stability analysis for a large subset of languages.
Serbian, Croatian, Polish, Czech, Russian, Ukrainian, and Belarusian
all belong to the highest evidence category~($E_{RS+DS}$), with ESS
values ranging from $0.83$ to $0.85$.
In contrast, Slovene, Bosnian, Bulgarian, Macedonian, and Slovak
belong to the single-cluster evidence category~($E_{SC}$), where
multiple datasets are available but provide redundant
evaluation evidence.

The task exhibits a clear separation between two groups of models.
For the higher-evidence $E_{RS+DS}$ languages, \textit{LaBSE-en-ru} emerges as the most consistent model, achieving
TC scores between $0.83$ and $0.88$.
In contrast, the $E_{SC}$ languages are dominated by
\textit{multilingual-e5-large-instruct}, which achieves perfect
consistency~($TC=1.00$) in all such settings.
This pattern is not primarily driven by linguistic differences, but
rather by benchmark composition: the languages where \textit{LaBSE-en-ru} dominates largely share the same underlying bitext-mining datasets,
while the languages favoring \textit{multilingual-e5-large-instruct}
are evaluated on a set of highly correlated benchmarks.
 
Bitext mining is the only task where top-$k$ consistency can be
interpreted together with strong evidence strength.
Although \textit{multilingual-e5-large-instruct} achieves perfect
consistency in several languages, these results are supported only by
single-cluster evidence~($E_{SC}$, $ESS=0.23$) and therefore provide
weaker stability evidence.
By contrast, the slightly lower TC scores of \textit{LaBSE-en-ru} are
supported by substantially stronger evidence~($E_{RS+DS}$,
$ESS>0.80$), making these conclusions considerably more reliable.

\begin{table}[!ht]
  \caption{Models with highest task-specific top-$k$ consistency (TC) by language on the bitext mining task.}
  \label{tab:summary_bitextmining}
  \begin{center}
    \begin{small}
      \begin{tabularx}{\columnwidth}{llcccccccX}
        \hline
        Group & Lang. & Level & ESS & $n$ & $q$ & $U$ & \#Ranks & TC &
        Most Consistent Model \\
        \hline
        \multirow{4}{*}{W. South}
        & SLV & $E_{SC}$     & 0.23 & 3 & 1 & 1 & 1 & 1.00 &
          multilingual-e5-large-instruct \\
        & SRP & $E_{RS+DS}$  & 0.83 & 4 & 2 & 3 & 45 & 0.86 & LaBSE-en-ru \\
        & HRV & $E_{RS+DS}$  & 0.83 & 4 & 2 & 3 & 45 & 0.86 & LaBSE-en-ru \\
        & BOS & $E_{SC}$     & 0.23 & 3 & 1 & 1 & 1 & 1.00 &
          multilingual-e5-large-instruct \\
        \hline
        \multirow{2}{*}{E. South}
        & BUL & $E_{SC}$     & 0.23 & 3 & 1 & 1 & 1 & 1.00 &
          multilingual-e5-large-instruct \\
        & MKD & $E_{SC}$     & 0.23 & 3 & 1 & 1 & 1 & 1.00 &
          multilingual-e5-large-instruct \\
        \hline
        \multirow{4}{*}{West}
        & POL & $E_{RS+DS}$  & 0.83 & 4 & 2 & 3 & 45 & 0.86 & LaBSE-en-ru \\
        & CES & $E_{RS+DS}$  & 0.83 & 4 & 2 & 3 & 45 & 0.86 & LaBSE-en-ru \\
        & SLK & $E_{SC}$     & 0.23 & 3 & 1 & 1 & 1 & 1.00 &
          multilingual-e5-large-instruct \\
        \hline
        \multirow{3}{*}{East}
        & RUS & $E_{RS+DS}$  & 0.85 & 5 & 2 & 3 & 45 & 0.83 & bge-m3 \\
        & UKR & $E_{RS+DS}$  & 0.83 & 4 & 2 & 3 & 45 & 0.86 & LaBSE-en-ru \\
        & BEL & $E_{RS+DS}$  & 0.83 & 4 & 2 & 3 & 45 & 0.88 & LaBSE-en-ru \\
        \hline
      \end{tabularx}
    \end{small}
  \end{center}
\end{table}

\paragraph{Sensitivity to threshold $k$}

The sensitivity analysis in~\ref{appendix:topk_sensitivity} evaluates the effect of varying the top-$k$ threshold across the task-language combinations. Relative to the reference value $k=10$, the closest alternatives ($k=8$ and $k=12$) yield a mean absolute difference of 0.0468 in consistency scores, with mean Spearman correlations of 0.9469 and 0.9656, respectively. Larger changes in $k$ produce greater score differences, although model ordering remains comparatively stable for $k=15$ and $k=20$. These findings indicate that the consistency scores depend on the definition of top-performing models, while their relative ordering is generally preserved across the examined configurations.

\subsection{Language-Specific Cross-Task Consistency}
 
\subsubsection{Evidence Strength}
 
\begin{figure*}[!ht]
\centering
  \includegraphics[width=\linewidth]{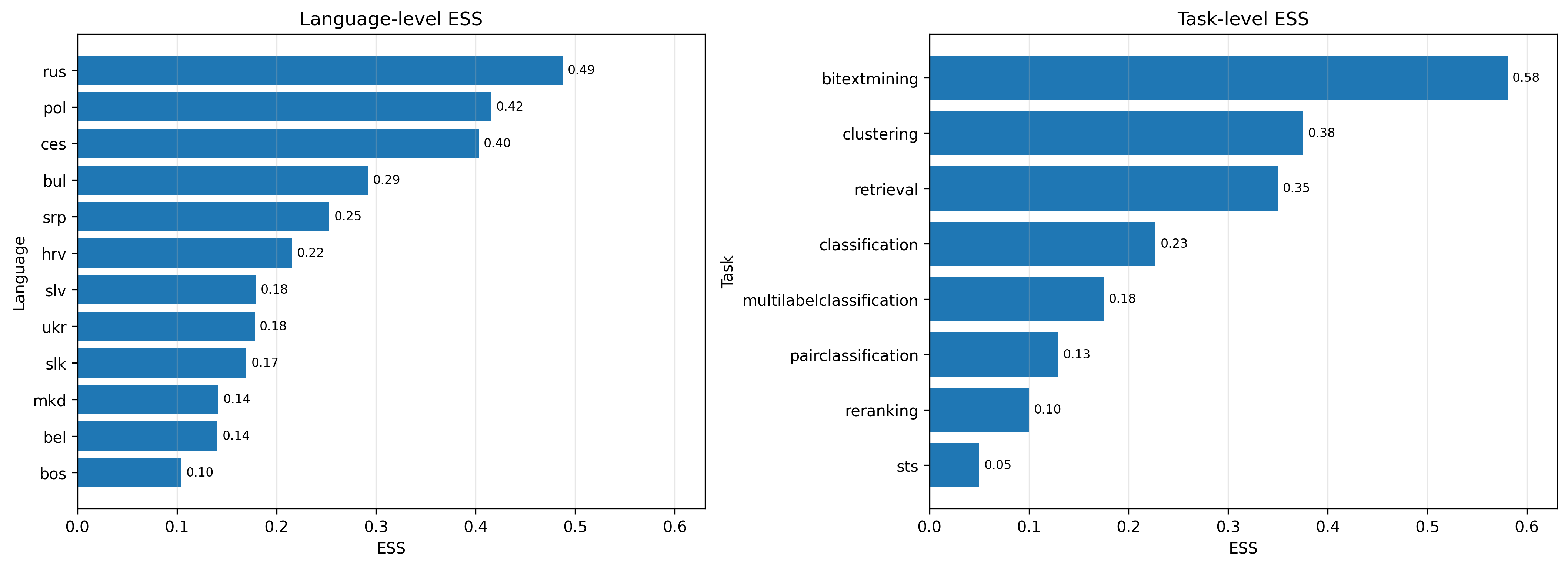}
  \caption{Aggregated ESS across languages~(left) and tasks~(right).}
  \label{fig:evidence_strength_language_task_bars}
\end{figure*}

Figure~\ref{fig:evidence_strength_language_task_bars} summarizes the aggregated ESS across languages and tasks, providing important context for interpreting the reliability of the cross-task consistency analysis. Since language-level ESS averages task-specific evidence across all analyzed tasks, including tasks without available datasets, it jointly reflects benchmark coverage, dataset diversity, and stability assessability at the language level. Consequently, higher language-level ESS indicates that cross-task consistency conclusions are supported by broader and more reliable evaluation evidence.
 
Russian exhibits the highest language-level ESS~($0.49$), followed by Polish~($0.42$) and Czech~($0.40$), indicating that cross-task consistency results for these languages are supported by comparatively strong and diverse benchmark evidence across multiple tasks. Bulgarian and Serbian form an intermediate group with moderate evidence strength, while Croatian, Slovene, Ukrainian, and Slovak exhibit weaker evidence profiles due to sparse benchmark coverage and limited stability assessability across several tasks. Bosnian represents the weakest evidence settings overall.
 
The task-level ESS further explains these language-level differences. \textit{Bitext mining} clearly dominates the benchmark landscape with the highest task-level ESS~($0.58$), followed by \textit{clustering}~($0.38$) and \textit{retrieval}~($0.35$). Consequently, cross-task consistency is heavily influenced by these tasks. In contrast, tasks such as \textit{STS}~($0.05$), \textit{reranking}~($0.10$), and \textit{pair classification}~($0.13$) contribute only weak evidence due to sparse or highly redundant benchmark coverage.
 
These results demonstrate that cross-task consistency should not be interpreted independently of evidence strength. Apparent consistency for lower-coverage languages may partially reflect limited benchmark variability rather than genuinely stable multilingual behavior. In contrast, consistency results for Russian, Polish, and Czech are considerably more reliable because they are supported by stronger evidence across multiple tasks and evaluation settings.
 
\subsubsection{Cross-Task Consistency}

\begin{figure*}[!ht]
\centering
  \includegraphics[width=\linewidth]{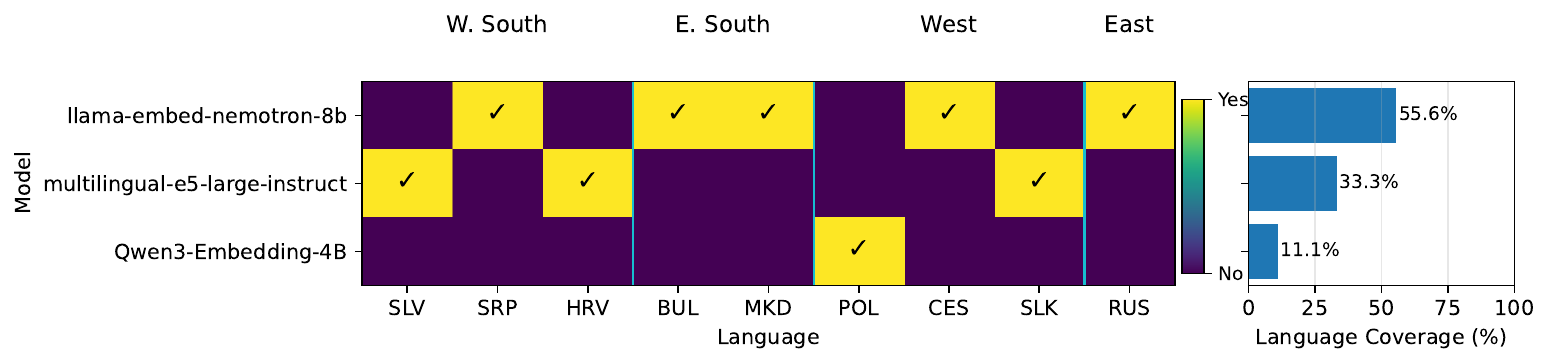}
  \caption{Binary heatmap showing whether a model is among the most consistent cross-task models for each language~(left). Overall coverage of each model, measured as the percentage of languages for which it is top ranked~(right).}
  \label{fig:robust_models_heatmap_coverage}
\end{figure*}

Table~\ref{tab:cross_task_model_consistency} and Figure~\ref{fig:robust_models_heatmap_coverage} summarize cross-task consistency for languages with task coverage exceeding~50\%, with \ref{appendix:interpretationESS} providing details on the qualitative interpretation of the ESS score given in the Evidence column. \ref{appendix:cross_task_detailed} provides details for a selected set of languages. The project repository contains detailed results for all evaluated languages.

While task-specific analysis evaluates stability within individual tasks, cross-task consistency measures how consistently a model remains among the highest-ranked methods across the evaluated task families within the same language.
The results reveal a considerably more stable picture than the task-specific analysis. While different models dominate individual tasks, cross-task consistency is concentrated within a very small set of multilingual embedding models. In particular, \textit{llama-embed-nemotron-8b}, \textit{multilingual-e5-large-instruct}, and \textit{Qwen3-Embedding} variants repeatedly occupy the top positions across languages. Figure~\ref{fig:robust_models_heatmap_coverage} further illustrates this concentration: \textit{llama-embed-nemotron-8b} leads in~55.6\% of the analyzed languages, \textit{multilingual-e5-large-instruct} in 33.3\%, and \textit{Qwen3-Embedding-4B} in~11.1\% (Polish only). However, the differences between the top-ranked models are often small, indicating that a small group of large multilingual models exhibit comparable consistency
within the evaluated benchmark settings rather than establishing a
single universally superior model.

\begin{table}[!ht]
  \caption{Most cross-task consistent models by language. Evidence denotes the language-level evidence category, ESS is the language-level Evidence Strength Score, Tasks is the number of evaluable tasks out of all analyzed tasks, and $CW$ is the coverage-weighted cross-task consistency score.}
  \label{tab:cross_task_model_consistency}
  \begin{center}
    \begin{small}
      \begin{tabularx}{\columnwidth}{llccccX}
        \hline
        Group & Lang. & Evidence & ESS & Tasks & $CW$ &
        Most Consistent Model \\
        \hline
        \multirow{3}{*}{W. South}
        & SLV & Weak     & 0.18 & 5/8 & 0.33 & multilingual-e5-large-instruct \\
        & SRP & Moderate & 0.25 & 4/8 & 0.31 & llama-embed-nemotron-8b \\
        & HRV & Weak     & 0.22 & 4/8 & 0.30 & multilingual-e5-large-instruct \\
        \hline
        \multirow{2}{*}{E. South}
        & BUL & Moderate & 0.29 & 7/8 & 0.52 & llama-embed-nemotron-8b \\
        & MKD & Weak     & 0.14 & 4/8 & 0.31 & llama-embed-nemotron-8b \\
        \hline
        \multirow{3}{*}{West}
        & POL & Comparatively strong   & 0.42 & 7/8 & 0.53 & Qwen3-Embedding-4B \\
        & CES & Comparatively strong   & 0.40 & 7/8 & 0.53 & llama-embed-nemotron-8b \\
        & SLK & Weak     & 0.17 & 5/8 & 0.40 & multilingual-e5-large-instruct \\
        \hline
        \multirow{1}{*}{East}
        & RUS & Comparatively strong   & 0.49 & 8/8 & 0.70 & llama-embed-nemotron-8b \\
        \hline
      \end{tabularx}
    \end{small}
  \end{center}
\end{table}
 
At the subgroup level, the strongest competition is observed in the West South languages. Slovene is nominally led by \textit{multilingual-e5-large-instruct} ($CW^{(\mathrm{slv})}=0.33$), but \textit{llama-embed-nemotron-8b} follows almost immediately, indicating that
both models achieve similar cross-task consistency scores on the
evaluated tasks. Serbian exhibits the reverse ordering, with \textit{llama-embed-nemotron-8b} slightly outperforming \textit{multilingual-e5-large-instruct}, again with a negligible margin. Croatian again favors \textit{multilingual-e5-large-instruct}. Taken together, the West South subgroup suggests that cross-task consistency is effectively shared between \textit{llama-embed-nemotron-8b} and \textit{multilingual-e5-large-instruct}.
 
The East South subgroup is more homogeneous. Both Bulgarian and Macedonian identify \textit{llama-embed-nemotron-8b} as the strongest cross-task model, with \textit{multilingual-e5-large-instruct} remaining highly competitive. Bulgarian represents one of the most reliable results in the study, combining high task coverage~(7/8 tasks) with moderate evidence strength~($ESS=0.29$).
 
The West subgroup exhibits the greatest diversity. Polish stands out because \textit{Qwen3-Embedding-4B} achieves the highest cross-task consistency, making it the only language where a \textit{Qwen3} model clearly occupies the first position. However, both \textit{llama-embed-nemotron-8b} and \textit{Qwen3-Embedding-8B} achieve nearly identical scores, suggesting that the entire \textit{Qwen3} family performs well on Polish benchmarks. Czech favors \textit{llama-embed-nemotron-8b}, but \textit{Qwen3-Embedding-8B} remains very close. Slovak favors \textit{multilingual-e5-large-instruct}, with \textit{llama-embed-nemotron-8b} as a second, with larger difference.
 
The East subgroup is represented by Russian, which provides the strongest and most comprehensive evidence in the entire study. Russian is the only language with complete task coverage~(8/8 tasks), the highest ESS~($0.49$), and the highest coverage-weighted cross-task consistency~($CW^{(\mathrm{rus})}=0.70$). Although \textit{llama-embed-nemotron-8b} occupies the first position, both \textit{Qwen3-Embedding-8B} and \textit{Octen-Embedding-8B} achieve similarly strong consistency.
 
Overall, the cross-task analysis identifies a small set of multilingual embedding models that consistently achieve high rankings across the evaluated task families. \textit{llama-embed-nemotron-8b} most frequently achieves the highest coverage-weighted cross-task consistency score across the analyzed languages, including those with comparatively stronger benchmark evidence. However, its advantage is often modest relative to \textit{multilingual-e5-large-instruct} and \textit{Qwen3-Embedding-8B}. The results therefore suggest that observed cross-task ranking consistency is concentrated among several competitive multilingual models whose relative ordering depends on the language and the available benchmark evidence. These findings describe consistency within the analyzed benchmark snapshot, which should not be interpreted as generalization to unseen tasks.
 
\section{Discussion}
 
The cross-task analysis reveals an important distinction between task-specific superiority and cross-task high rank consistency. The inspection of the task-level consistency scores shows that many task-specific winners do not emerge as the strongest cross-task models. Instead, the models that dominate the cross-task rankings are typically those that maintain consistently strong performance across several task families rather than achieving first place in any single evaluation setting. This distinction highlights the importance of evaluating both task-specific and cross-task consistency when selecting multilingual embedding models.
 
A first observation is the existence of clear task specialization. Different task families favor different model families. Classification and pair classification tasks are dominated by \textit{Qwen3-Embedding} variants, particularly \textit{Qwen3-Embedding-8B} and \textit{Qwen3-Embedding-4B}. Clustering and reranking are strongly associated with \textit{llama-embed-nemotron-8b}, which emerges as the most consistent model in the majority of evaluated languages. Retrieval exhibits a completely different set of leaders, with \textit{LaBSE-ru-turbo}, \textit{bilingual-embedding-large}, and \textit{jina-embeddings-v3} outperforming models that dominate other tasks (potentially due to the zero-shot model filtering). Similarly, STS is led by \textit{Octen-Embedding-8B}, multilabel classification by \textit{KaLM-Embedding-Gemma3-12B-2511}, and bitext mining by either \textit{multilingual-e5-large-instruct} or \textit{LaBSE-en-ru} depending on the evidence regime. These patterns indicate that no single architecture dominates all embedding use cases and that different task families emphasize substantially different capabilities.
 
At the same time, the cross-task rankings reveal that task specialization alone is insufficient to explain overall model utility. Several task-specific leaders achieve exceptional performance within a narrow evaluation domain but rarely appear among the strongest cross-task models. For example, \textit{LaBSE-ru-turbo} dominates retrieval benchmarks, \textit{Octen-Embedding-8B} dominates STS, \textit{KaLM-Embedding-Gemma3-12B-2511} dominates multilabel classification. Despite their strong task-specific consistency, these models generally do not emerge as the most consistent cross-task models. In contrast, models such as \textit{llama-embed-nemotron-8b}, \textit{multilingual-e5-large-instruct}, and \textit{Qwen3-Embedding} variants consistently rank near the top across multiple task families, accumulating substantially higher cross-task consistency scores even when they are not the best-performing model within a particular task.
 
The task-level analysis also provides insight into why \textit{llama-embed-nemotron-8b} emerges as the dominant cross-task model. One contributing factor is the broad availability of clustering benchmarks across languages: clustering is among the best-covered tasks and is consistently dominated by this model. Because cross-task consistency aggregates information across all evaluable tasks, strong performance in a widely available task family contributes substantially to the final rankings. However, clustering alone does not explain the observed dominance, since the model also achieves competitive results in reranking, classification, and several retrieval settings, allowing it to remain near the top even when it is not the single strongest performer in the task.
 
Another important observation is that the cross-task results exhibit considerably greater stability than the task-specific results. Task-specific analyses often identify different leaders depending on the language, dataset composition, and task. In contrast, the cross-task rankings repeatedly highlight the same small set of models, suggesting that the principal source of variability in multilingual embedding evaluation is task specialization rather than language variation. Once performance is aggregated across multiple task families, the rankings become substantially more stable and less sensitive to individual benchmark characteristics.
 
Taken together, these findings suggest that multilingual embedding models can be divided into two broad categories. The first consists of \textit{task specialists}, such as \textit{LaBSE-ru-turbo}, \textit{Octen-Embedding-8B}, \textit{KaLM-Embedding-Gemma3-12B-2511}, which achieve state-of-the-art performance in specific task families but do not consistently lead the cross-task rankings. The second consists of \textit{generalist models}, including \textit{llama-embed-nemotron-8b}, \textit{multilingual-e5-large-instruct}, and \textit{Qwen3-Embedding} variants, which may not always achieve the highest task-specific scores but consistently perform well across diverse evaluation settings. However, these results should not be interpreted as evidence of model generalization to unseen tasks, since the analysis is restricted to the available benchmark tasks and their evaluation coverage within the evaluated benchmark snapshot.

\section{Conclusion}
 
We presented a language-aware stability analysis of multilingual embedding models across Slavic languages, considering both task-specific stability and cross-task consistency. Our results show that evaluation is largely constrained by dataset scarcity: most language-task pairs rely on a single dataset or none, limiting stability assessment and leading to conclusions based on isolated outcomes. 
We introduced evidence levels and the Evidence Strength Score to characterize the adequacy of available benchmark evidence, distinguishing ranking stability supported by diverse datasets from apparent stability arising from sparse or redundant evaluation resources.
Despite these limitations, a small group of large multilingual models, such as \textit{llama-embed-nemotron-8b}, \textit{Qwen3-Embedding}, and \textit{multilingual-e5}, consistently demonstrates stable cross-task consistency, while most models remain task- and language-specific within the evaluated benchmark settings. These findings characterize observed ranking patterns rather than generalization to unseen tasks. Overall, the analyzed benchmark snapshot provides insufficient evidence for reliable conclusions in many Slavic languages, emphasizing the need for richer and more task-balanced evaluation resources.
 
\section*{Limitations}
 
Our work focuses on benchmarking meta-analysis rather than model selection, examining how evaluation conclusions change under different evaluation settings, namely dataset composition and aggregation methods. In this sense, it complements standard aggregate metrics by assessing the stability of benchmark results. The study inherits limitations typical of leaderboard-based evaluations: (i)~it relies on a single snapshot of publicly reported MTEB results, which may change over time, but releases the source code to allow rerunning on new snapshots; (ii)~it treats reported dataset-level performance scores as fixed
observations and does not account for uncertainty arising from
repeated evaluation runs or alternative data splits (i.e., the proposed framework assesses the adequacy of
available benchmark evidence and the stability of rankings under
different aggregation methods and dataset compositions, rather than
the statistical reliability of the underlying model performance
estimates); (iii)~it may inherit inaccuracies from MTEB's user-submitted model metadata and scores; and 
(iv)~the number and identity of eligible models vary across
task-language pairs, and the rank-based consistency measure
does not explicitly normalize for candidate set size.
Consequently, differences in cross-task consistency may partly
reflect differences in the evaluated model pools rather than
model behavior alone.
 
\section*{Acknowledgments}

This work is funded by the Slovenian Research and Innovation Agency under program grant P2-0098, project grants No. GC-0001 and No. J2-70078; and by the European Union under Grant Agreement 101211695 (HE MSCA-PF AutoLLMSelect) and Grant Agreement 101187010 (HE ERA Chair AutoLearn-SI).

\bibliographystyle{unsrt}

\begin{thebibliography}{10}

\bibitem{feng2022language}
Fangxiaoyu Feng, Yinfei Yang, Daniel Cer, Naveen Arivazhagan, and Wei Wang.
\newblock Language-agnostic bert sentence embedding.
\newblock In {\em Proceedings of the 60th annual meeting of the association for computational linguistics (volume 1: Long papers)}, pages 878--891, 2022.

\bibitem{reimers2019sentence}
Nils Reimers and Iryna Gurevych.
\newblock Sentence-bert: Sentence embeddings using siamese bert-networks.
\newblock In {\em Proceedings of the 2019 conference on empirical methods in natural language processing and the 9th international joint conference on natural language processing (EMNLP-IJCNLP)}, pages 3982--3992, 2019.

\bibitem{wang2024multilingual}
Liang Wang, Nan Yang, Xiaolong Huang, Linjun Yang, Rangan Majumder, and Furu Wei.
\newblock Multilingual e5 text embeddings: A technical report.
\newblock {\em arXiv preprint arXiv:2402.05672}, 2024.

\bibitem{muennighoff2023mteb}
Niklas Muennighoff, Nouamane Tazi, Lo{\"\i}c Magne, and Nils Reimers.
\newblock Mteb: Massive text embedding benchmark.
\newblock In {\em Proceedings of the 17th Conference of the European Chapter of the Association for Computational Linguistics}, pages 2014--2037, 2023.

\bibitem{enevoldsen2025mmteb}
Kenneth Enevoldsen, Isaac Chung, Imene Kerboua, M{\'a}rton Kardos, Ashwin Mathur, David Stap, Jay Gala, Wissam Siblini, Dominik Krzemi{\'n}ski, Genta Winata, et~al.
\newblock Mmteb: Massive multilingual text embedding benchmark.
\newblock In {\em International Conference on Learning Representations}, volume 2025, pages 101715--101771, 2025.

\bibitem{gjorgjevikj2026robustness}
Ana Gjorgjevikj, Barbara~Korou{\v{s}}i{\'c} Seljak, and Tome Eftimov.
\newblock On the robustness of multilingual text embedding rankings across learning tasks, languages, and benchmark datasets.
\newblock {\em arXiv preprint arXiv:2605.31142}, 2026.

\bibitem{colombo2022best}
Pierre Colombo, Nathan Noiry, Ekhine Irurozki, and St{\'e}phan Cl{\'e}men{\c{c}}on.
\newblock What are the best systems? new perspectives on nlp benchmarking.
\newblock {\em Advances in neural information processing systems}, 35:26915--26932, 2022.

\bibitem{rofin2023vote}
Mark Rofin, Vladislav Mikhailov, Mikhail Florinsky, Andrey Kravchenko, Tatiana Shavrina, Elena Tutubalina, Daniel Karabekyan, and Ekaterina Artemova.
\newblock Vote’n’rank: Revision of benchmarking with social choice theory.
\newblock In {\em Proceedings of the 17th Conference of the European Chapter of the Association for Computational Linguistics}, pages 670--686, 2023.

\bibitem{frank2026pteb}
Manuel Frank and Haithem Afli.
\newblock Pteb: Towards robust text embedding evaluation via stochastic paraphrasing at evaluation time with llms.
\newblock In {\em Proceedings of the 19th Conference of the European Chapter of the Association for Computational Linguistics (Volume 1: Long Papers)}, pages 2832--2851, 2026.

\bibitem{sussex2006slavic}
Roland Sussex and Paul Cubberley.
\newblock {\em The slavic languages}.
\newblock Cambridge University Press Cambridge, 2006.

\bibitem{piskorski2023proceedings}
Jakub Piskorski, Micha{\l} Marci{\'n}czuk, Preslav Nakov, Maciej Ogrodniczuk, Senja Pollak, Pavel P{\v{r}}ib{\'a}{\v{n}}, Piotr Rybak, Josef Steinberger, and Roman Yangarber.
\newblock Proceedings of the 9th workshop on slavic natural language processing 2023 (slavicnlp 2023).
\newblock In {\em Proceedings of the 9th Workshop on Slavic Natural Language Processing 2023 (SlavicNLP 2023)}, 2023.

\bibitem{piskorski2025proceedings}
Jakub Piskorski, Pavel P{\v{r}}ib{\'a}{\v{n}}, Preslav Nakov, Roman Yangarber, and Micha{\l} Marci{\'n}czuk.
\newblock Proceedings of the 10th workshop on slavic natural language processing (slavic nlp 2025).
\newblock In {\em Proceedings of the 10th Workshop on Slavic Natural Language Processing (Slavic NLP 2025)}, 2025.

\bibitem{zhang2025qwen3}
Yanzhao Zhang, Mingxin Li, Dingkun Long, Xin Zhang, Huan Lin, Baosong Yang, Pengjun Xie, An~Yang, Dayiheng Liu, Junyang Lin, et~al.
\newblock Qwen3 embedding: Advancing text embedding and reranking through foundation models.
\newblock {\em arXiv preprint arXiv:2506.05176}, 2025.

\bibitem{babakhin2025llama}
Yauhen Babakhin, Radek Osmulski, Ronay Ak, Gabriel Moreira, Mengyao Xu, Benedikt Schifferer, Bo~Liu, and Even Oldridge.
\newblock Llama-embed-nemotron-8b: A universal text embedding model for multilingual and cross-lingual tasks.
\newblock {\em arXiv preprint arXiv:2511.07025}, 2025.

\bibitem{poswiata2026pl}
Rafa{\l} Po{\'s}wiata, S{\l}awomir Dadas, and Micha{\l} Pere{\l}kiewicz.
\newblock Pl-mteb: Polish massive text embedding benchmark.
\newblock In {\em Findings of the Association for Computational Linguistics: ACL 2026}, pages 35601--35619, 2026.

\bibitem{enevoldsen2024scandinavian}
Kenneth Enevoldsen, M{\'a}rton Kardos, Niklas Muennighoff, and Kristoffer~L Nielbo.
\newblock The scandinavian embedding benchmarks: Comprehensive assessment of multilingual and monolingual text embedding.
\newblock {\em Advances in Neural Information Processing Systems}, 37:40336--40358, 2024.

\bibitem{joshi2020state}
Pratik Joshi, Sebastin Santy, Amar Budhiraja, Kalika Bali, and Monojit Choudhury.
\newblock The state and fate of linguistic diversity and inclusion in the nlp world.
\newblock In {\em Proceedings of the 58th annual meeting of the association for computational linguistics}, pages 6282--6293, 2020.

\bibitem{klemen2024si}
Matej Klemen, Ale{\v{s}} {\v{Z}}agar, Jaka {\v{C}}ibej, and Marko Robnik-{\v{S}}ikonja.
\newblock Si-nli: A slovene natural language inference dataset and its evaluation.
\newblock In {\em Proceedings of the 2024 Joint International Conference on Computational Linguistics, Language Resources and Evaluation (LREC-COLING 2024)}, pages 14859--14870, 2024.

\bibitem{knez2025semi}
Timotej Knez, Miha {\v{S}}travs, and Slavko {\v{Z}}itnik.
\newblock Semi-supervised relation extraction corpus construction and models creation for under-resourced languages: A use case for slovene.
\newblock {\em Information}, 16(2):143, 2025.

\bibitem{piskorski2024cross}
Jakub Piskorski, Micha{\l} Marci{\'n}czuk, and Roman Yangarber.
\newblock Cross-lingual named entity corpus for slavic languages.
\newblock In {\em Proceedings of the 2024 Joint International Conference on Computational Linguistics, Language Resources and Evaluation (LREC-COLING 2024)}, pages 4143--4157, 2024.

\bibitem{suppa2025sklep}
Marek Suppa, Andrej Ridzik, Daniel Hl{\'a}dek, Tom{\'a}{\v{s}} Javurek, Vikt{\'o}ria Ondrejov{\'a}, Krist{\'\i}na S{\'a}sikov{\'a}, Martin Tamajka, and Marian Simko.
\newblock sklep: A slovak general language understanding benchmark.
\newblock In {\em Findings of the Association for Computational Linguistics: ACL 2025}, pages 26716--26743, 2025.

\bibitem{koeva2012bulgarian}
Svetla Koeva, Ivelina Stoyanova, Svetlozara Leseva, Rositsa Dekova, Tsvetana Dimitrova, and Ekaterina Tarpomanova.
\newblock The bulgarian national corpus: Theory and practice in corpus design.
\newblock {\em Journal of Language Modelling}, (1):65--110, 2012.

\bibitem{schober2018correlation}
Patrick Schober, Christa Boer, and Lothar~A Schwarte.
\newblock Correlation coefficients: appropriate use and interpretation.
\newblock {\em Anesthesia \& analgesia}, 126(5):1763--1768, 2018.

\bibitem{chen1992fuzzy}
Shu-Jen Chen and Ching-Lai Hwang.
\newblock Fuzzy multiple attribute decision making methods.
\newblock In {\em Fuzzy multiple attribute decision making: Methods and applications}, pages 289--486. Springer, 1992.

\bibitem{opricovic1998multicriteria}
Serafim Opricovic.
\newblock Multicriteria optimization of civil engineering systems.
\newblock {\em Faculty of civil engineering, Belgrade}, 2(1):5--21, 1998.

\bibitem{brans1982ingenierie}
Jean-Pierre Brans, R~Nadeau, and M~Landry.
\newblock L’ing{\'e}nierie de la d{\'e}cision.
\newblock {\em Elaboration d’instruments d’aide {\`a} la d{\'e}cision. La m{\'e}thode PROMETHEE. In l’Aide {\`a} la D{\'e}cision: Nature, Instruments et Perspectives d’Avenir}, pages 183--213, 1982.

\bibitem{keshavarz2021determination}
Mehdi Keshavarz-Ghorabaee, Maghsoud Amiri, Edmundas~Kazimieras Zavadskas, Zenonas Turskis, and Jurgita Antucheviciene.
\newblock Determination of objective weights using a new method based on the removal effects of criteria (merec).
\newblock {\em Symmetry}, 13(4):525, 2021.

\bibitem{diakoulaki1995determining}
Danae Diakoulaki, George Mavrotas, and Lefteris Papayannakis.
\newblock Determining objective weights in multiple criteria problems: The critic method.
\newblock {\em Computers \& operations research}, 22(7):763--770, 1995.

\end{thebibliography}

\appendix
 
\section{ESS Design Choices}
\label{appendix:ess}
 
The ESS was designed to quantify the reliability of the available benchmark evidence rather than model performance itself. The motivation behind ESS is that stability conclusions in low-resource multilingual settings are strongly affected by benchmark sparsity, dataset redundancy, and limited stability assessability. Consequently, apparently stable benchmark rankings may arise from insufficient evaluation coverage rather than genuine model stability.
 
The first design component, dataset availability, measures the amount of available evaluation evidence for a task-language pair. Rather than using the raw number of datasets directly, we normalize using a saturation threshold~$n_{\min}$. This reflects the assumption that additional datasets provide diminishing returns once a sufficiently diverse evaluation setting has been reached. Without saturation, high-resource languages would dominate the score purely due to dataset quantity. We set $n_{\min}=5$ for the analyzed benchmark snapshot, treating five or more datasets as providing saturated availability for reliable stability assessment. This choice has been further analyzed in~\ref{appendix:ess_saturation}.
 
The second component, effective dataset diversity, addresses the fact that multiple datasets do not necessarily correspond to independent evaluation evidence. Datasets with highly correlated model rankings may provide nearly identical information. We therefore cluster datasets based on similarity of model-performance profiles and quantify diversity as the ratio between decorrelated dataset clusters and the total number of datasets. This allows ESS to distinguish between genuinely diverse evaluation settings and benchmark collections dominated by redundant datasets.
 
The final two components quantify stability assessability. The aggregation stability indicator measures whether multiple ranking schemes can be compared, while the composition stability indicator measures whether multiple non-redundant dataset compositions exist. These components explicitly capture whether stability analysis is feasible for a given task-language pair.
 
The final ESS formulation combines these four components equally. Equal weighting was selected to preserve interpretability and avoid introducing additional assumptions regarding the relative importance of dataset quantity, diversity, and stability assessability. The resulting score therefore provides a normalized estimate of how reliable benchmark conclusions are for a given task-language pair. The ESS component weighting has been further analyzed in~\ref{appendix:ess_wighting}.
 
Importantly, ESS is not intended as a replacement for stability metrics such as Kendall's coefficient of concordance or top-$k$ consistency. Instead, it serves as a heuristic confidence measure for interpreting those results. High stability together with high ESS indicates that conclusions are supported by sufficient and diverse benchmark evidence, whereas high stability under low ESS suggests that apparent stability may primarily reflect sparse or redundant evaluation conditions.

\section{Interpretation of ESS Values}
\label{appendix:interpretationESS}

The ESS should be interpreted as a heuristic confidence measure in the stability conclusions rather than a direct measure of model quality. Since ESS jointly captures benchmark coverage, dataset diversity, and stability assessability, higher values indicate that ranking stability and consistency results are supported by broader and more reliable evaluation evidence. Table~\ref{tab:ess_interpretation} summarizes the proposed interpretation ranges.
 
\begin{table}[t]
\centering
\small
\caption{Interpretation of Evidence Strength Score~(ESS) values.}
\label{tab:ess_interpretation}
\begin{tabular}{ll}
\hline
ESS Range & Interpretation \\
\hline
$[0.0,0.1)$   & Extremely weak evidence \\
$[0.1,0.25)$  & Weak evidence \\
$[0.25,0.4)$  & Moderate evidence \\
$[0.4,0.6)$   & Comparatively strong evidence \\
$[0.6,1.0]$   & Very strong evidence \\
\hline
\end{tabular}
\end{table}
 
Values below $0.1$ typically correspond to missing benchmark coverage or highly sparse evaluation settings, where stability conclusions should be interpreted with substantial caution.
Scores between $0.1$ and $0.25$ indicate weak evidence, usually reflecting isolated datasets or limited stability assessability. Values between $0.25$ and $0.4$ correspond to moderate evidence, where at least partial stability analysis is feasible but benchmark coverage remains incomplete. Scores above $0.4$ indicate comparatively strong evidence under
current multilingual benchmark conditions, while values above $0.6$ correspond to tasks or languages with relatively rich, diverse, and stability-assessable evaluation resources.
 
Importantly, the proposed ESS formulation intentionally penalizes missing tasks, dataset redundancy, and limited stability assessability. Consequently, even the strongest-performing languages in current multilingual benchmarks may still achieve only moderate ESS values. For example, Russian achieves the highest language-level ESS in our analysis~($0.49$), which should therefore be interpreted as comparatively strong evidence relative to the current benchmark landscape rather than near-saturated evaluation coverage. This observation itself highlights the substantial benchmark sparsity that characterizes multilingual evaluation for many Slavic languages.

\section{Language-Specific Cross-Task Analysis}
\label{appendix:cross_task_detailed}
 
In the language-specific cross-task analysis, we evaluate consistency across all eight tasks for each language. Since not all tasks support composition stability analysis, conclusions primarily capture stability with respect to aggregation methods. We set $k=10$ and compute cross-task consistency by ranking scheme and the coverage weighted aggregation. The cross-task consistencies by ranking scheme are visualized as dendrograms, where rows correspond to models and columns to ranking schemes. Based on the heatmaps, we identify the most consistent models per language. Here we show the dendrograms for the language with highest task coverage by Slavic language subgroup. The dendrograms for all other languages are available in the project repository, which can be used to perform the same language-specific analysis presented below. Additionally, the coverage-weighted cross-task consistency scores for the same set of models shown in the dendrograms are given as bar charts.
 
\subsection{Western South Slavic Languages}
 
\paragraph{Slovenian (SLV)}

\begin{figure*}[!ht]
  \includegraphics[width=0.8\linewidth]{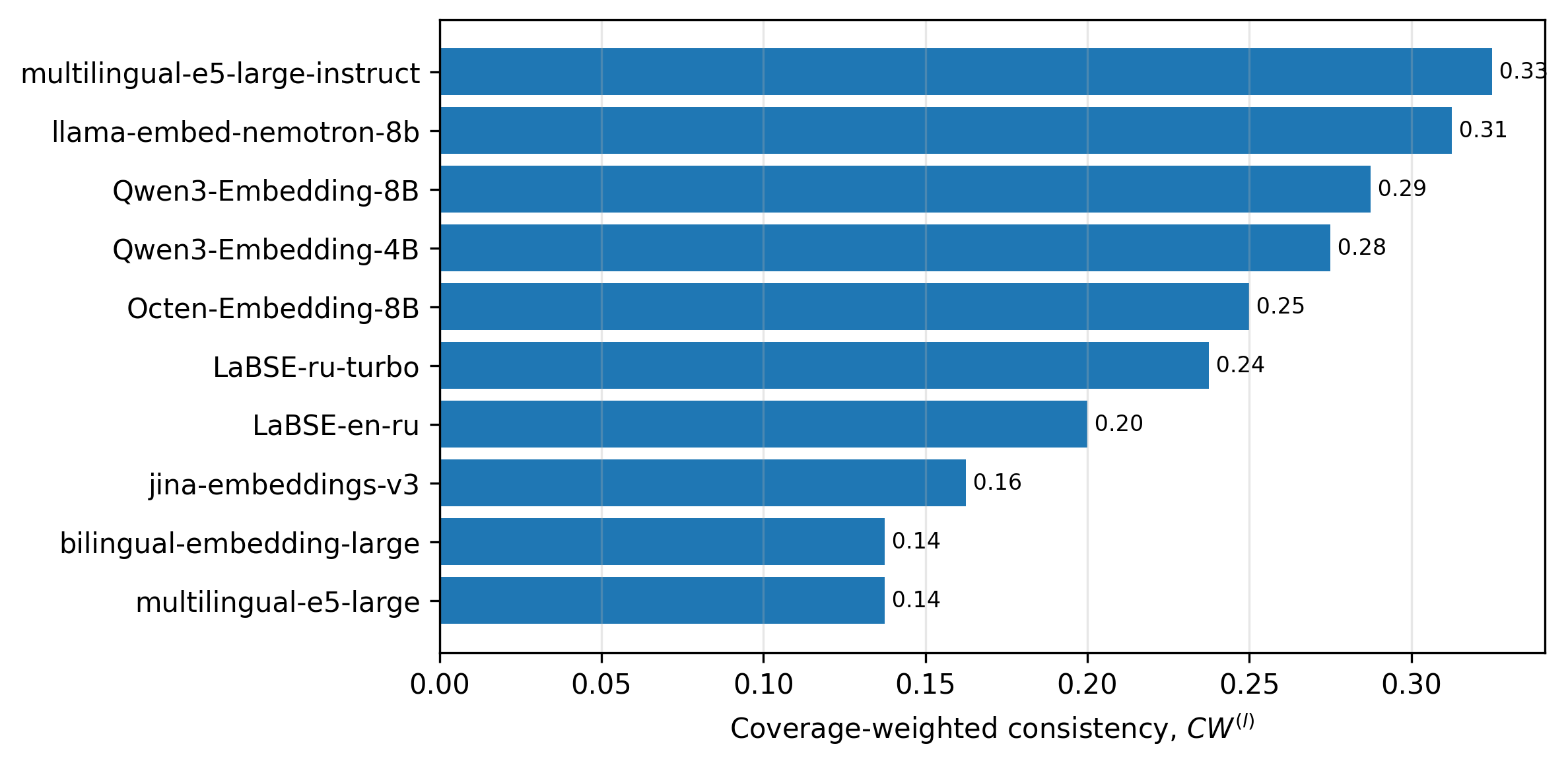}
  \caption{Coverage-weighted cross-task consistency for Slovenian~($k=10$).}
  \label{fig:slv-top-10-cross-task-consistency_bar}
\end{figure*}

\begin{figure*}[!ht]
  \includegraphics[width=\linewidth]{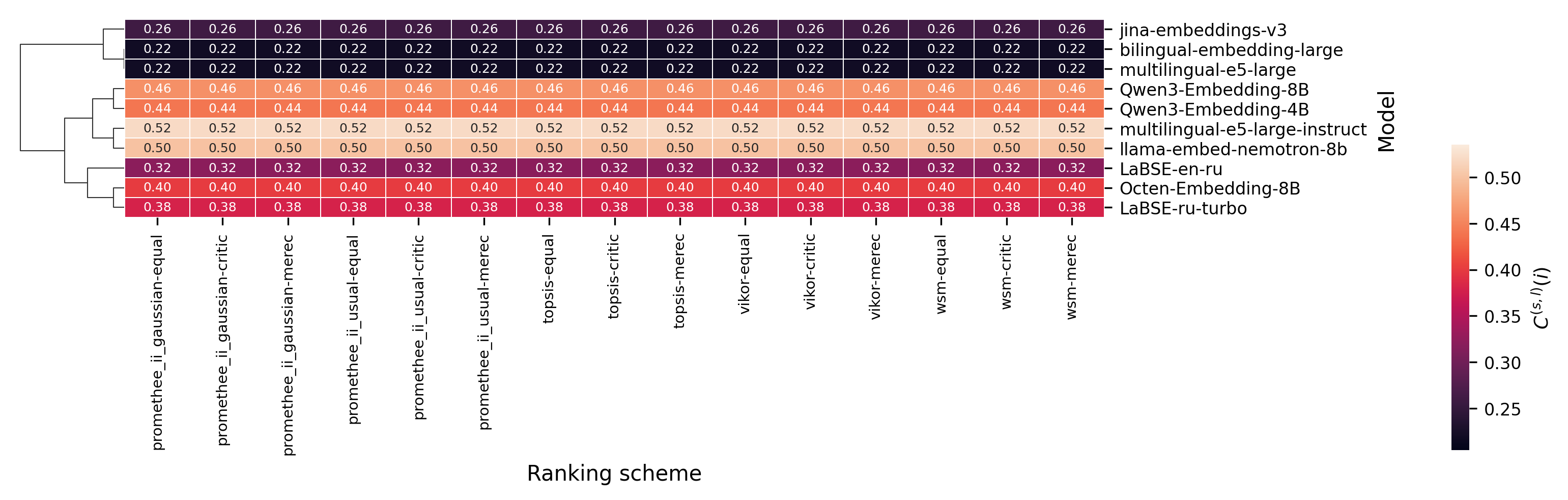}
  \caption{Cross-task consistency by ranking scheme for Slovenian~($k=10$).}
  \label{fig:slv-top-10-cross-task-consistency}
\end{figure*}

Although Slovenian has benchmark data for only five of the eight analyzed tasks, a clear group of highly consistent multilingual models emerges. The highest coverage-weighted cross-task consistency is achieved by \textit{multilingual-e5-large-instruct} ($CW=0.33$), closely followed by \textit{llama-embed-nemotron-8b} ($CW=0.31$). A second group of highly competitive models consists of \textit{Qwen3-Embedding-8B} ($CW=0.29$), \textit{Qwen3-Embedding-4B} ($CW=0.28$), and \textit{Octen-Embedding-8B} ($CW=0.25$) (Figure~\ref{fig:slv-top-10-cross-task-consistency_bar}). The dendrogram (Figure~\ref{fig:slv-top-10-cross-task-consistency}) further demonstrates that the relative ordering of the leading models is highly stable across all 15 ranking schemes. Overall, the results indicate that Slovenian cross-task consistency is characterized by a small group of large multilingual embedding models that consistently outperform the remaining candidates regardless of the ranking methodology. Nevertheless, the language-level evidence remains limited ($ESS=0.18$), and the conclusions should therefore be interpreted with appropriate caution.

\subsection{Eastern South Slavic Languages}
 
\paragraph{Bulgarian (BUL)}

\begin{figure*}[!ht]
  \includegraphics[width=0.8\linewidth]{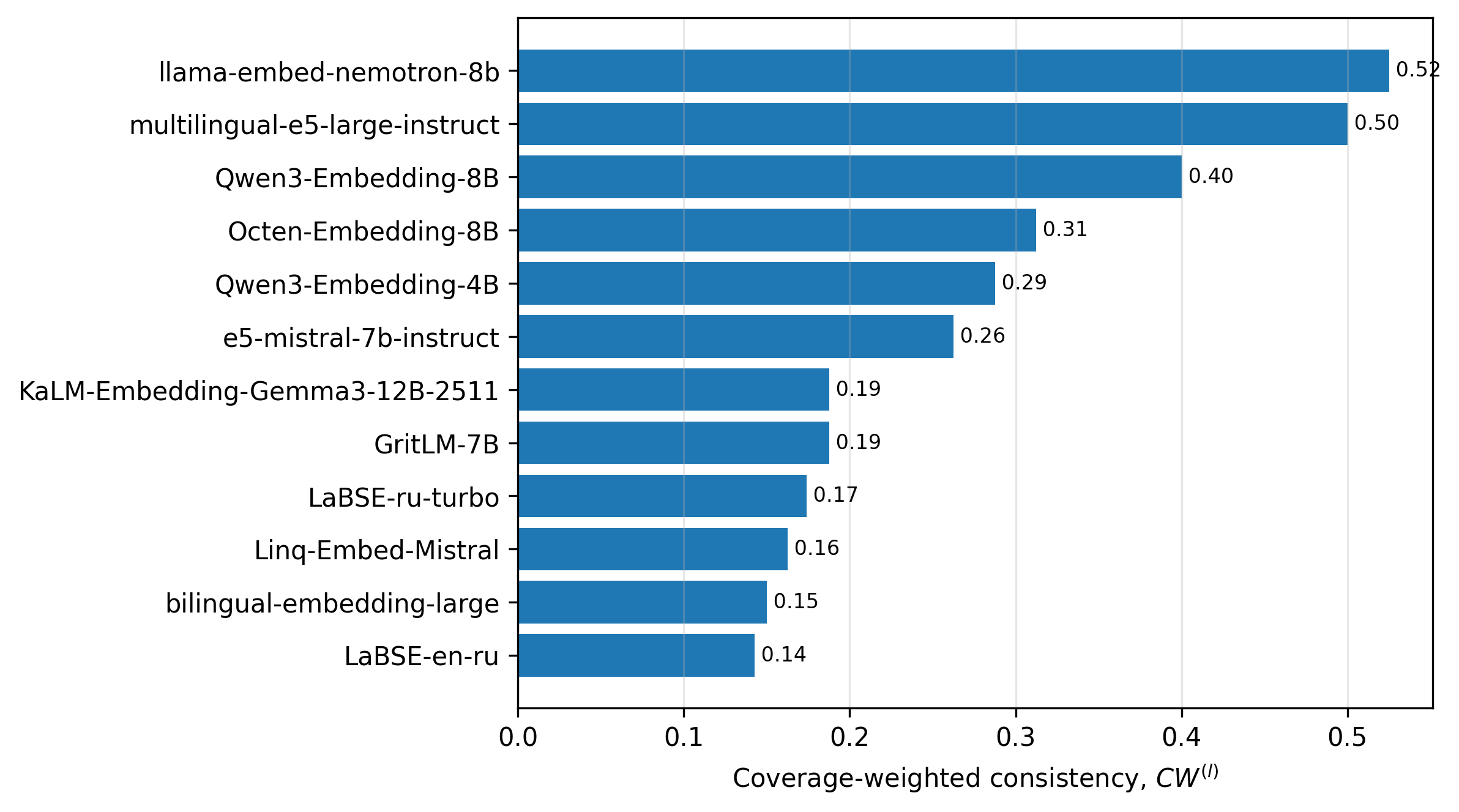}
  \caption{Coverage-weighted cross-task consistency for Bulgarian~($k=10$).}
  \label{fig:bul-top-10-cross-task-consistency_bar}
\end{figure*}

\begin{figure*}[!ht]
  \includegraphics[width=\linewidth]{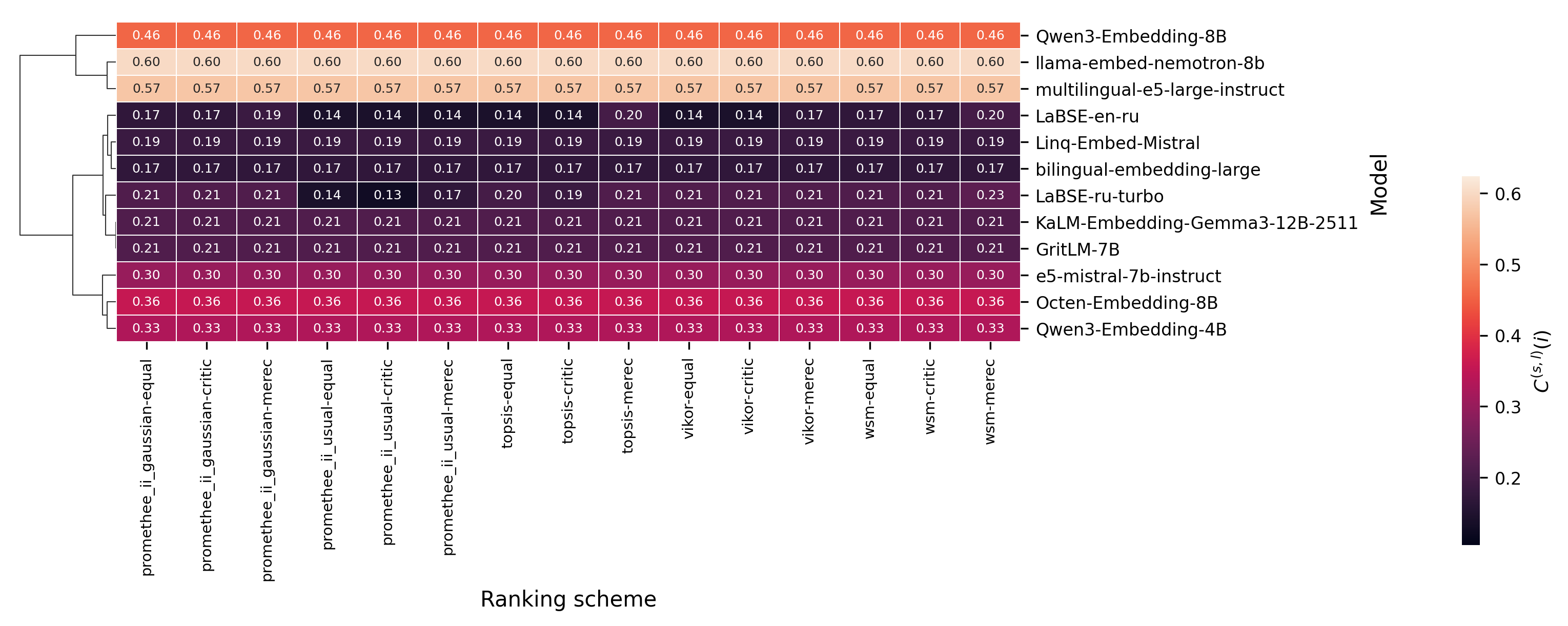}
  \caption{Cross-task consistency by ranking scheme for Bulgarian~($k=10$).}
  \label{fig:bul-top-10-cross-task-consistency}
\end{figure*}

With benchmark data available for seven of the eight analyzed tasks, Bulgarian provides one of the strongest settings for cross-task consistency analysis. A clear leading pair emerges, consisting of \textit{llama-embed-nemotron-8b} ($CW=0.52$) and \textit{multilingual-e5-large-instruct} ($CW=0.50$), which achieve nearly identical coverage-weighted consistency scores and are substantially ahead of the remaining models. \textit{Qwen3-Embedding-8B} forms a strong second tier ($CW=0.40$), followed by \textit{Octen-Embedding-8B} ($CW=0.31$), \textit{Qwen3-Embedding-4B} ($CW=0.29$), and \textit{e5-mistral-7b-instruct} ($CW=0.26$), which also demonstrate competitive cross-task performance (Figure~\ref{fig:bul-top-10-cross-task-consistency_bar}). The dendrogram (Figure~\ref{fig:bul-top-10-cross-task-consistency}) shows that the leading models exhibit stable consistency scores across all 15 ranking schemes, with only minor numerical fluctuations that do not affect the overall ordering. In particular, \textit{llama-embed-nemotron-8b} and \textit{multilingual-e5-large-instruct} consistently occupy the top positions, followed by \textit{Qwen3-Embedding-8B}, while the remaining models preserve a similar relative ranking across aggregation methods. Overall, the results indicate that Bulgarian cross-task performance is dominated by a small group of large multilingual embedding models, led by \textit{llama-embed-nemotron-8b} and \textit{multilingual-e5-large-instruct}, with \textit{Qwen3-Embedding-8B} representing the strongest alternative. The ESS score of 0.29 corresponds to a moderate level of evidence.

\subsection{West Slavic Languages}
 
\paragraph{Polish (POL)}

\begin{figure*}[!ht]
  \includegraphics[width=0.8\linewidth]{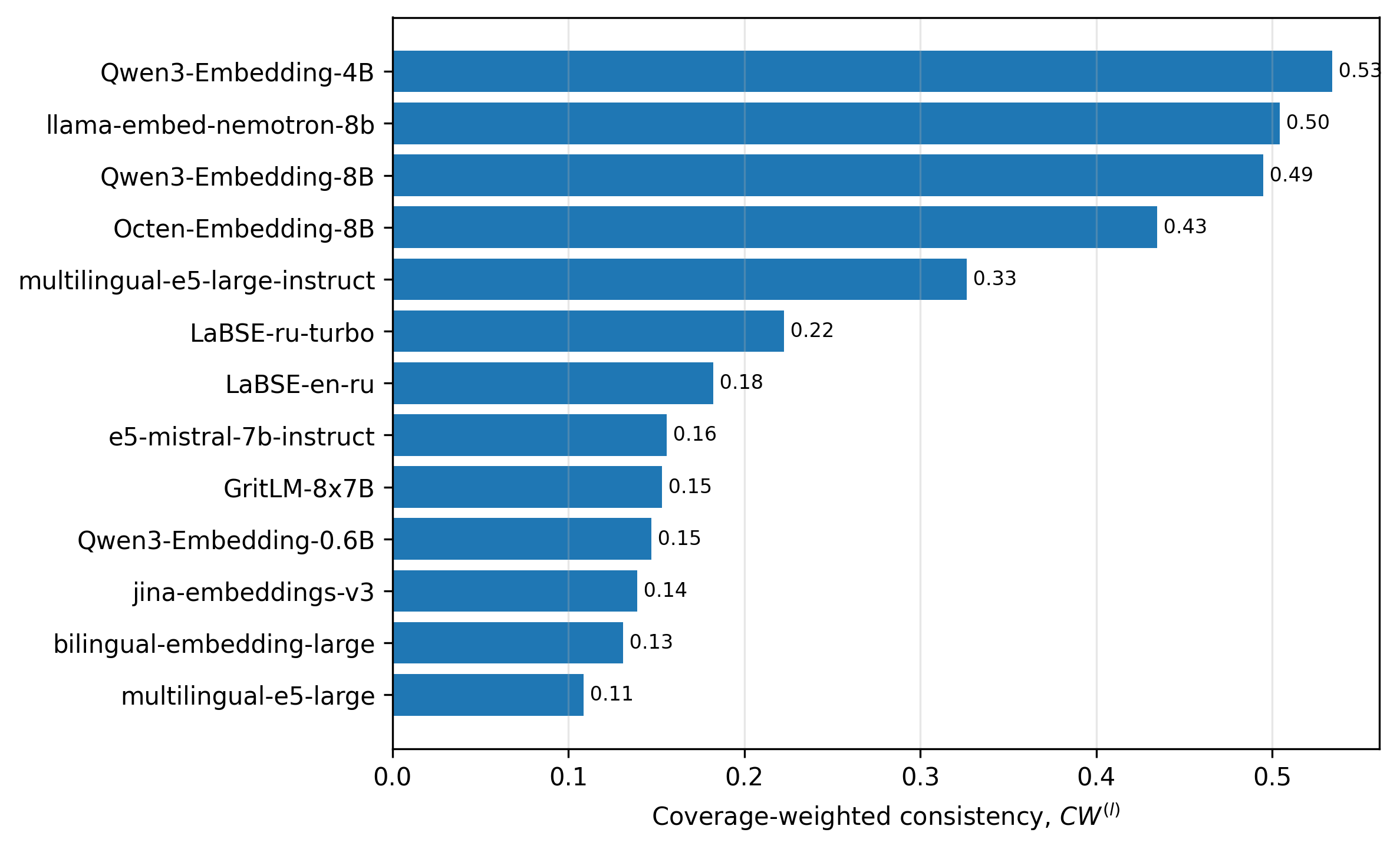}
  \caption{Coverage-weighted cross-task consistency for Polish~($k=10$).}
  \label{fig:pol-top-10-cross-task-consistency_bar}
\end{figure*}

\begin{figure*}[!ht]
  \includegraphics[width=\linewidth]{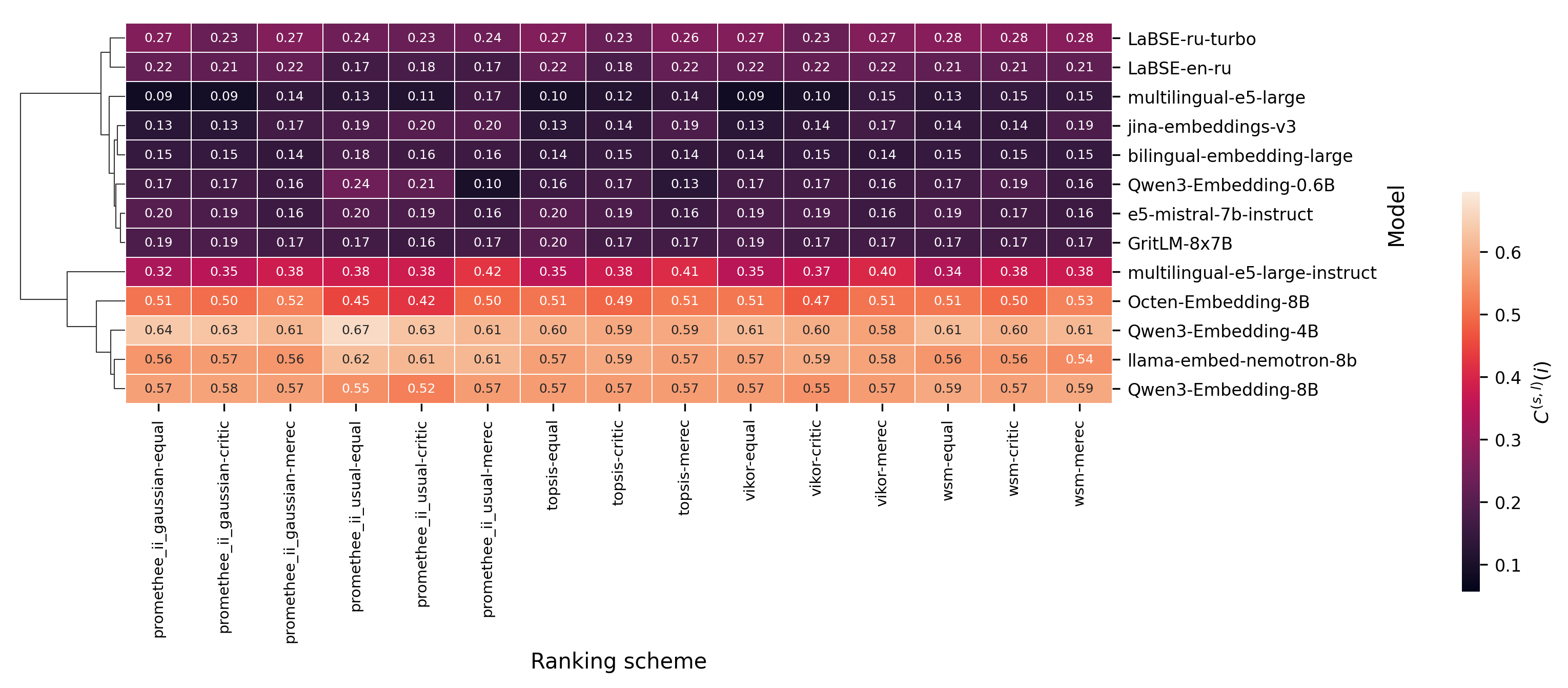}
  \caption{Cross-task consistency by ranking scheme for Polish~($k=10$).}
  \label{fig:pol-top-10-cross-task-consistency}
\end{figure*}

With benchmark data available for seven of the eight analyzed tasks, Polish provides one of the strongest settings for evaluating cross-task consistency. The highest coverage-weighted consistency is achieved by \textit{Qwen3-Embedding-4B} ($CW=0.53$), followed closely by \textit{llama-embed-nemotron-8b} and \textit{Qwen3-Embedding-8B} (both $CW=0.50$), with \textit{Octen-Embedding-8B} ($CW=0.43$) completing a clearly defined leading group (Figure~\ref{fig:pol-top-10-cross-task-consistency_bar}). \textit{multilingual-e5-large-instruct} forms a second tier ($CW=0.33$), while the remaining models exhibit substantially lower cross-task consistency. The dendrogram (Figure~\ref{fig:pol-top-10-cross-task-consistency}) shows that the leading models maintain highly consistent scores across all 15 ranking schemes, with only minor numerical fluctuations that do not alter their relative ordering. In particular, \textit{Qwen3-Embedding-4B}, \textit{llama-embed-nemotron-8b}, and \textit{Qwen3-Embedding-8B} remain the dominant models irrespective of the aggregation method, demonstrating robust performance across tasks. Overall, the results indicate that Polish exhibits one of the strongest and most competitive benchmark landscapes, where several large multilingual embedding models achieve consistently high cross-task consistency under diverse ranking methodologies. The ESS score of 0.42 corresponds to a comparatively strong level of evidence, indicating that these conclusions are supported by a broad coverage of benchmark tasks.
 
\subsection{East Slavic Languages}
 
\paragraph{Russian (RUS)}

\begin{figure*}[!ht]
  \includegraphics[width=0.8\linewidth]{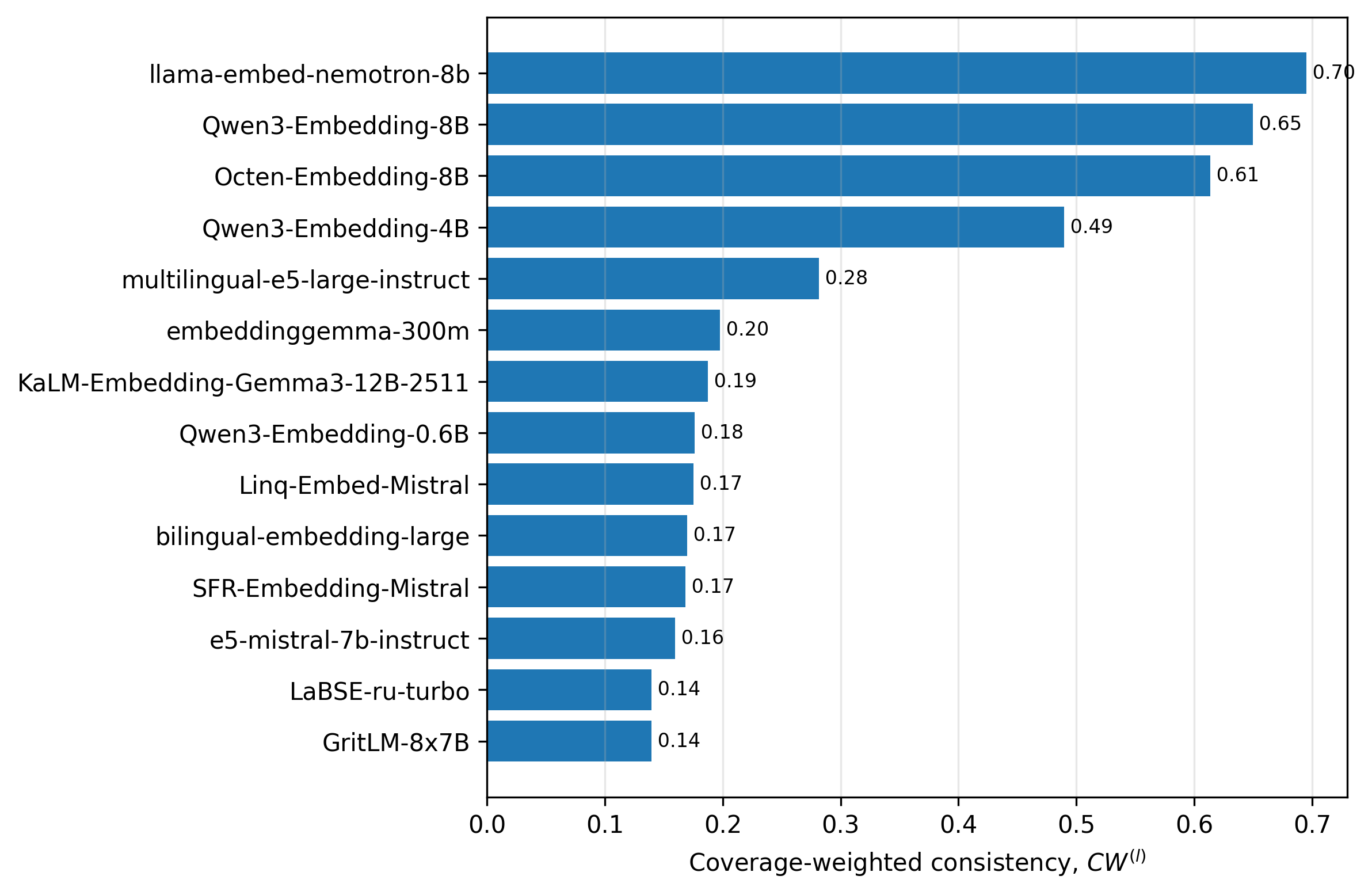}
  \caption{Coverage-weighted cross-task consistency for Russian~($k=10$).}
  \label{fig:rus-top-10-cross-task-consistency_bar}
\end{figure*}

\begin{figure*}[!ht]
  \includegraphics[width=\linewidth]{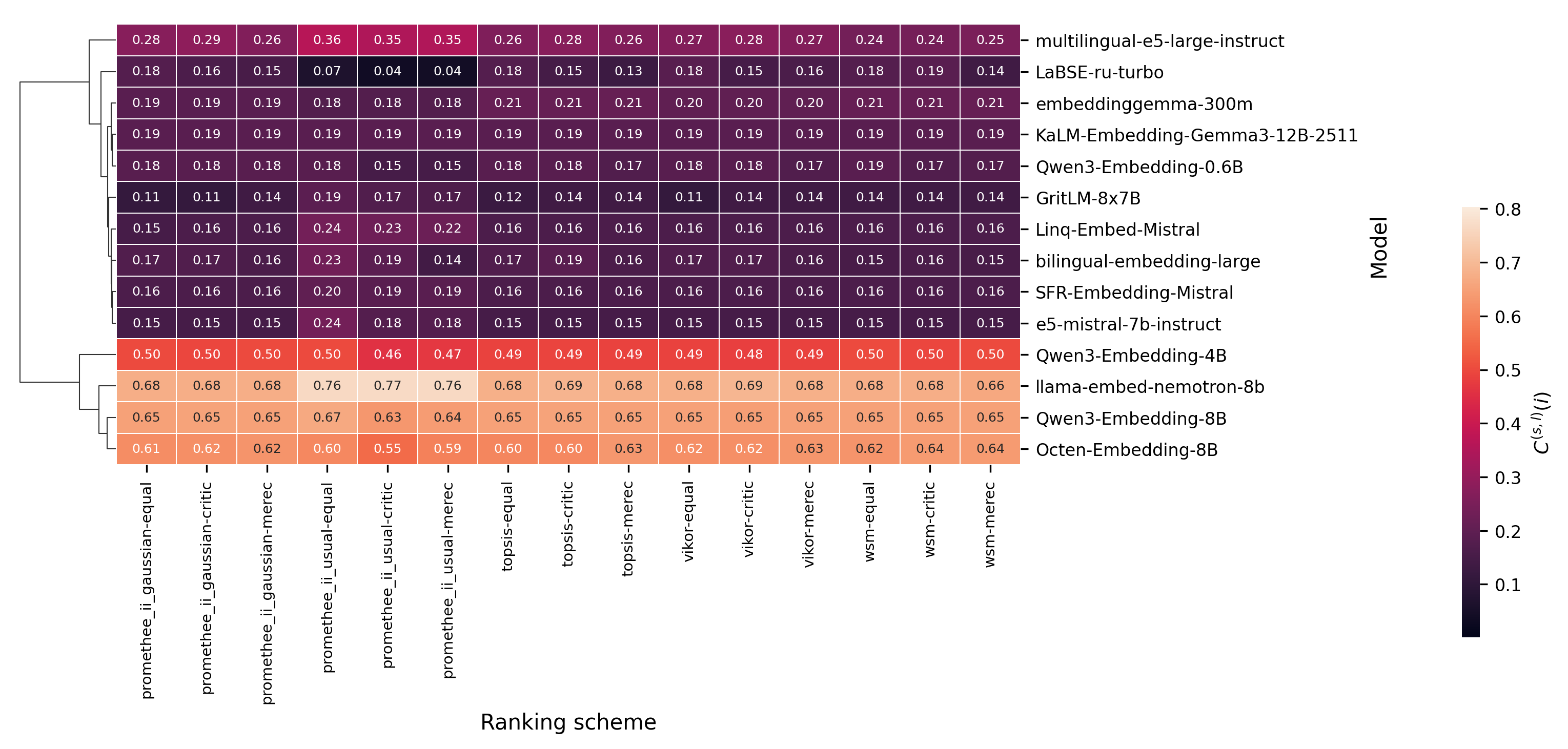}
  \caption{Cross-task consistency by ranking scheme for Russian~($k=10$).}
  \label{fig:rus-top-10-cross-task-consistency}
\end{figure*}

As the only language with complete benchmark coverage across all eight analyzed tasks, Russian provides the strongest evidence for evaluating cross-task consistency. \textit{Llama-embed-nemotron-8b} achieves the highest coverage-weighted consistency score ($CW=0.70$), followed closely by \textit{Qwen3-Embedding-8B} ($CW=0.65$) and \textit{Octen-Embedding-8B} ($CW=0.61$), while \textit{Qwen3-Embedding-4B} ($CW=0.49$) completes a clearly defined leading group of highly consistent models (Figure~\ref{fig:rus-top-10-cross-task-consistency_bar}). \textit{Multilingual-e5-large-instruct} ($CW=0.28$) forms a second tier below these four models. The dendrogram (Figure~\ref{fig:rus-top-10-cross-task-consistency}) shows that the leading models maintain highly consistent scores across all 15 ranking schemes, with only minor numerical variations that do not alter their relative ordering. In particular, \textit{llama-embed-nemotron-8b}, \textit{Qwen3-Embedding-8B}, and \textit{Octen-Embedding-8B} consistently occupy the top positions under every aggregation method, while \textit{Qwen3-Embedding-4B} remains a stable fourth-ranked model. Overall, the results indicate that Russian exhibits the strongest and most reliable cross-task consistency among the analyzed languages, supported by complete benchmark coverage and a clear separation between the leading multilingual embedding models and the remaining approaches. The ESS score of 0.49 corresponds to a comparatively strong level of evidence and confidence in the robustness of these conclusions.

\section{Sensitivity of ESS to Correlation and Dataset Saturation Thresholds}
\label{appendix:ess_saturation}

Sensitivity analysis were conducted to examine the dependence of the ESS values, calculated by language and task pair, on two methodological parameters, i.e., (1) the correlation threshold \(\tau\), used to identify groups of correlated datasets, and (2) the dataset saturation threshold \(n_{\min}\), used as part of the ESS formula. We considered three correlation thresholds, \(\tau\in\{0.7,0.8,0.9\}\), and six saturation thresholds, \(n_{\min}\in\{5,6,7,8,9,10\}\), which yielded 18 parameter configurations. The original configuration, \(\tau=0.9\) and \(n_{\min}=5\), served as a reference. For each configuration, ESS was recalculated for all 96 task–language combinations using the same four equally weighted components, i.e., dataset availability, dataset diversity, ranking-scheme stability assessability, and dataset-composition stability assessability. Changing \(n_{\min}\) affects only the availability component, whereas changing \(\tau\) may alter the number of dataset clusters (diversity component).

\begin{figure}[!th]
    \centering

    \begin{subfigure}[b]{0.48\textwidth}
        \centering
        \includegraphics[width=\linewidth]{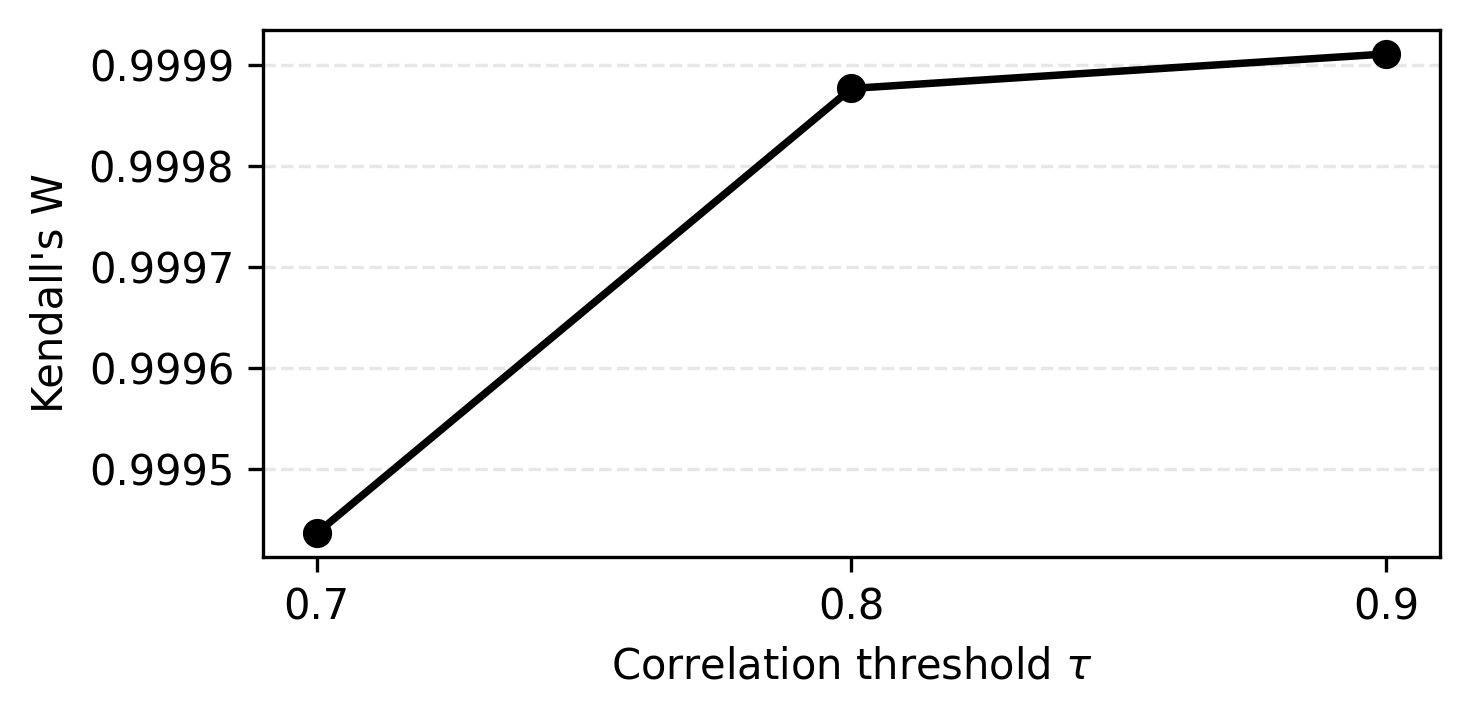}
        \label{fig:kendall_tau}
    \end{subfigure}
    \hfill
    \begin{subfigure}[b]{0.48\textwidth}
        \centering
        \includegraphics[width=\linewidth]{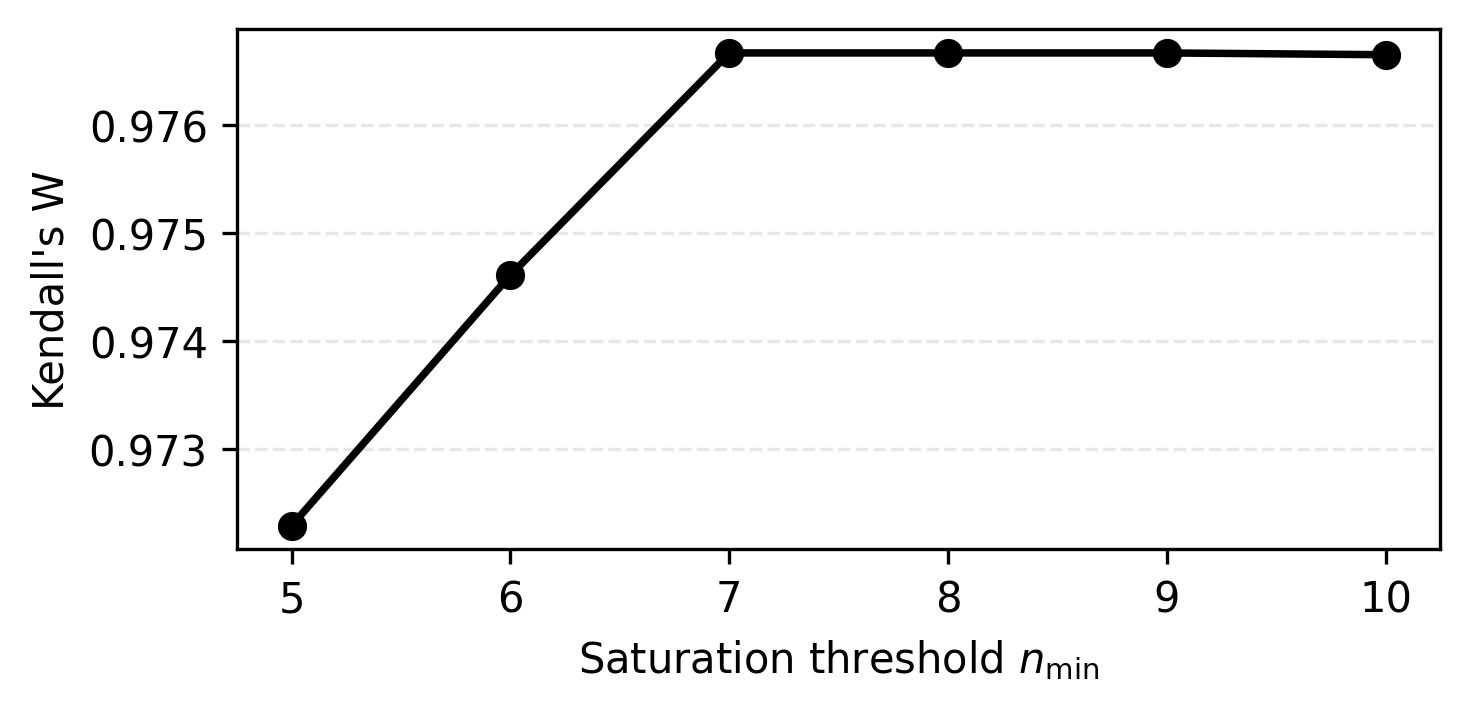}
        \label{fig:kendall_nmin}
    \end{subfigure}

    \caption{Agreement of ESS-based task-language orderings under alternative parameter configurations. Kendall's $W$ is calculated across six $n_{\min}$ values for each $\tau$ (left), and across three $\tau$ values for each $n_{\min}$ (right).}
    \label{fig:ess_sensitivity_kendall}
\end{figure}

To assess the stability of the relative ESS ordering of the task-language pairs, we calculated Kendall's coefficient of concordance (\(W\)). First, for each correlation threshold, we measured agreement among the six ESS orderings obtained with different \(n_{\min}\) values. The resulting coefficients were \(W=0.999911\) for \(\tau=0.9\), \(W=0.999877\) for \(\tau=0.8\), and \(W=0.999436\) for \(\tau=0.7\), as illustrated in Figure~\ref{fig:ess_sensitivity_kendall} (left). These values indicate near-perfect agreement in the relative ordering of task–language combinations across the examined saturation thresholds.

Second, for each fixed \(n_{\min}\), we assessed agreement among the three ESS orderings obtained with different correlation thresholds \(\tau\). Kendall's \(W\) ranged from 0.972295 at \(n_{\min}=5\) to approximately 0.9767 at larger saturation thresholds, as illustrated in Figure~\ref{fig:ess_sensitivity_kendall} (right). Although agreement remained high, it was consistently lower than the agreement observed when varying \(n_{\min}\) alone. This indicates that relative ESS ordering is more sensitive to the correlation threshold than to the dataset saturation threshold.

The component-level effects explain this difference. Increasing \(n_{\min}\) reduces or preserves the availability component without changing dataset counts, cluster assignments, or evidence levels. Consequently, absolute ESS values may decrease while the relative ordering remains largely unchanged. In contrast, decreasing \(\tau\) can merge dataset clusters, reducing measured diversity and potentially changing the evidence level of a task–language combination. Such changes can affect multiple ESS components simultaneously. This explains the slightly lower Kendall's \(W\) values observed across correlation thresholds compared with those obtained across saturation thresholds. Overall, the results demonstrate that the relative ordering of task-language combinations by ESS is highly stable across the examined parameter configurations, particularly with respect to \(n_{\min}\). The full heatmaps illustrating the change in the ESS value by language-task pair, \(n_{\min}\) value, and \(\tau\) are available in the project repository.

To further investigate the sensitivity of the ESS to the correlation threshold \(\tau\), we decomposed changes in ESS into contributions from its four components: dataset availability (\(A\)), dataset diversity (\(Div\)), ranking-scheme stability availability (\(RS\)), and dataset-selection stability availability (\(DS\)). Unlike Kendall's coefficient of concordance, which evaluates agreement in relative ESS ordering, this analysis identifies which components contribute most to changes in absolute ESS values.

For each task-language combination \(i\), the change in ESS between two correlation thresholds, \(\tau_a\) and \(\tau_b\), was expressed as their difference (Eq.~\ref{eq:ess_change}). The contribution of each component \(j\) was calculated as the difference between its values at the two thresholds, multiplied by its weight of \(1/4\) (Eq.~\ref{eq:ess_pair_change}). Here, \(x_{ij}(\tau)\) denotes the value of component \(j\) for task–language combination \(i\) at threshold \(\tau\), where \(j\in\{A,Div,RS,DS\}\). A positive contribution indicates that the component increases the ESS when moving from \(\tau_a\) to \(\tau_b\), whereas a negative contribution indicates a decrease. The sum of all four contributions therefore equals the total change in ESS for that task-language combination.

\begin{equation}
\label{eq:ess_change}
\Delta ESS_i^{a\rightarrow b}
=
ESS_i(\tau_b)-ESS_i(\tau_a)
=
\sum_{j\in\{A,Div,RS,DS\}}
C_{ij}^{a\rightarrow b}
\end{equation}

\begin{equation}
\label{eq:ess_pair_change}
C_{ij}^{a\rightarrow b}
=
\frac{x_{ij}(\tau_b)-x_{ij}(\tau_a)}{4}.
\end{equation}

We calculated two summary measures across the \(N=96\) task-language combinations. The mean signed contribution indicates whether a component increases or decreases ESS on average, while the mean absolute contribution quantifies the magnitude of its contribution irrespective of direction. We evaluated three threshold comparisons: \(0.9\rightarrow0.8\), \(0.8\rightarrow0.7\), and \(0.9\rightarrow0.7\), keeping the dataset saturation threshold fixed at \(n_{\min}=5\).

\begin{figure}[!th]
    \centering

    \begin{subfigure}[b]{0.48\textwidth}
        \centering
        \includegraphics[width=\linewidth]{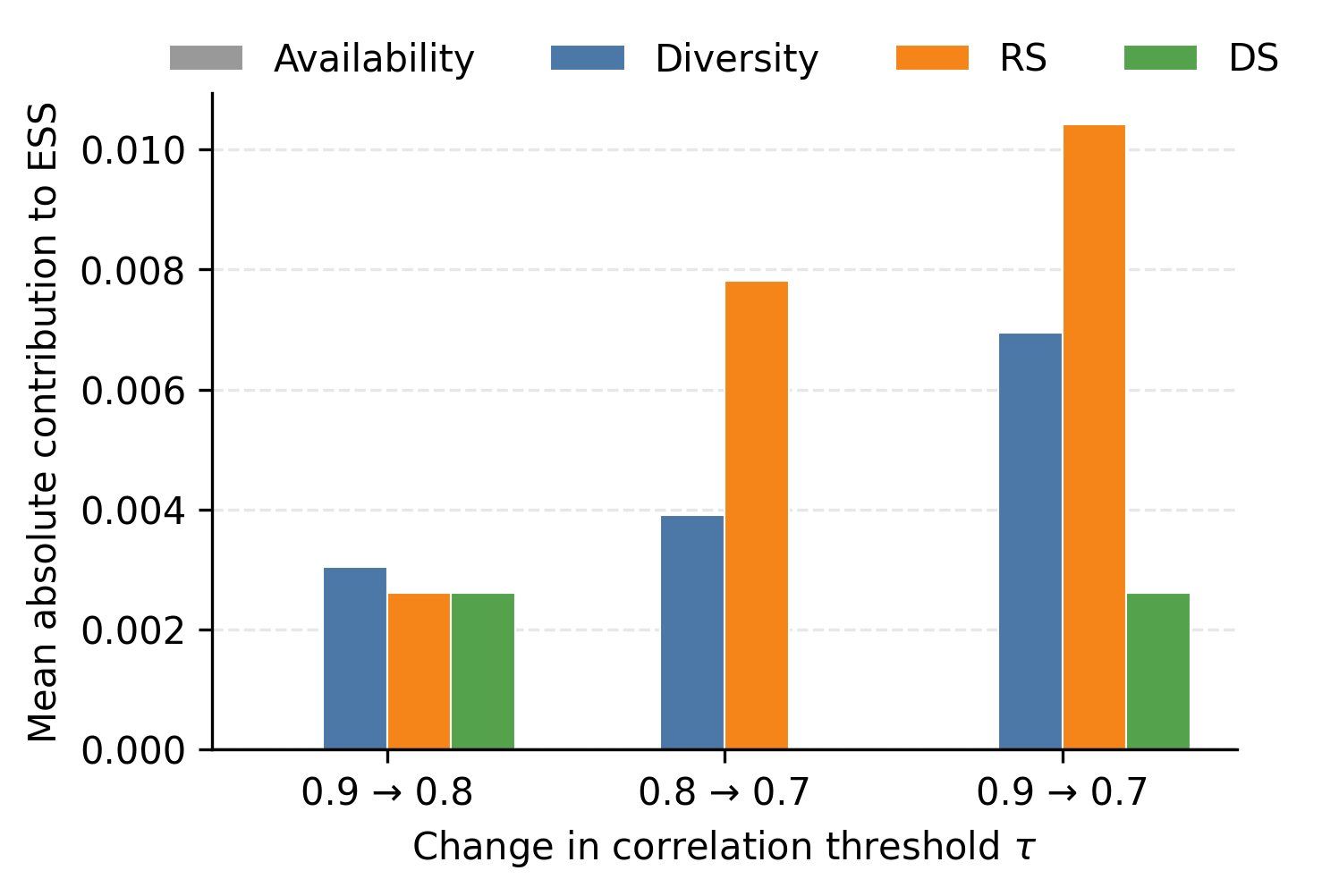}
    \end{subfigure}
    \hfill
    \begin{subfigure}[b]{0.48\textwidth}
        \centering
        \includegraphics[width=\linewidth]{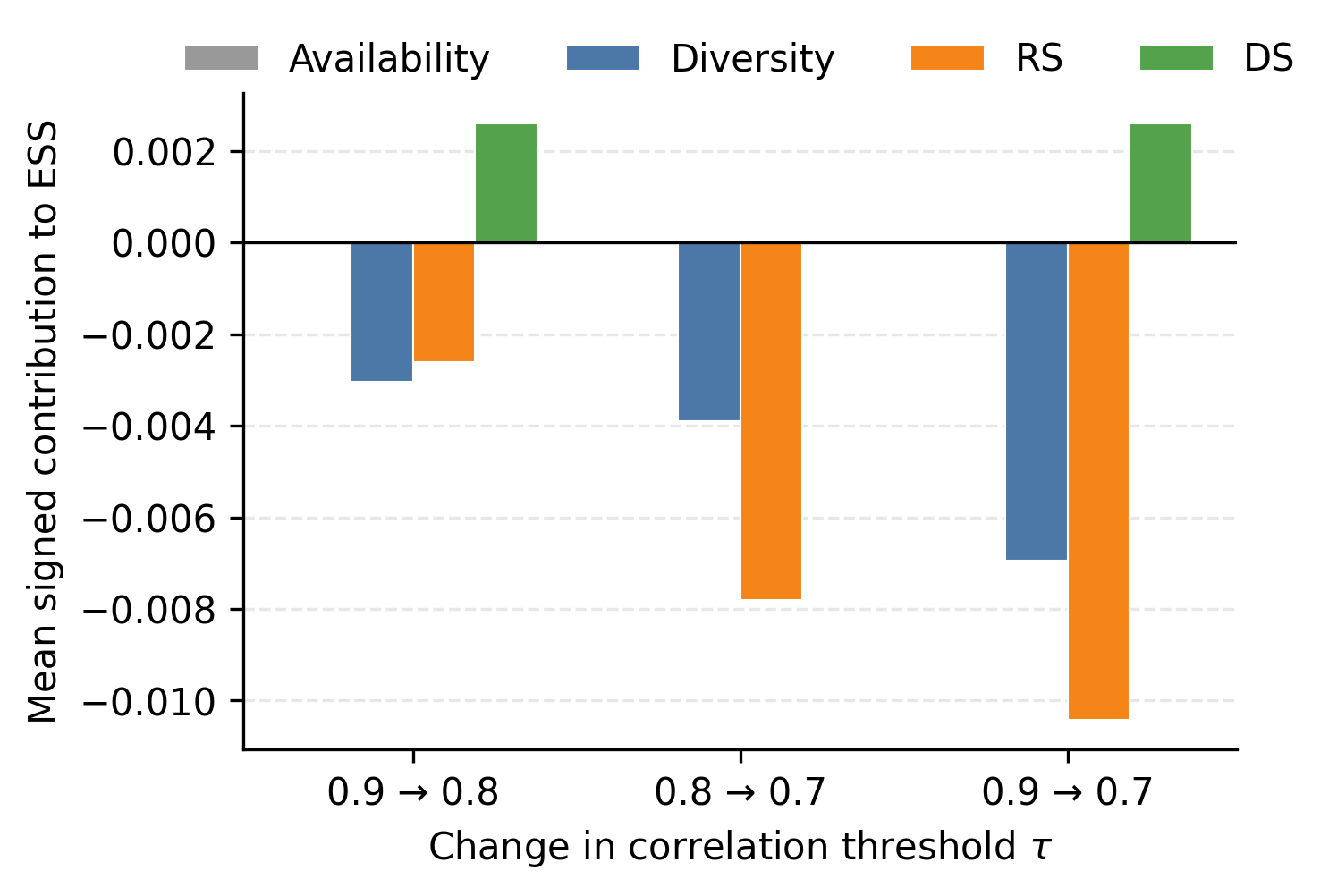}
    \end{subfigure}

    \caption{Component-wise contributions to changes in the ESS task-language score under alternative correlation thresholds \(\tau\). (left) Mean absolute contributions quantify the magnitude of component-level changes, irrespective of direction. (right) Mean signed contributions indicate whether each component increases or decreases ESS. Contributions are averaged across 96 task-language combinations using equal component weights and a fixed saturation threshold of \(n_{\min}=5\).}
    \label{fig:ess_changes}
\end{figure}

Figure~\ref{fig:ess_changes} presents the mean absolute and signed contributions for each comparison. Dataset availability remained unchanged across all thresholds, as the number of datasets is independent of \(\tau\). In contrast, dataset diversity consistently decreased as the correlation threshold was lowered, reflecting reductions in the number of independent dataset clusters.

Across the full threshold range (\(\tau=0.9\) to \(\tau=0.7\)), RS availability exhibited the largest mean absolute contribution (0.01042), followed by dataset diversity (0.00694) and DS availability (0.00260). Diversity decreased for five task–language combinations, whereas RS availability decreased for four combinations. Although diversity changes were more frequent, changes in the binary RS indicator produced larger individual contributions, reaching 0.25 in absolute magnitude. The signed contributions further reveal that RS availability and diversity both reduced ESS, with mean contributions of \(-0.01042\) and \(-0.00694\), respectively. Conversely, DS availability contributed \(+0.00260\), partially offsetting these reductions. This increase occurs because merging previously independent datasets into a correlated group can enable dataset-selection stability assessment, changing the evidence level from \(E_{RS}\) to \(E_{RS+DS}\). Consequently, the overall mean ESS decreased by approximately 0.01476 between \(\tau=0.9\) and \(\tau=0.7\).

The effects were not evenly distributed across threshold comparisons. The transition from \(\tau=0.9\) to \(\tau=0.8\) produced a relatively small mean ESS reduction (0.00304), whereas the transition from \(\tau=0.8\) to \(\tau=0.7\) resulted in a larger reduction (0.01172), primarily due to changes in RS availability (datasets moving to correlated groups).

Overall, the component-wise analysis indicates that ESS sensitivity to the correlation threshold is driven primarily by changes in RS availability and dataset diversity. However, these changes affect only a small subset of task-language combinations, consistent with the high agreement in relative ESS ordering observed in the Kendall's \(W\) analysis. Importantly, component contributions to absolute ESS changes should not be interpreted as direct measures of their influence on relative ranking stability.
 
\section{Sensitivity of ESS to Component Weighting}
\label{appendix:ess_wighting}

To evaluate the robustness of the ESS to the assumption of equal component importance, we conducted a sensitivity analysis in which the weights assigned to the four ESS components were systematically varied. The original ESS formulation assigns equal weights of 0.25 to dataset availability (\(A\)), dataset diversity (\(Div\)), ranking-scheme stability availability (\(RS\)), and dataset-selection stability availability (\(DS\)). The objective of this analysis was to determine the extent to which alternative weighting schemes affect the resulting ESS values and, consequently, whether equal weighting provides a reasonably stable reference formulation.

For each task-language combination \(i\), the weighted ESS was defined as weighted sum of the four components \(j\in\{A,Div,RS,DS\}\),
$ESS_i(\mathbf{w}) =
\sum_{j} w_j x_{ij},$
where \(x_{ij}\) denotes the value of component \(j\) for task-language combination \(i\), and \(w_j\) is its corresponding weight, subject to all weights summing to 1.0. We adopted a one-at-a-time weighting strategy, varying the weight of each component independently over the set
$w_j\in\{0.10,0.20,0.25,0.40,0.60\},$
while distributing the remaining weight equally among the other three components
$w_{k\neq j}=(1-w_j) / 3.$

The configuration \(w_j=0.25\) corresponds to the original equal-weight ESS. The tested values cover both reductions and increases in component importance, including substantial departures from the baseline. This procedure resulted in 17 unique weighting configurations, comprising the shared equal-weight baseline and four alternative weights for each of the four components.

To assess the sensitivity of absolute ESS values, we calculated the mean absolute difference (MAD) between each alternatively weighted ESS and the corresponding equal-weight baseline (Eq.~\ref{eq:ess_mad}), where \(N=96\) is the number of task-language combinations, \(ESS_i^{(j,w,\tau,n_{\min})}\) denotes the ESS obtained by assigning weight \(w\) to component \(j\), and \(ESS_i^{(j,0.25,\tau,n_{\min})}\) represents the original equal-weight ESS calculated under the same correlation threshold \(\tau\) and saturation threshold \(n_{\min}\). A MAD of zero indicates identical ESS values, whereas larger values indicate greater sensitivity to the alternative weighting scheme.

\begin{equation}
\label{eq:ess_mad}
MAD_{j,w}^{(\tau,n_{\min})}
=
\frac{1}{N}
\sum_{i=1}^{N}
\left|
ESS_i^{(j,w,\tau,n_{\min})}
-
ESS_i^{(j,0.25,\tau,n_{\min})}
\right|
\end{equation}

The analysis was repeated for all combinations of correlation thresholds \(\tau\in\{0.7,0.8,0.9\}\) and dataset saturation thresholds \(n_{\min}\in\{5,6,7,8,9,10\}\), resulting in 18 parameter configurations. For each component and tested weight, the MAD values obtained across these configurations were summarized using their mean and standard deviation.

\begin{figure}[!th]
    \centering
    \includegraphics[width=\linewidth]{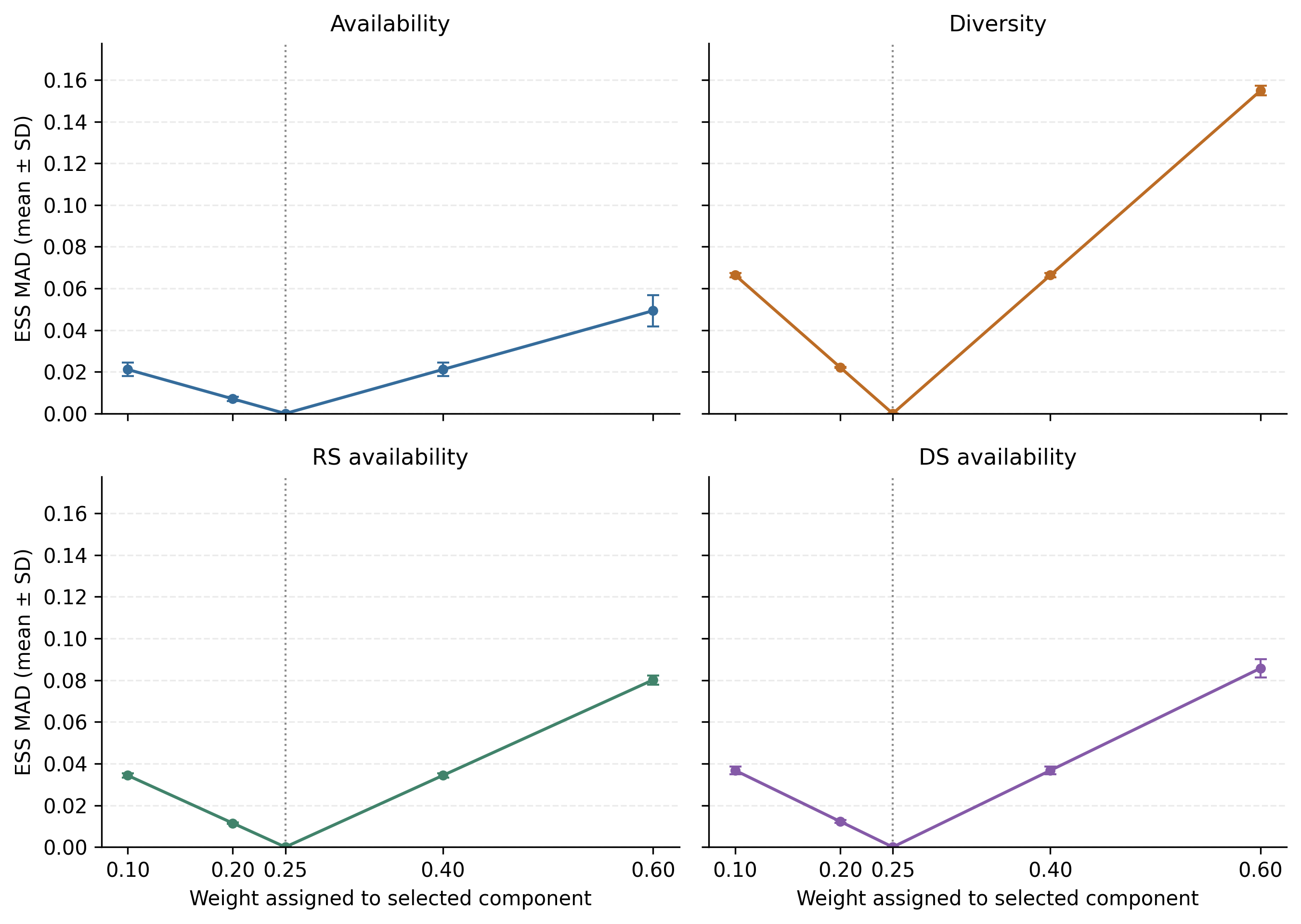}

    \caption{Sensitivity of the ESS to alternative component weights. Each panel presents the MAD between the alternatively weighted ESS and the corresponding equal-weight baseline for one ESS component: (top left) dataset availability, (top right) dataset diversity, (bottom left) ranking-scheme stability availability, and (bottom right) dataset-selection stability availability. Points represent the mean MAD across 18 combinations of correlation threshold \(\tau\) and dataset saturation threshold \(n_{\min}\), while error bars indicate the standard deviation. The vertical dashed line marks the equal-weight baseline (\(w=0.25\)).}
    \label{fig:ess_mad}
\end{figure}

Figure~\ref{fig:ess_mad} presents the mean MAD and corresponding standard deviation for each ESS component across the tested weights. Deviations from this baseline result in progressively larger differences in ESS values, with the magnitude depending on the component being reweighted. Dataset diversity exhibited the greatest sensitivity to weight changes. When its weight was increased to 0.60, the mean MAD reached $0.1548 (\pm0.0023)$, corresponding to approximately 15.5\% of the normalized ESS scale. In comparison, increasing the availability weight to 0.60 produced a mean MAD of $0.0493 (\pm0.0075)$. The corresponding values for RS and DS availability were $0.0801 (\pm0.0022)$ and $0.0857 (\pm0.0044)$, respectively.

The results indicate that ESS values are relatively robust to moderate deviations from equal component weighting, while larger perturbations produce more pronounced changes. In particular, increasing a single component's weight to 0.60 represents a substantial departure from the original formulation, as that component receives 2.4 times its baseline weight, while each remaining component receives approximately 0.133 weight. Even under this relatively extreme configuration, mean absolute differences remain below 0.16 on the normalized ESS scale of 0-1.

The standard deviations are generally small, suggesting that the magnitude of weighting sensitivity is relatively consistent across the examined correlation and saturation thresholds. Nevertheless, the differences between components indicate that the ESS formulation is more sensitive to the weighting of dataset diversity than to the weighting of the other components.

Overall, the weighting sensitivity analysis indicates that equal weighting provides a reasonably stable reference formulation within the examined parameter ranges. Moderate changes in component importance produce limited changes in absolute ESS values, whereas substantial deviations can lead to non-negligible differences, particularly when emphasizing dataset diversity. These findings support the use of equal weighting as a transparent and reproducible baseline, while acknowledging that the relative importance assigned to individual components influences the resulting ESS values. Importantly, the analysis establishes robustness within the tested weighting scenarios rather than demonstrating that equal weighting is uniquely optimal.

\section{Sensitivity Analysis of $k$ in Task-Specific Top-$k$ Consistency}
\label{appendix:topk_sensitivity}

This section examines the sensitivity of the proposed task-specific rank-weighted top-$k$
consistency measure to the choice of the parameter $k$. The analysis considered
$k\in\{5,8,10,12,15,20\}$, with $k=10$ serving as the reference
configuration (also used in the main results). For each value of $k$, consistency scores were
recomputed using the same underlying rankings and a fixed set of
models, defined as the union of models appearing among the top 20
in at least one ranking scheme. This ensured that differences between
configurations resulted exclusively from changes in $k$, rather
than changes in the evaluated model population.

The analysis covered 58 task-language combinations for which
task-specific model consistency could be computed. For each
combination $(t,l)$, we quantified score sensitivity using the
MAD between the consistency scores
obtained with $k$ and those obtained with the reference value
$k=10$ (Eq.~\ref{eq:topk_mad}). $N_{t,l}$ denotes the number of models in the fixed
top-20 union for the corresponding task--language combination.

\begin{equation}
\label{eq:topk_mad}
\mathrm{MAD}_{k}^{(t,l)}
=
\frac{1}{N_{t,l}}
\sum_{i=1}^{N_{t,l}}
\left|
TC_{k}^{(t,l)}(i)
-
TC_{10}^{(t,l)}(i)
\right|
\end{equation}

To assess whether changes in $k$ affected the relative ordering
of models, we additionally computed Spearman's rank correlation
coefficient between the consistency scores obtained with any value of $k$
and $k=10$.
Both measures were calculated separately for each task-language
combination and subsequently aggregated by taking the arithmetic
mean across combinations, thereby assigning equal weight to
each combination irrespective of the number of evaluated models.
Standard deviations were used to characterize variability
across combinations.

\begin{figure}[!th]
    \centering
    \includegraphics[width=\linewidth]
    {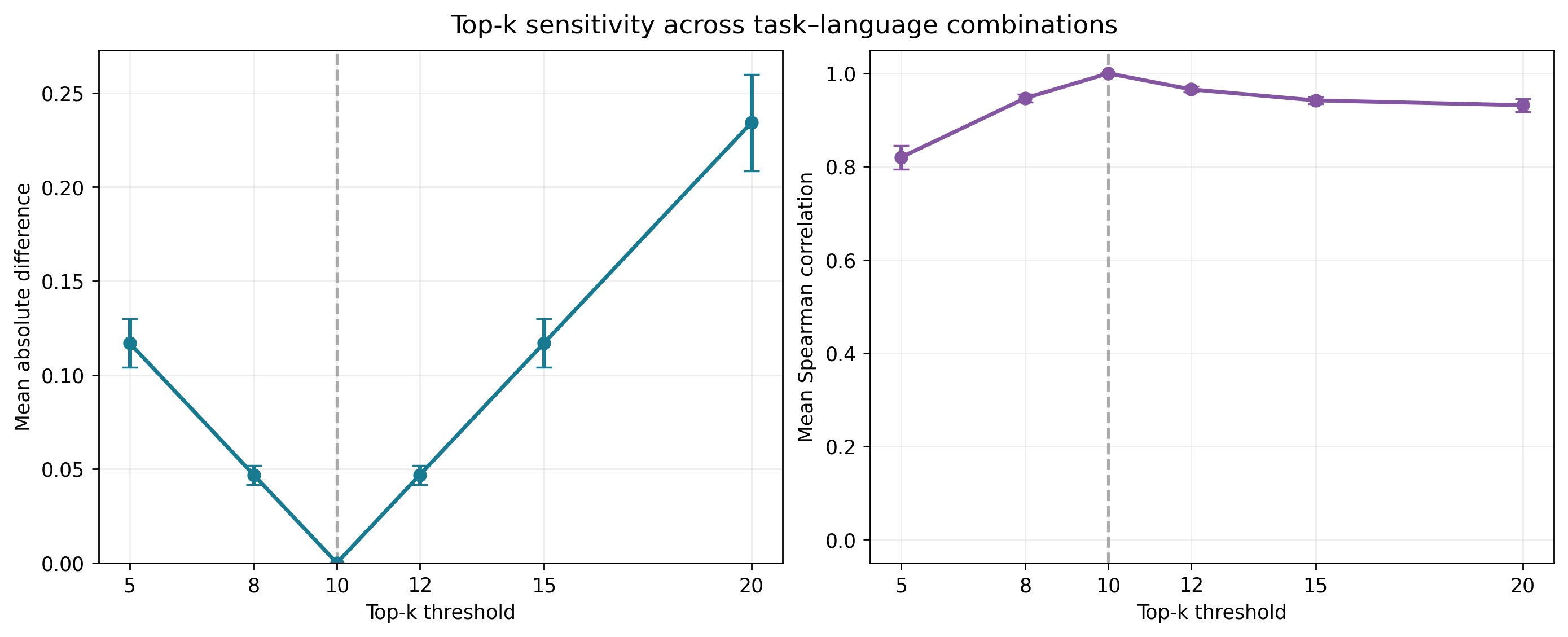}
    \caption{Sensitivity of rank-weighted top-$k$ model
    consistency to the choice of $k$, relative to the reference
    value $k=10$. The panels show the mean absolute difference
    (MAD) in consistency scores and the mean Spearman rank
    correlation across 58 task--language combinations. Error
    bars represent one standard deviation across combinations.}
    \label{fig:topk_sensitivity_overall}
\end{figure}

Figure~\ref{fig:topk_sensitivity_overall} summarizes the overall
results. The smallest deviations from the reference configuration
were observed for $k=8$ and $k=12$, both yielding a mean MAD of
0.0468. The corresponding mean Spearman correlations were
0.9469 and 0.9656, respectively, indicating substantial
preservation of model ordering. Increasing the distance from
the reference value of $k$ resulted in larger absolute differences.
For $k=5$ and $k=15$, mean MAD was approximately 0.1170,
whereas $k=20$ produced the largest mean MAD of 0.2342.
Nevertheless, the mean Spearman correlation remained relatively
high for $k=15$ (0.9420) and $k=20$ (0.9317). In contrast,
$k=5$ yielded a lower mean correlation of 0.8200, indicating
greater sensitivity of model ordering to a more restrictive
definition of top-performing models.

\begin{figure}[!th]
    \centering
    \includegraphics[width=\linewidth]
    {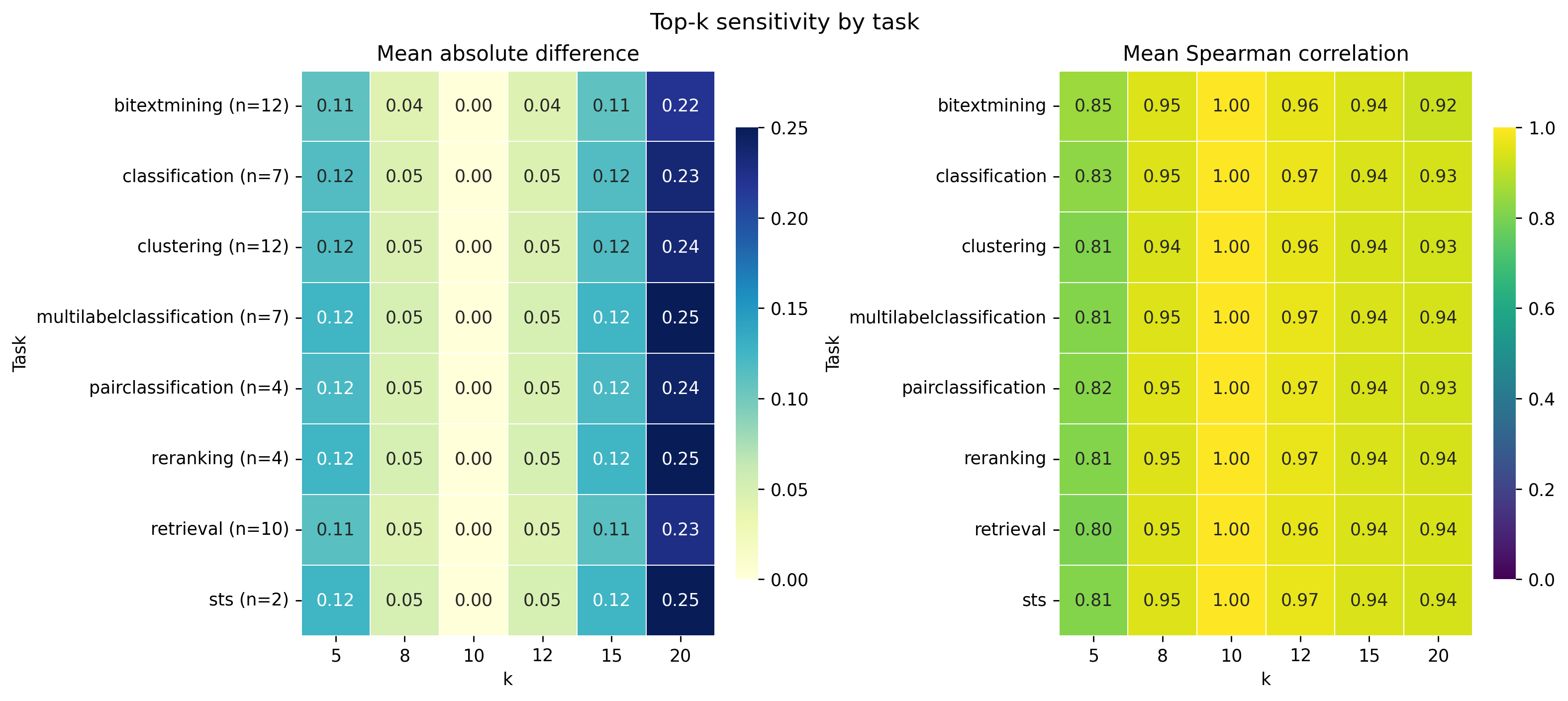}
    \caption{Task-specific top-$k$ score sensitivity to the $k$
    parameter by task. Heatmaps report mean MAD and mean Spearman
    rank correlation relative to $k=10$, aggregated across
    available languages for each task ($n$ value given in brackets).}
    \label{fig:topk_sensitivity_tasks}
\end{figure}

\begin{figure}[!th]
    \centering
    \includegraphics[width=\linewidth]
    {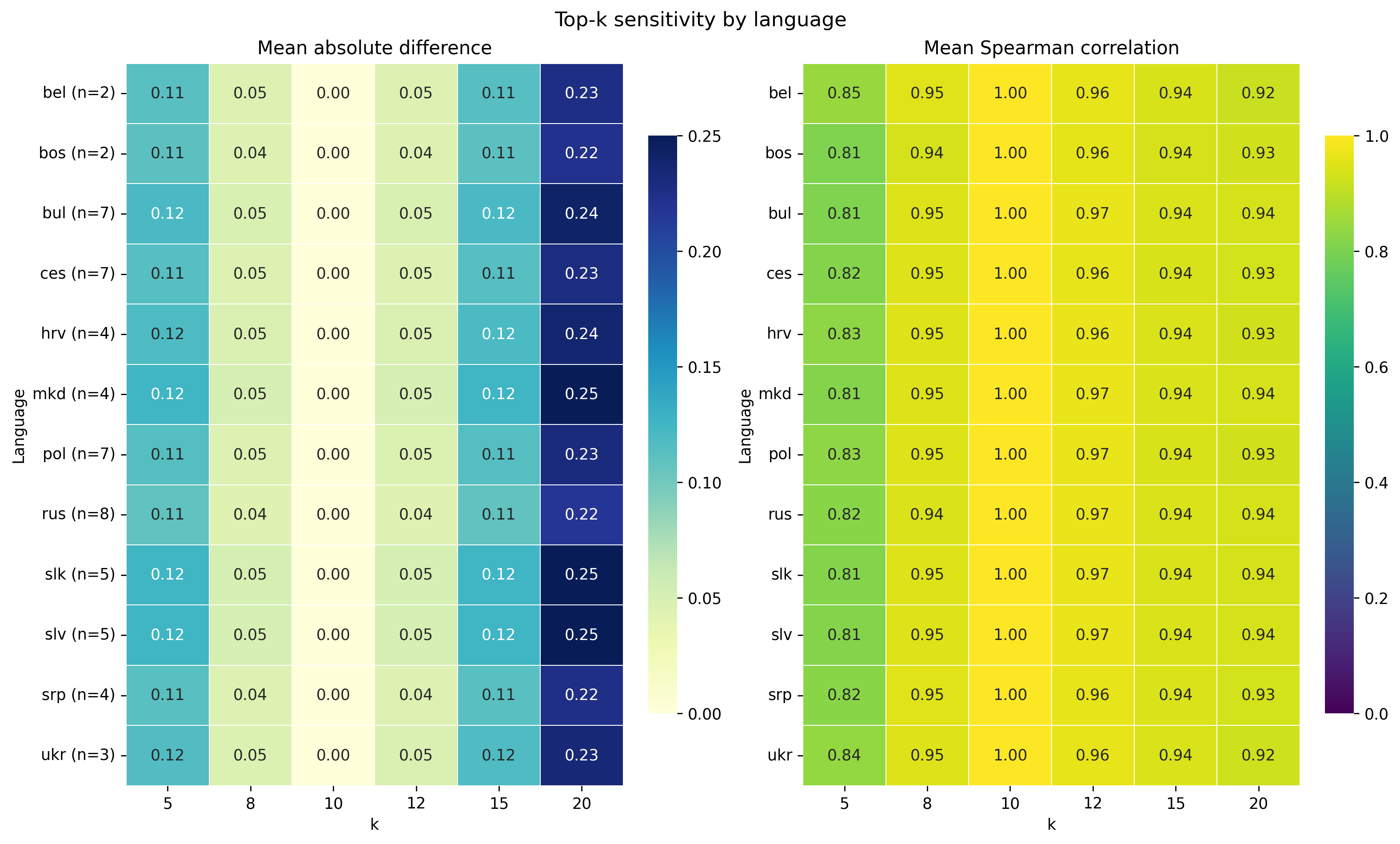}
    \caption{Task-specific top-$k$ score sensitivity to the $k$
    parameter by language. Heatmaps report mean MAD and mean Spearman
    rank correlation relative to $k=10$, aggregated across
    available tasks for each language ($n$ value given in brackets).}
    \label{fig:topk_sensitivity_languages}
\end{figure}

\begin{figure}[!th]
    \centering
    \includegraphics[width=\linewidth]
    {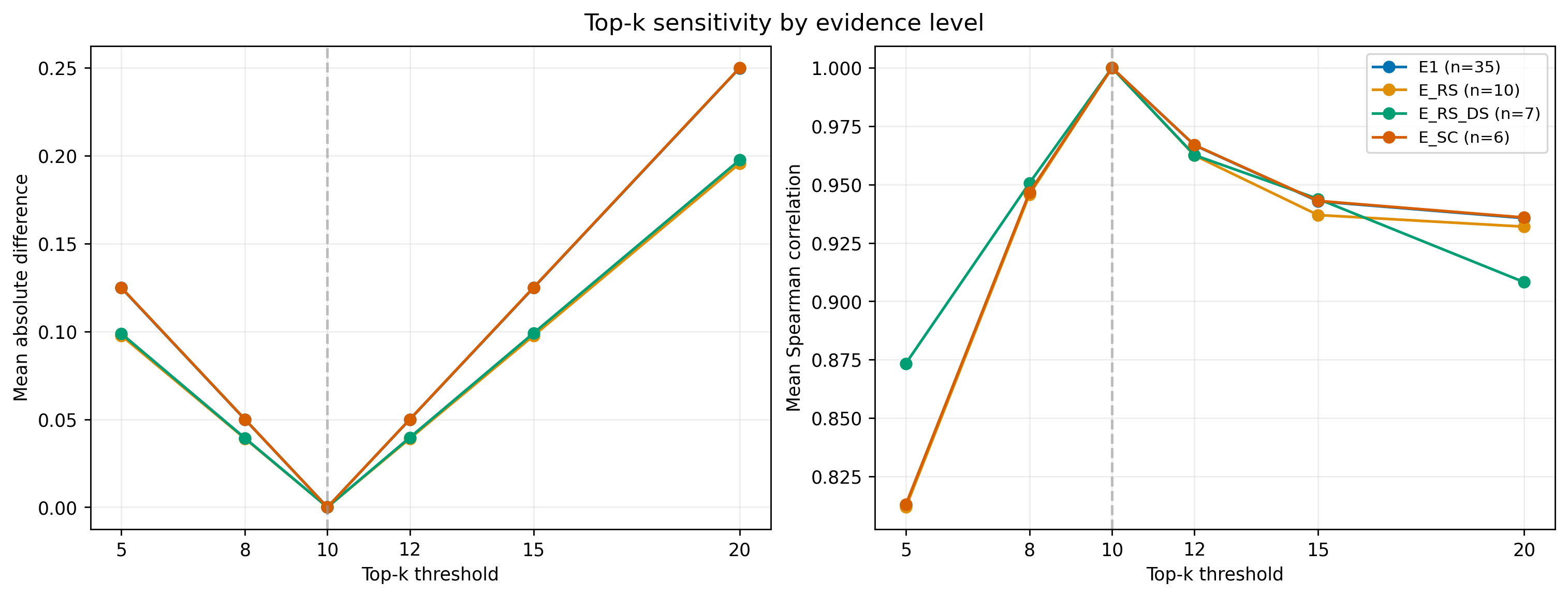}
    \caption{Task-specific top-$k$ score sensitivity to the $k$
    parameter by
    evidence level. The panels show mean MAD and mean
    Spearman rank correlation relative to $k=10$ for
    task-language combinations grouped by evidence level ($n$ value given in brackets).}
    \label{fig:topk_sensitivity_evidence}
\end{figure}

To investigate potential heterogeneity, we further aggregated
the task-language-level MAD and Spearman coefficients separately
by task, language, and evidence level. Task-specific results
were obtained by averaging across the available languages
within each task, whereas language-specific results were
obtained by averaging across the available tasks within each
language. Evidence-level results were obtained by averaging
across combinations belonging to the same evidence category.
The corresponding comparisons are presented in
Figures~\ref{fig:topk_sensitivity_tasks},
\ref{fig:topk_sensitivity_languages}, and
\ref{fig:topk_sensitivity_evidence}.

Across tasks, sensitivity was generally limited for values of $k$ close to the reference configuration, while larger deviations were observed for more distant values. At $k=20$, mean MAD ranged from 0.219 for bitext mining to 0.250 for multilabel classification, reranking, and STS. Despite these differences in consistency scores, mean Spearman correlations remained above 0.91 across all tasks, indicating that model ordering was largely preserved. A similar pattern was observed across languages: at $k=20$, mean MAD ranged from 0.217 to 0.250, while mean Spearman correlations ranged from 0.918 to 0.936.

The sensitivity results by evidence level should be interpreted in light of the evidence level. For $E_1$ and $E_{SC}$, top-$k$ consistency is calculated from a single ranking. Consequently, changing $k$ only changes the weights assigned to model positions in that ranking, without introducing variation across ranking schemes or dataset compositions. For $E_{RS}$ and $E_{RS+DS}$, consistency is averaged across multiple ranking schemes and, in the latter case, dataset compositions. Changing $k$ can therefore also affect how differences between these rankings contribute to the final consistency scores. Thus, although Spearman's correlation measures changes in model ordering consistently across all evidence levels, its interpretation differs: in single-ranking scenarios, it reflects sensitivity to the top-$k$ weighting function alone, whereas in multiple-ranking scenarios, it also reflects how the aggregation of rankings responds to changes in $k$.

The evidence-level analysis revealed a more pronounced distinction in score sensitivity. At $k=20$, the single-ranking scenarios ($E_1$ and $E_{SC}$) exhibited mean MAD values of approximately 0.250, compared with 0.196 and 0.198 for $E_{RS}$ and $E_{RS+DS}$, respectively. However, these smaller score deviations did not necessarily correspond to greater preservation of model ordering. In particular, $E_{RS+DS}$ exhibited the lowest mean Spearman correlation at $k=20$ (0.908), compared with approximately 0.936 for both single-ranking scenarios. These findings suggest that averaging across multiple rankings can reduce changes in the magnitude of consistency scores without necessarily reducing changes in the relative ordering of models.

Overall, the results indicate that the task-specific top-$k$ consistency measure
is relatively insensitive to moderate changes around $k=10$,
particularly with respect to the relative ordering of models.
Larger changes in $k$ produce more pronounced differences in
consistency scores, as expected because the parameter directly
determines both the set of ranks receiving positive weight
and the weight assigned to each position. The results
therefore support $k=10$ as a reasonable reference
configuration, while not establishing it as a uniquely
optimal choice.

\end{document}